%% file: main.tex
\documentclass{article}
\usepackage[final]{corl_2018}

\usepackage{times}
\usepackage{wrapfig}
\usepackage{natbib}
\setcitestyle{numbers,sort&compress}
\usepackage{multicol}
\usepackage{hyperref}
\usepackage{amsmath}
\usepackage{amssymb}
\usepackage{algorithm,algpseudocode}
\usepackage{graphicx}
\usepackage{capt-of,etoolbox}
\usepackage{caption}
\usepackage{array,multirow}

\DeclareMathOperator*{\argmax}{arg\,max}

\algnewcommand{\Inputs}[1]{%
  \State \textbf{Inputs:}
  \Statex \hspace*{\algorithmicindent}\parbox[t]{.8\linewidth}{\raggedright #1}
}
\algnewcommand{\Initialize}[1]{%
  \State \textbf{Initialize:}
  \Statex \hspace*{\algorithmicindent}\parbox[t]{.8\linewidth}{\raggedright #1}
}

\usepackage[resetlabels]{multibib}
\newcites{apndx}{References}

\title{Personalized Dynamics Models \\ for Adaptive Assistive Navigation Systems} 
 \author{
   Eshed Ohn-Bar \hspace{.2in} Kris Kitani \hspace{.22in} Chieko Asakawa \vspace{.1cm}\\
   The Robotics Institute, Carnegie Mellon University
 }
\begin{document}
\maketitle

\begin{abstract}
Consider an assistive system that guides visually impaired users through speech and haptic feedback to their destination. Existing robotic and ubiquitous navigation technologies (\emph{e.g.}, portable, ground, or wearable systems) often operate in a generic, user-agnostic manner. However, to minimize confusion and navigation errors, our real-world analysis reveals a crucial need to adapt the instructional guidance across different end-users with diverse mobility skills. To address this practical issue in scalable system design, we propose a novel model-based reinforcement learning framework for personalizing the system-user interaction experience. When incrementally adapting the system to new users, we propose to use a weighted experts model for addressing data-efficiency limitations in transfer learning with deep models. A real-world dataset of navigation by blind users is used to show that the proposed approach allows for (1) more accurate long-term human behavior prediction (up to 20 seconds into the future) through improved reasoning over personal mobility characteristics, interaction with surrounding obstacles, and the current navigation goal, and (2) quick adaptation at the onset of learning, when data is limited. 
\end{abstract}



\input{intro.tex}

\input{related.tex}

\input{approach.tex}

\input{experiment.tex}

\input{analysis.tex}

\section{Conclusion} 
\label{sec:conclusion}
Adaptive assistive navigation systems have the potential to better meet the dynamic information needs of the blind end-user. Informed by a real-world navigation study, we proposed an approach for adapting a dynamics model across users. The models can be used to personalize instructional guidance, hence taking a step towards less manual design of generic interfaces that fail to flexibly accommodate large variability in blind user behavior. 

Our study focused on understanding and learning user dynamics, with several future directions. First, we would like to continue and study issues in scalability of navigation interfaces, where the personalization must efficiently deal with an ever increasing pool of users. Although the weighted experts model was designed to handle scenarios where the pre-trained expert models may be suboptimal, a practical realization could involve an additional sampling mechanism over the experts. Learning more robust long-term prediction models that can reason over uncertainty can also be pursued in the future. Extending the personalized navigation interface framework to plan over the path and the instructions concurrently can allow it to choose more user-friendly routes. While we emphasized prediction and timing during the most critical navigation task of turning ({\em i.e.} concise notifications), we can see how a more elaborate instruction generation planner ({\em i.e.} natural language) could be added. Surrounding pedestrians can also be detected and incorporated into the state. Given that we studied practical design issues in scalable adaptive navigation interfaces, an important next step would be further validation in real-world, large-scale studies.

\section{Acknowledgments}

This work was sponsored in part by a JST CREST (JPMJCR14E1), NSF NRI award (1637927), and the Shimizu Corporation. We greatly appreciate assistance of collaborators Daisuke Sato, Hironobu Takagi (IBM Research Tokyo), Jo\~{a}o Guerreiro, and Dragan Ahmetovic (Cognitive Assistance Lab).

\bibliographystyle{plainnat}
\bibliography{references}

\input{appendix.tex}

\end{document}

%% file: intro.tex
\section{Introduction}
\label{sec:intro}

\textit{Generalized} assistive navigation systems for people with visual impairments have been extensively studied in the robotics community over the past decades \cite{1191706,survey2,LACEY1998245,icrabackpack,Manduchi:2014:LMB:2556288.2557328,faria2010electronic,luca2016towards,legge,fallah2012user}. However, many of these systems have not considered \textit{personalization}, to take into account individual differences between users~\cite{variabilityohnbar,Azenkot:2016:EBS:2906831.2906835}. Differences between users can significantly impact the navigation experience and the ultimate success of the assistive navigation system. Based on our analysis of real-world user data, we observe high variability in the reaction time and response to instructional guidance provided by the navigation system. Towards a more personalized interaction experience, we envision an assistive navigational system that is able to automatically identify the user's specific information needs (\emph{e.g.,} environmental context, walking speed, reaction time, and cane strategies) and adapt the timing and content of its instructions accordingly in real-time. 

\begin{figure}[!t]
   \centering
  \includegraphics[trim=0cm 9.25cm 15.7cm 0cm,clip,width=4in]{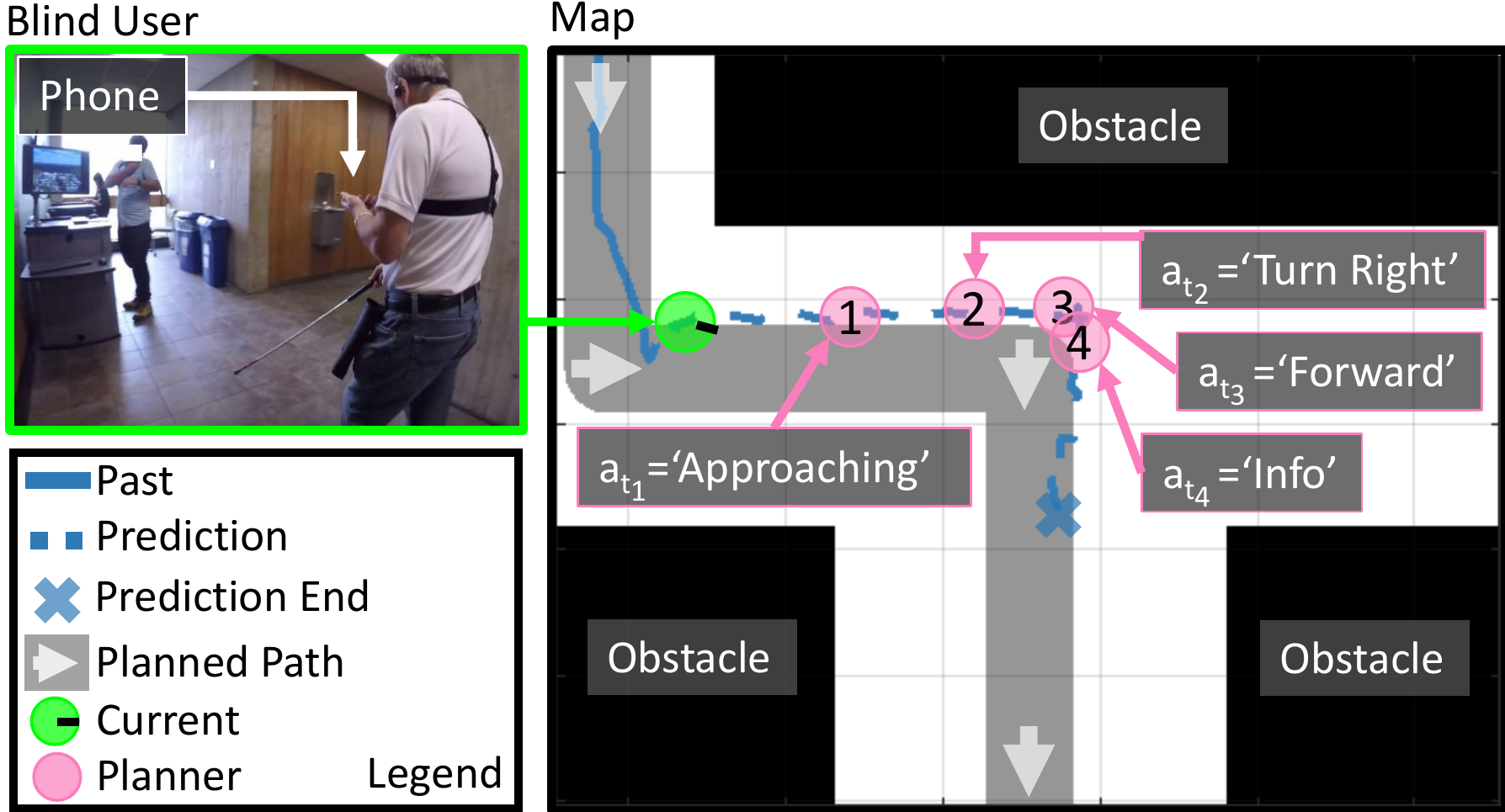}
   \caption{\textbf{Blind user dynamics modeling for adaptive assistive navigation interfaces.} Towards a more personalized interaction experience, our approach (PING) learns robust dynamics models of user behavior and plans for a sequence of instructions for arrival to a navigation goal.
   }~\label{fig:fig1final}
   \vspace{-0.32in}
   \end{figure}

Designing a system which can personalize its interaction across users is challenging. Consider an assistive system (\emph{e.g.,} a service robot, wearable, or smartphone system) that instructs a blind person along the planned path in Fig.~\ref{fig:fig1final}. As the system observes the user approaching a turn, it must determine when to announce `turn right.' In this scenario, a slight delay or premature signaling of a turn can result in a navigation error. For instance, the user could miss the turn or get confused by proceeding into the wrong hallway or encountering a wall. Hence, we can see that the ability of a navigational system to give appropriately timed instruction is critical, in particular when guiding a blind person in intricate indoor environments. However, appropriately selecting the time and content of instructional feedback requires knowledge of the user's dynamics, which are not known a priori (as we demonstrate) and must be inferred on the fly.

We propose to frame the interaction between the assistive navigational system and the blind user as a Markov Decision Process (MDP). We term our complete reinforcement learning (RL) framework \textbf{PING} (for \textbf{P}ersonalized \textbf{IN}struction \textbf{G}eneration). Actions in this dynamical system are sampled from a policy $P(a|s)$ (the solution to the MDP), which maps the current spatial context and user state $s$ (\emph{e.g.}, user position and orientation) to an instructional guidance action $a$ (\emph{e.g.}, `turn left' or `proceed forward'). 

While adaptive human-robot interaction (HRI) systems have been previously proposed~\cite{torrey13, tariq,dorsa2,zhou2017expressive,daniele17,daniele16,AAAI113623}, quick adaptation to new users of a system is still an open problem. For instance, model-free RL algorithms do exist for learning optimal policies by sampling state-action transitions during interaction with the environment~\cite{mnih2015human,watkins1989learning}, yet those approaches can require much time and exploration (\emph{i.e.}, trial and error) to find the optimal policy. In the context of navigating a blind person through a new environment, we simply do not have the capacity for such costly exploration of the policy space. Instead, to minimize navigation errors in our application, we must develop a technique that can adapt as fast as possible to the user's behavior. As model-based approaches are known to benefit from sample efficiency~\cite{pilco,NNdynamics}, we emphasize approximating the state transition function $P(s'|s,a)$, which in our scenario is a model of how a specific user will react to the guidance.
\subsection{Contributions}
We build on previous research in adaptive learning~\cite{torrey13,tariq,dorsa2}, human dynamics modeling~\cite{sfmodelorig,7780479,6907442,ELFRING2014591,Ziebart:2009:PPP:1732643.1732694,kitani2012activity,fragkiadaki2015recurrent}, and model-based RL~\cite{7487509,Lenz2015DeepMPCLD,pilco,NNdynamics,levine2014learning1,doerr2017optimizing} to make the following contributions.

\textbf{Dataset for Predicting Real-World Navigation by Blind People.} To understand practical issues in assistive navigation, we study a large dataset collected in real-world assisted navigation. The dataset is used to perform long-term user prediction, and includes unique challenges compared to common trajectory prediction datasets which observe humans passively ({\em e.g.}, surveillance or social settings~\cite{7780479}), such as delayed reactions, veering, and highly non-linear trajectories during turns. We use the dataset to identify a practical aspect of the interaction which has been little explored before--the extent of inter-user variability and need for real-time personalization in assistive navigation.

\textbf{User-Specific Dynamics Model via an Efficient Weighted Experts Algorithm.} Our main contribution is in a personalization scheme which allows for training high capacity \textit{and} sample efficient dynamics models. We demonstrate that a weighted experts model is able to quickly adapt the state transition dynamics model to match the reactions of the user nearly an order of magnitude faster when compared to the common transfer learning (\emph{i.e.,} adaptation) baseline of single model fine-tuning. When evaluating long-term future prediction of user position with a 20 second horizon, the proposed model is able to improve prediction accuracy by nearly the width of a hallway on average compared to the baseline, especially at the crucial navigation junctions of turns or veering paths. 

\textbf{Assistive Navigation System with Adaptive Instruction Generation for Blind People.} Towards a system which can adapt to a variety of blind walkers, we pair the proposed dynamics personalization scheme with a Model Predictive Control (MPC)-based planner, and evaluate it on a diverse set of simulated walkers. To the best of our knowledge, PING is the first application of ideas from model-based RL and online learning to develop a continuously adapting assistive navigation system for people with visual impairments.

%% file: related.tex
\section{Related Research}
\begin{table}[!t]
\centering
\caption{Representative related research studies in assistive navigation systems.}~\label{tab:report}
\vspace{0.1cm}
\resizebox{14.2cm}{!}{
\centering
    \begin{tabular}{l |p{2.6cm} | p{1.2cm} | p{2.5cm} | p{1.39cm} | p{1.5cm} | p{2.5cm} | p{1.5cm} |p{1.6cm}  }
            Reserch Study & Participants  & Indoor & Personalized Route Preference & Visual Impairment & Motion Tracking & Real-World & Route Length &   Adaptive Interaction (none) 
            \\
            \hline
            Wang~\emph{et al.}~\citeapndx{icrabackpack} & 11 (only 3 in real) & \checkmark & & \checkmark & \checkmark & \checkmark  & 90m &  \\
             Baldi~\emph{et al.}~\cite{baldi2018haptic} & 18 & \checkmark &  &  &  &  & 4m &    
           \\
            Zhang and Ye~\cite{zhang2017indoor} & 7 & \checkmark &  &  & \checkmark  & \checkmark  & 45m &   
            \\
            Akasaka and Onisawa~\cite{akasaka2008personalized} & 11 &   & \checkmark  &  & \checkmark & & - &  \\
            Thorsten and Gerhard~\cite{Volkel} & - & \checkmark & \checkmark & \checkmark & & & - &  \\
            Macik~\emph{et al.}~\cite{macik2015software} & - & \checkmark & \checkmark & & & & - &  \\
            Fallah~\emph{et al.}~\cite{fallah2012user} & 6 & \checkmark  &  & \checkmark & \checkmark & \checkmark & 55m &  \\ 
            Riehle~\emph{et al.}~\cite{riehle2013indoor} & 8 & \checkmark  &  & \checkmark & \checkmark & \checkmark & 160m &  \\ 
            Ahmetovic~\emph{et al.}~\cite{ahmetovic2016navcog2} & 6 & \checkmark  &  & \checkmark & \checkmark & \checkmark & 350m & \\
            Ivanov~\cite{ivanov2010indoor} & 8 & \checkmark & & \checkmark  & \checkmark & \checkmark & - & \\
     Adebiyi~\emph{et al.}~\cite{adebiyi2017assessment} & 11 & \checkmark  &  & \checkmark &  & \checkmark & 55m &  \\
        \end{tabular}
}
\end{table}


\textbf{Modeling Human Dynamics.} The goal of our work is to adapt an assistive interface to the current user by learning a personalized dynamics model. By formulating assistive navigation in a model-based framework, our work is related to data-efficient RL frameworks~\cite{7487509,finn2017deep,Lenz2015DeepMPCLD,rsspilco,pilco,NNdynamics,levine2014learning2,levine2014learning1}. Model-based approaches often employ a non-parametric Gaussian Process~\cite{pilco} or Neural Network~\cite{fragkiadaki2015recurrent} dynamics model, as in our study. Modeling the dynamics of people is an important yet challenging problem, extensively studied in literature~\cite{torrey13, tariq,dorsa1,dorsa2,zhou2017expressive}. Much attention has been given to predicting human behavior under passive observation ({\em e.g.}, surveillance or social settings ~\cite{Bulling:2014:THA:2578702.2499621,sfmodelorig,5995468,basket,7780479,5509779,6907442,ELFRING2014591,Ziebart:2009:PPP:1732643.1732694,kitani2012activity,jain2016structural,fragkiadaki2015recurrent}). While relevant, such studies do not generally model interaction with an instructional interface, nor do they study the role of personalization for modeling the behavior of a blind person. Moreover, the prediction horizon studied in related literature is often short, not allowing for planning over delayed reactions, navigation tasks which may span multiple seconds (as users may stop to listen to instructions, scan the environment with their cane before proceedings, \emph{etc.}), and allow for sufficient plan-ahead (\emph{i.e.,} including several seconds of instruction utterance). In contast, we focus on long-term prediction of blind users' trajectories during interaction with a navigation system of up to 20 seconds into the future. 


\textbf{Designing Instructional Guidance Systems.} Previous studies suggest that users with visual impairements generally prefer speech output during navigation~\cite{arditi2013user}. Designing appropriate instructional guidance for such users has been extensively studied in research~\cite{look2005location,waller2007landmarks,4287440}. In particular, a diverse set of factors is known to contribute to navigation performance, including perceptions towards technology and previous experience~\cite{ohn2018modeling,guerreiro2017virtual}, Orientation and Mobility (O\&M) skills and the ability to deal with unknown environments, level of impairment, and more~\cite{fallah2012user,quinones2011supporting,survey2,baus2007auditory,sanchez2014audio,ahmetovic2016navcog2,banovic2013uncovering,hriblind,Azenkot:2016:EBS:2906831.2906835}. The possible design choices motivate our approach of an adaptive interface which can better meet the dynamic and specific needs of different types of end-users.


\textbf{Reinforcement Learning for Adaptive Assistive Interfaces.} We aim to find a policy that is optimized to provide clear and concise navigation instructions. Hence, the research in learning instructional timing and content is related to dialogue and human-robot communication systems~\cite{AAAI113623}. RL provides a framework for interactive learning
with an end-user, in particular for assistive applications~\cite{amershi2014power,thomaz2006reinforcement,wang2003user,Hemminghaus17,kober2013reinforcement}. Several previous studies formulated general collaborative navigation interfaces (\emph{i.e.}, for sighted users) in an MDP framework~\cite{cuay,6907334,Seo:2000:RLA:325737.325859,dealorno}. To select and realize human-like instructions, methods may also employ a database of human-generated instructions~\cite{daniele17,daniele16}. In contrast, we do not assume access to such demonstrations, and complement previous studies by emphasizing efficient system adaptation. Specifically in the domain of assistive technologies for blind people, the overall majority of systems are not adaptive but static in their operation, and an in-depth analysis of the role of real-time personalization is lacking in literature. Instead, related studies often provide a preliminary prototype or analysis in a highly simplified environment~\cite{icrabackpack,baldi2018haptic,zhang2017indoor}, limiting insight into the role of personalized instruction. 



%% file: approach.tex
\section{Personalized Instruction Generation for Navigation}

In order to provide a personalized instructional interface for navigation, we will model the interaction between the navigation system (agent) and the user (environment) as an MDP, described in Section \ref{sec:def}. The MDP will be adapted to each user by learning a user-specific dynamics model in an incremental manner. Specially, the user-specific dynamics models will be learned using a weighted majority algorithm and a recurrent network based function approximator to represent the user's dynamics model, described in Section \ref{sec:dynamics}. Based on the MDP and the learned user-specific dynamics model, we employ Model Predictive Control to determine the optimal instructional feedback (\emph{i.e.}, action of the MDP) for a personalized human-computer interface, described in Section \ref{sec:mpcontrol}.

\begin{figure}[t]
\begin{minipage}{0.99\textwidth}
\begin{algorithm}[H]
 \caption{PING: \textbf{P}ersonalized \textbf{IN}struction \textbf{G}eneration Agent for Assistive Navigation}
    \begin{algorithmic}[1]
   	\Procedure{LearnPersonalizedAgent}{Layout, Localizer, ParticleFilter, EndGoal, $\theta_{1:M}$ (pretrained weights), $\pi_0$ (initial policy), $L$ (planning horizon), $(\eta, \lambda)$ (update parameters), MDP($\mathcal{S}$, $\mathcal{I}$, $\mathcal{T}$, $\mathcal{R}$,  $\gamma$)}
		\State $s_0$.Init(Localizer), $\theta$.Init($\theta_{1:M}$)
		\State $D$ $\gets \{\}$, $\pi \gets \pi_0$, $i = 0$
			  \State $p \gets $ PathPlan($s_0$, Layout.Graph, EndGoal)
			\While{$s_i \neq$ EndGoal}
			\State $s^g$ $\gets$ DetermineNextWayPoint($L$, $p$, $s_i$)
			\State $s_i^H = [x_i,y_i,\alpha_i] \gets$ ParticleFilter.Track($s_i$)
			\State $s_i^O = [o_i,l_i] \gets$ Layout.Occupancy($s_i^H$)
			\State $s_i$ $\gets$ $[s_i^H, s_i^O]$, $a_i \gets \pi(s_i)$, $s_{i+1} \gets \mathcal{T}(s_i,a_i)$
			\State $D \gets D \cup \{$Reparameterize$(s_{i+1},s_i,a_i) \}$  
	       \State is\textunderscore model \textunderscore update $\gets$ DetermineUpdate($L$, $D$)
		  \If{ is\textunderscore model\textunderscore update }
			
			  	\State $\theta \gets$ {UpdateWeights}($D$, $\theta$, $\eta$, $\lambda$) (Algorithm~\ref{alg:update})
			  	\State $D \gets \{\}$				
			\EndIf
			\State $\pi(s_{i+1}) \gets$  MPCPlan($\pi$, $D$, $p$, $s^g$, Layout.Occupancy,   $\theta$, $L$, $\gamma$) 
			(Equation~6)
			\State is\textunderscore off\textunderscore path $\gets$ DistanceThreshold($s^H$, $p$)
			\If{ is\textunderscore off\textunderscore path }
				\State $p \gets $ PathPlan($s_i$, Layout.Graph, EndGoal)
			\EndIf	
			\State $i \gets i+1$
		
		\EndWhile
		\EndProcedure
    \end{algorithmic}
    \label{alg:ping}
\end{algorithm}
\end{minipage}
\end{figure}

\subsection{Formulation}\label{sec:def}
The adaptive navigation system PING is an MDP defined by a tuple~({$\mathcal{S}, \mathcal{I}, \mathcal{T},\mathcal{R}, \gamma$}). $\mathcal{S}$ and $\mathcal{I}$ are the sets of states and possible actions (\textit{i.e.}, navigation instructions), respectively. Given a state ${s}_t \in \mathcal{S}$ at time $t$, the instructional system performs an action ${a}_t \in \mathcal{I}$ and observes a reward and an updated state $s_{t+1} = \mathcal{T}(s_t,a_t)$ (the transition function). The learning goal is to solve for a policy, $\pi(s_t) = p(a_t|s_t)$, which is a mapping from state to action. Actions taken should maximize the discounted sum of future rewards, $\sum_{t' = t}^{t+L-1} \gamma^{t' - t}r({s}_{t'},{a}_{t'})$, where $\gamma \in (0,1]$ is a discount factor and $1 < L < \infty$ a time horizon for planning.

\textbf{States.} As the blind user is reacting to the instructions provided by the interface as well as the environmental context, we decompose the observations of the system to two components, a human user component $s^H$, and a surrounding occupancy component $s^O$. The state is then defined as 
\begin{equation}
s = [s^H, s^O] = [x,y,\alpha,o,l]
\end{equation}
where the state of the human is the position $(x,y)$ and heading $\alpha$ of the user in the map, and the surrounding state $s^O$ includes occupancy and landmark features. For the obstacle descriptor ${o}$ a vector of binary values encoding obstacles in discretized ego-centered and ego-oriented polar spatial bins around the user is used (we use 8 orientation and 3 distance bins). A similar encoding is used to represent landmark types (annotated in the floor plan) ${l}$ as one-hot vectors. 

\textbf{Actions.} We define a set of actions for the interface to provide instructional guidance. While current state-of-the-art assistive turn-by-turn navigation systems~\cite{navcog3} select speech instructions from a fixed vocabulary of about 100 possibilities, their possible combinations ({\em e.g.}, distance amounts, landmark types) lead to a problematically large action set. Instead of directly operating in this large set, the instructions are clustered into a set of semantically similar groups including `Landmark,' `Forward,' `Approaching a Turn,' `Turn,'  `Diagonal Turn,' `Slight Turn,' `Obstacle,' `Information,' and `Distance from Point of Interest.' These types ($a_{type}$) may be joined with additional arguments, the ego-centered direction ($a_{dir}$, {\em e.g.}, `Left,' `Right') or a quantity amount ($a_{amount}$, {\em e.g.}, `Diagonal Turn', `Slight Turn'). Also, two types of vibration feedback occur during turns, once at a turn onset notification, and again when correct heading is achieved. The actions $a \in \mathcal{I}$ are encoded as a concatenation of three one-hot vectors,
\begin{equation}
a = [a_{type}, a_{dir},a_{amount}]
\end{equation}
In this work, this semantically clustered language space is used to efficiently learn dynamics models. Clustering is commonly employed in related studies for learning navigational interfaces, and decoding it further into natural language ({\em i.e.,} realization) can be studied in the future~\cite{daniele17}. 

\textbf{Reward.} While we focus on quickly learning personalized dynamics models, a suitable reward function for path-following is also defined for planning instructions with PING. The navigation task reward is a function of the current state, a planned path, a set of waypoints, and the floor plan layout. First, we reduce the map into a graph of connected links and nodes in the floor plan ({\em i.e.,} corridors are reduced to a single link placed at the center of the corridor) and plan a global path with A*~\cite{hart1968formal}. The task is decomposed into subtasks ({\em e.g.}, completing a turn), defined via waypoints at fixed distance thresholds following turns and landmarks. Arrival to within a distance of 1.5 meters to a waypoint with the correct heading results in a trivial reward 0. The agent receives a negative reward of at every time step to encourage timely arrival to the destination. This magnitude is decreased further depending on the distance to the planned graph link (-1 within 1.5m, lower otherwise). We select 1.5m as a conservative bound for path following in environments with dense layout characteristics and adjacent corridors. 

\subsection{Learning a User-specific Dynamics Model}
\label{sec:dynamics}

Model-based RL can be used to quickly compute an optimal policy due to lower sample complexity~\cite{pilco}. We learn a model for approximating $\mathcal{T}$, denoted as ${f_{\theta}}$, which is a function parameterized by a set of weights $\theta$ that are being updated over incoming observations. The model is then used for sampling future trajectories, {\em i.e.}, predicting behavior of a specific blind user, and finding the sequence of actions which maximizes expected future rewards. The goal is to learn a predictive model,
\begin{equation}
{s_{t+1}^H} = {f}_{\theta}(s_t,a_t)
\end{equation}

with a basic understanding of how the blind user will react (a process of interpretation and decision making which is known to be complex~\cite{Turano1998MentalER,guerreiro2017virtual,variabilityohnbar}). To reason over scenarios where the blind user exhibits delayed responses to instructions on the order of multiple seconds, we require a model which can reason over a history of observations. Hence, we employ a state-of-the-art Recurrent Neural Network (RNN) with Long-Short Term Memory (LSTM) units~\cite{Hochreiter:1997:LSM:1246443.1246450,lipton2015critical}. 

\textbf{User State Parameterization.} Efficiently learning a dynamics model for navigation scenarios is difficult with limited data. If a `Left Turn' instruction always results in similar changes to the regression output space, the model can more easily learn a basic understanding of what a left turn is, regardless of the user position, initial heading, or initial velocity direction. We therefore propose to change the output space to reflect the underlying dynamics during a turn motion in navigation, and find that it results in significantly more robust predictions around turns. Specifically, we convert the user's state to velocity values, so that it is invariant to actual user position, and convert these to polar coordinates (the appendix contains additional information).  This modified output space is denoted as $\hat{s}^H$, such that 
\begin{equation}
{\hat{s}_{t+1}^H} = \hat{f}_{\theta}(\hat{s}_t^H, s^O ,a_t)
\end{equation}
When updating the model over incoming observations, the Mean Squared Error (MSE) of the three components in $\hat{s}_{t+1}^H$ is used. Multi-step prediction is achieved by re-applying the dynamics function over previously predicted values and the updated instruction (sampled from the planner).

\textbf{Personalized Dynamics with Weighted Experts.} Adaptation of the dynamics is performed by updating the model parameters over incoming observations. When adapting the dynamics model to a new user, the model should be able to efficiently leverage the data from previously observed users. Hence, adaptation can be formulated as a form of transfer learning~\cite{tlearn}, where a deep network is pre-trained over the source data (in our case, a transitions dataset obtained by previous users of the system) and fine-tuned to the new target data ({\em i.e.}, incoming data from the new user). However, while a single Neural Network model provides a popular high-capacity model choice~\cite{NNdynamics}, it can also have high sample complexity (\emph{i.e.,} when data is limited at the onset of adaptation).

We propose to use a weighted experts approach~\cite{littlestone1994weighted} to adapt the dynamics model in a sample efficient manner, where data from previously observed users is leveraged by using predictions of person-specific models as the experts. In adaptation to a new user, the off-line dataset is clustered to train $M$ experts (person-specific dynamics models). To avoid issues with the experts pool, {\em i.e.}, a small number of similar experts or a very different new user's distribution, we learn a weighted experts model over both previous users' models and a new user model learned from scratch. 

\begin{figure}[t]
\begin{minipage}{0.45\textwidth}
\begin{algorithm}[H]
\caption{Update Weighted Experts Dynamics Model}
    \begin{algorithmic}[1] 
    	\Procedure{UpdateWeights}{$D$, $\theta_W$, $\theta_{M+1}$, $\theta_{1:M}$, $\eta$, $\lambda$}
			\State $\theta_l = [\theta_W, \theta_{M+1}] $
			    \State $\theta = [\theta_l, \theta_{1:M}] $
				\State $E(\theta_l) \gets$ $\mathcal{L}$\textsubscript{MSE}($D$, $\theta$) + $\frac{\lambda}{2}\theta_l^T \theta_l$
				
				\State $\theta_l	\gets \theta_l - \eta \nabla_{\theta_l} E (\theta_l) $		
		\State \Return $\theta \gets [\theta_l, \theta_{1:M}] $
		\EndProcedure
    \end{algorithmic}
    \label{alg:update}
\end{algorithm}
\end{minipage}
\hfill
\begin{minipage}{0.45\textwidth}
\begin{figure}[H]
   \centering
     \includegraphics[width=2.5in,trim={0cm 11.5cm 20cm 0cm},clip]{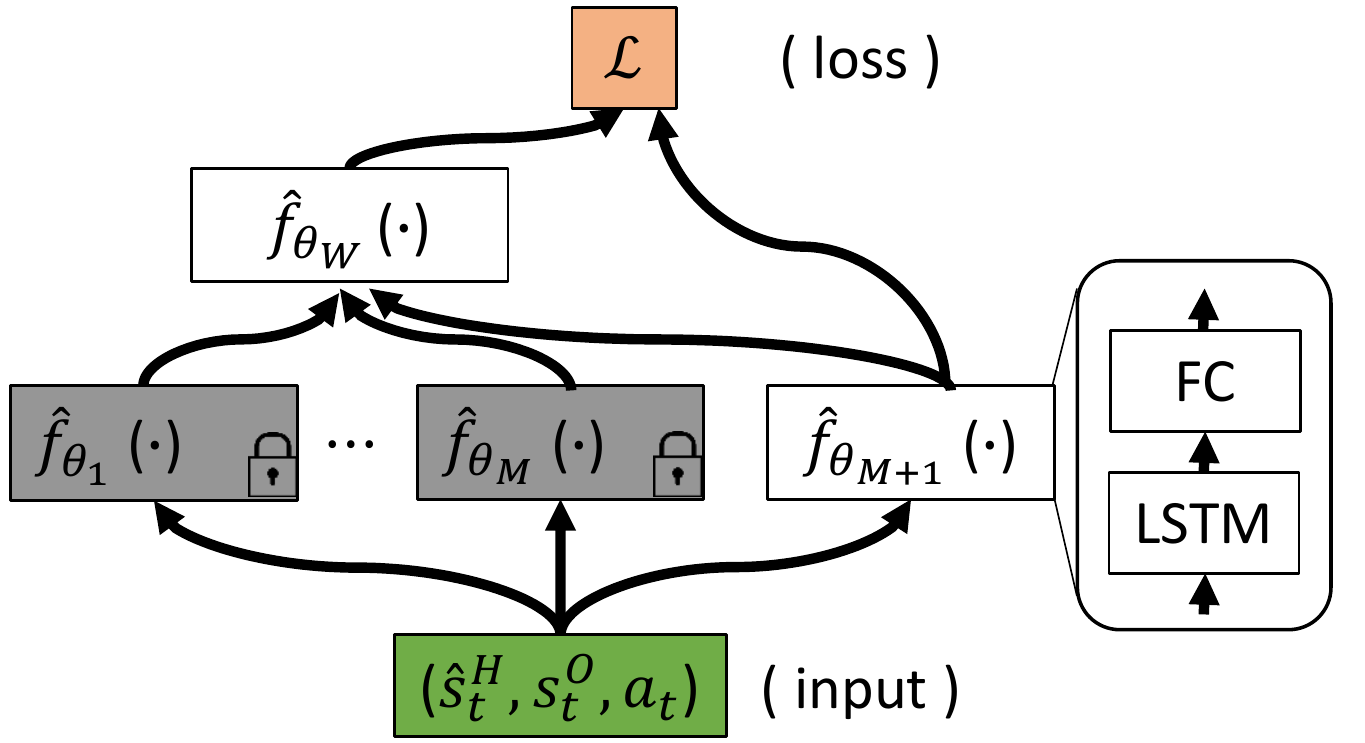}
   \caption{\textbf{Weighted experts model.}}~\label{fig:dynamicsexperts}
   \end{figure}
   \end{minipage}
\end{figure}

Formally, when adapting to a new user, we learn a combination module $\theta_W$ ({\em i.e.}, a fully-connected layer) over the previously learned personal user-specific models' output ($\hat{f}_{\textcolor{blue}{\theta_{i=1:M}}}$) and the current user-specific model ($\hat{f}_{\theta_{M+1}}$)  
\begin{equation}
\hat{f}_{\theta} = \theta_W^T \lbrack \hat{f}_{\textcolor{blue}{\theta_1}},\dots, \hat{f}_{\textcolor{blue}{\theta_M}}, \hat{f}_{\theta_{M+1}} \rbrack
\end{equation}
where we drop the input notation of state and action at the current time for convenience, and note that $\textcolor{blue}{\theta_{i=1:M}}$ are color coded as they are kept frozen during the adaptation phase (only $[\theta_W, \theta_{M+1}]$ are updated, see Fig.~\ref{fig:dynamicsexperts}). In this manner, the model can leverage the well-trained person-specific models while the predictions of the new person-specific model improve, and both modules can complement each other when making predictions. We demonstrate this approach to have both high data efficiency and modeling capacity when adapting the dynamics to a new blind user in real-world settings.

\subsection{Personalized Instructional Guidance Planning with Model Predictive Control}
\label{sec:mpcontrol}

For adaptive instructions with PING, the incremental personalized dynamics model needs to be paired with a planning algorithm. Model-based planning involves making future predictions and using the navigation task's reward function $r$ to select the instructions provided by the interface. Due to the continuous state space, we employ an approximate solution method which is suitable for our navigation application domain. The MPC-based planner~\cite{Lenz2015DeepMPCLD,camacho2013model,6907174} can handle compensating for errors in the model by re-planning at every step and adds minimal learning overhead.   

Given a finite planning horizon ($L$), we aim to find the sequence of actions that maximizes,
\begin{equation}
\label{eqn:mpc}
\begin{split}
    \argmax_{A = (a_{t}, \dots, a_{t+L-1})} \sum_{t'=t}^{t+L-1} r(s_{t'}, p, w),  
    \text{s. t. } \hat{s}_{t'+1}^H = \hat{f}_\theta(\hat{s}_{t'}^H, s_{t'}^O, a_{t'}), s_{t'+1} =  \text{Integrate}(\hat{s}_{t'+1}^H, s_{t'}) 
 \end{split}
\end{equation}
where `Integrate' is an operation to recover the original state parameterization for calculating the reward, $p$ is a globally planned path, and $w$ is a set of waypoints. The discount factor is set to $\gamma = 1$. 

We approximate the solution to Equation~(\ref{eqn:mpc}) by sampling a set of plausible instructional sequences and choose the one which maximizes expected cumulative reward. Specifically, we unroll the trajectory with the current policy to obtain an initial action sequence, and use random sampling and time-shifting to generate additional plausible sequences~\cite{mpcmethods}. This approximate solution to Equation~\ref{eqn:mpc} can be used to select the next action to be executed, observe the resulting state, and re-plan the instructions. The planned path is re-planned in cases where the walker drifts off too much (Alg.~\ref{alg:ping}).

%% file: experiment.tex
\section{Formative Study and Experimental Results}

Robust prediction of the user state is a main challenge in instructional planning with PING. Hence, we begin our work by analyzing a large dataset of real-world user dynamics. Besides uncovering the impact of personalization on the end-user experience, the dataset is used to (1) train person-specific dynamics models as experts, (2) perform on the fly personalization to new users, (3) evaluate long-term prediction as adaptation occurs to compare personalization schemes, and (4) inform the dynamics of a set of diverse simulated walkers for evaluating PING. 

\textbf{Platform.} To understand and model dynamics of real users, we employ a dataset collected with the smartphone app of~\cite{navcog3}. While several other assistive navigation solutions have been proposed, including a service~\cite{Azenkot:2016:EBS:2906831.2906835} or portable  robot~\cite{icrabackpack}, the open-source smartphone app~\cite{navcog3} can be quickly disseminated across researchers and blind users. Using a smartphone also allows users to use their preferred mobility aid while navigating without much difference to their daily mobility. Besides the possibility of reaching wide-spread adoption, we made efforts to keep our formulation as general as possible, so that it may be useful for other applications in assisted navigation. For instance, studying adaptation of the dynamics model is not just relevant to transferring to new users, but also to different use modes, platforms, and mobility aids. 

\textbf{Dataset.} IRB-approved data from 9 blind participants was used (4 female, 5 male, ages 37-70). One participant used a guide-dog and the rest white cane. Following a training session, each user was asked to navigate in a large public space. The {\raise.17ex\hbox{$\scriptstyle\sim$}}$400$ meters route spanned three floors with over 200 Bluetooth beacons placed every 6 to 8 meters. The smartphone's IMU and environment's Bluetooth beacon sensor measurements are fused with a particle filter~\cite{arulampalam2002tutorial} to provide continuous user position and heading information. As in any real-world system, the measurements contain inherent noise. The localization error, evaluated by manual annotation of the position of the users from video, averages 1.8m, which is competitive for indoor settings.  

\textbf{Data Processing.} Each trajectory was resampled to $500$ millisecond intervals. Given over two hours of total data analyzed, this results in {\raise.17ex\hbox{$\scriptstyle\sim$}}$15K$ samples overall. Prediction is fixed at $T=20$ seconds into the future in most of the experiments. When visualizing motion during this time span (Fig.~\ref{fig:fig1final}) we can see how this challenging long time horizon for predicting human behavior takes into account the approaching, task onset, re-orientation, and motion continuation along turns. Also, the interface must be given sufficient time to plan ahead, including announcement time. 

\textbf{Evaluation Measures.} Emphasizing long-term prediction of human behavior, we compute \textbf{end-point displacement error} (\textbf{EPE}), in meters, between the predicted future trajectory and the observed trajectory. We also employ the average displacement error between the two trajectories.

\textbf{Adaptation Experiment.} When adapting to new users, data from a new test user is partitioned into \textit{two temporally disjoint sets}, an adaptation set and a test set. As the model observes the new user's motion from the onset of the experiment at time 0, we collect the transitions, update the model weights, discard the collected data, and repeat. Following each update, we use the dynamics model to perform long-term prediction on all transitions in the test set and average the errors across all sampled points along the trajectory as a summarizing evaluation measure. As the model observes more from the adaptation set, the \textit{test set remains fixed}. Unlike batch error, this evaluation procedure provides a constant measure for the improvement, and therefore a more meaningful way to benchmark the impact of personalization schemes while ensuring generalization across environments. We update the weights in batches of 30 second intervals to provide some learning stability while identifying clear performance trends, yet this update rule can be modified as needed in practice.

%% file: analysis.tex
\begin{figure}[!t]
   \centering
   \begin{tabular}{ccccc}
   \multicolumn{4}{c}{\textbf{User Motion Statistics - Large Turns}} & \\
   \includegraphics[width=1.2in]{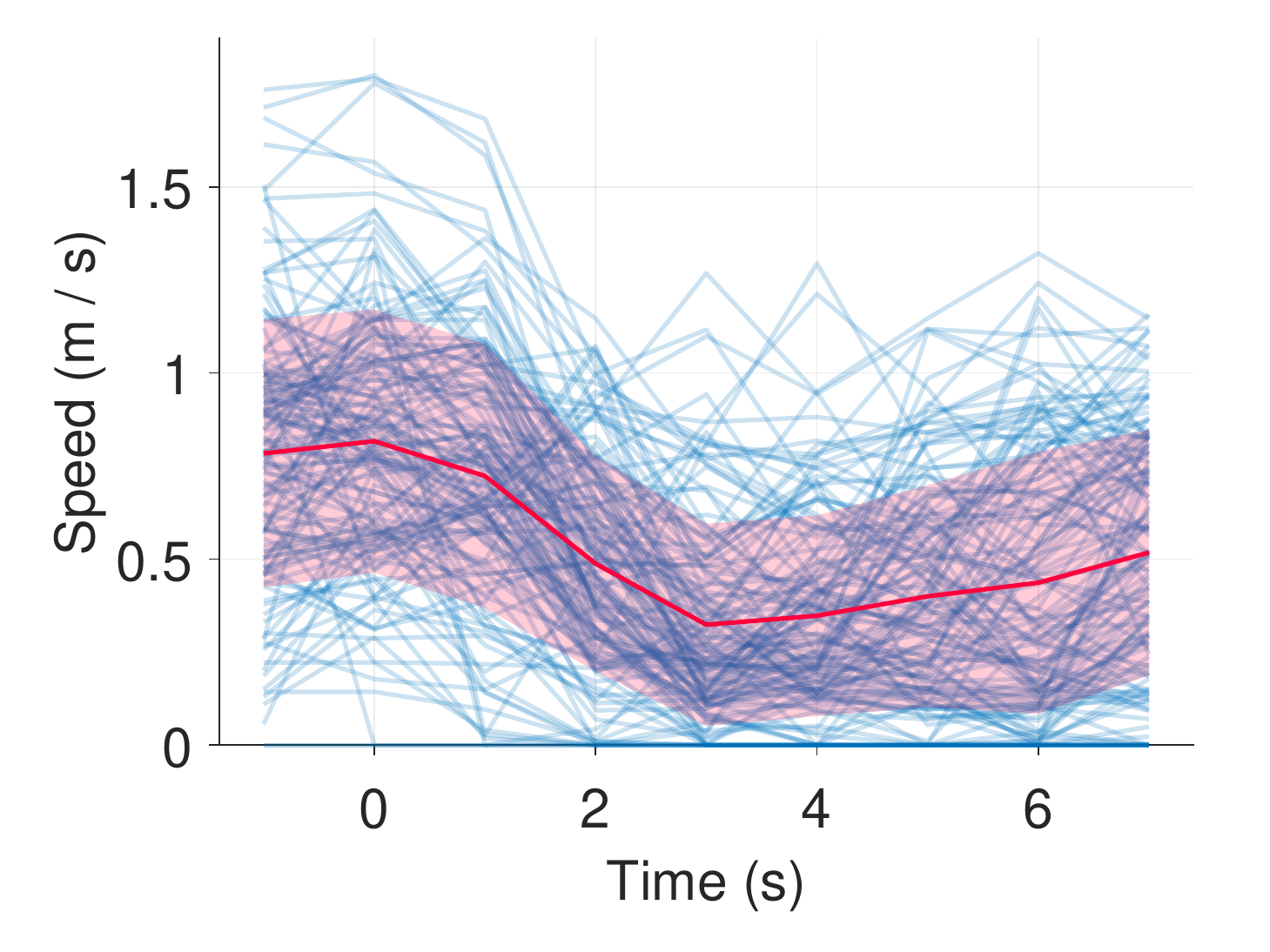}
   &
   \includegraphics[width=1.2in]{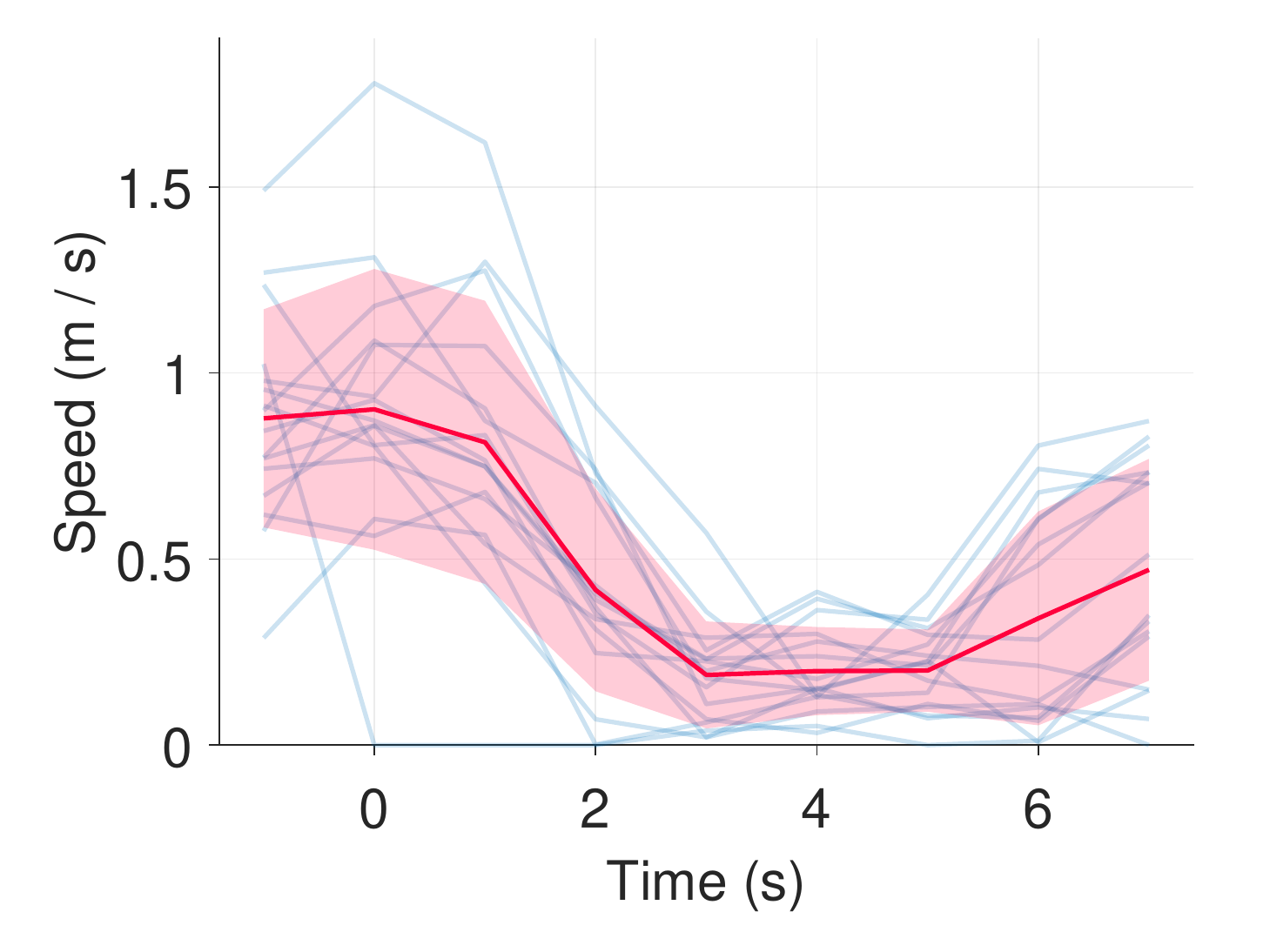} & 
\includegraphics[width=1.2in]{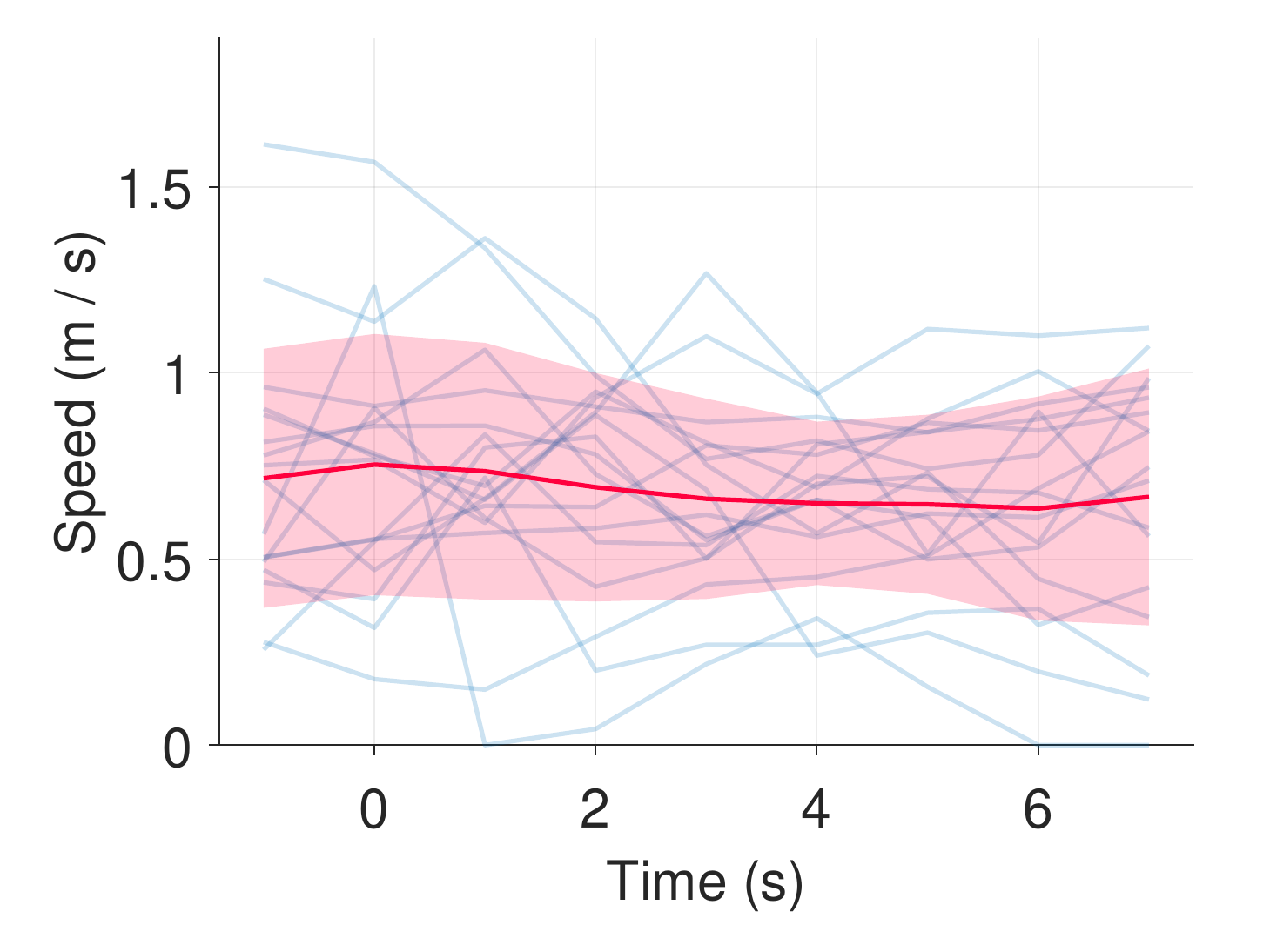}  
&
\includegraphics[width=1.0in,trim={0cm 0 2.8cm 0},clip]{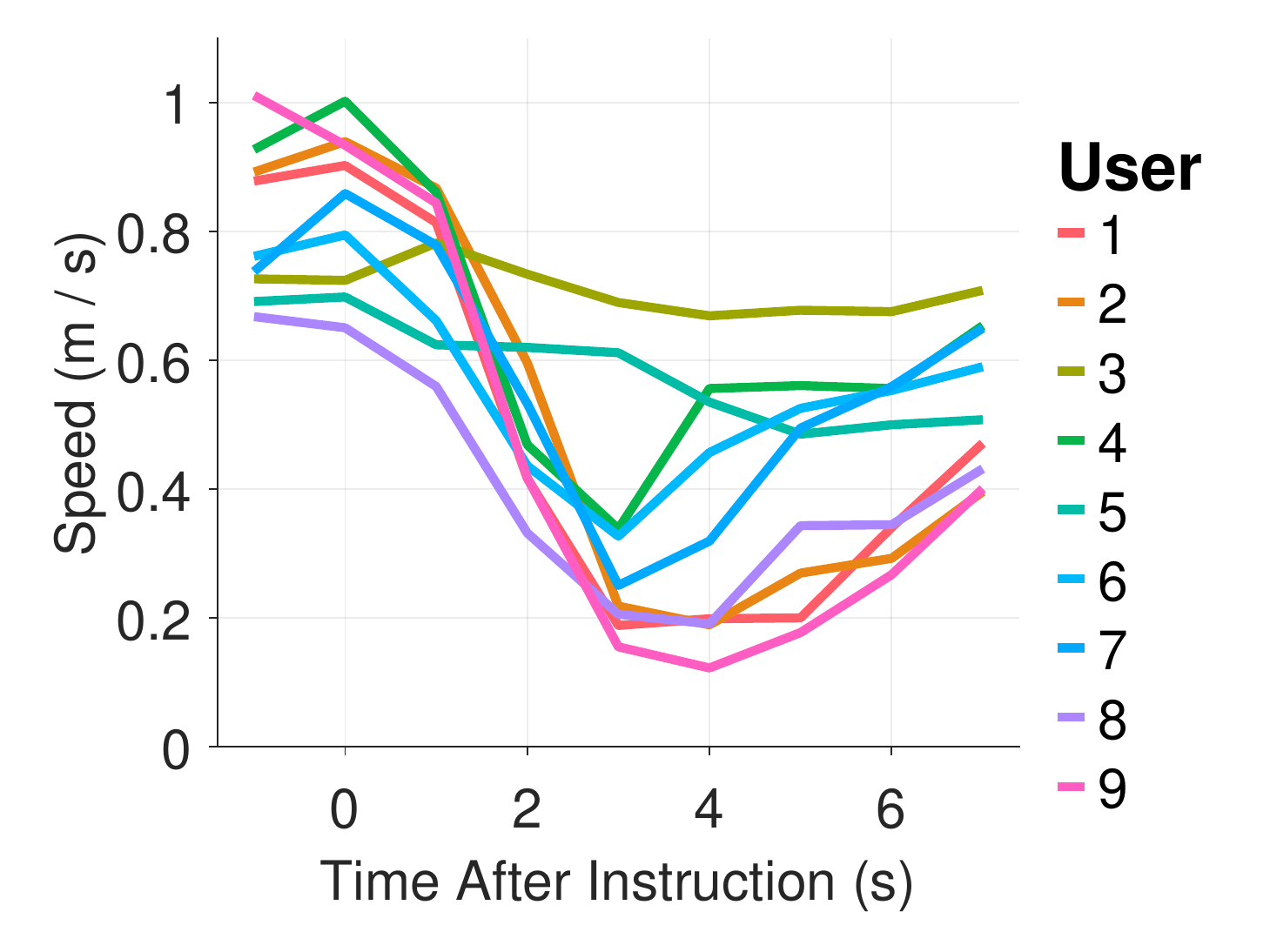} 
&
    \includegraphics[width=0.17in,trim={0cm 2cm 13cm 0},clip]{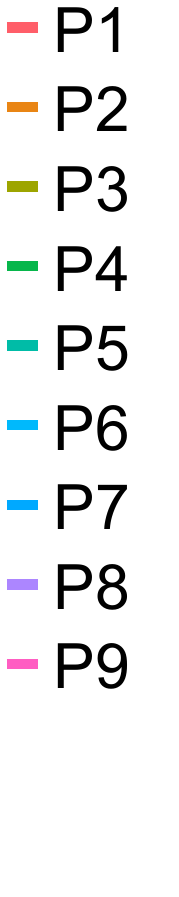}
     \\
(a) All Users & (b) User 1 & (c) User 3  & (d) User Averages & 
     \end{tabular}
   \caption{\textbf{Individual differences among users when responding to turn instructions.} (a-c) Overall speed statistics in the dataset (mean and standard deviation overlaid), where time 0 is the onset of a `turn' instruction (data shown for 90$^{\circ}$ turns) and the y-axis is the speed. (d) shows averages for all the users in the data.}~\label{fig:turnssupp}
   \vspace{-0.3in}
   \end{figure}
   
\subsection{Formative Study: Characterizing the Dynamics of Blind Users}
To motivate our adaptive approach, we performed statistical analysis of user behavior which can be found in the appendix, as well as Fig~\ref{fig:turnssupp}. Analyzed by leveraging continuous tracking of the users, the extent of this phenomenon has not been rigorously studied in related studies. With the aim of understating the role of personalization, we find that in long real-world navigation routes with diverse stimuli, user variability has significant implications to system design. Specifically, beyond individual walking pace, we find user reaction (\emph{e.g.,} speed change, task timing) varies in a person-specific manner, in particular during the most crucial elements of the navigation such as when turning and encountering obstacles. For instance, users with similar initial speeds may respond very differently to the same instruction, \emph{e.g.}, one slows down to a near stop during turns and another maintaining pace, so that a constant velocity or user-agnostic dynamics model will perform poorly in long-term prediction. We also observe user-specific consistency in under/over-turning and behavior around obstacles. We found such individual differences in the dynamics to result in navigation errors in open spaces and turns, motivating an adaptive interface which can predict ahead and adjust accordingly. 

\subsection{Evaluation of Learned User-specific Dynamics Model}
\begin{figure}[!t]
\centering
\begin{tabular}{cccc}
\multicolumn{2}{c}{\textbf{Entire Dataset}} & \textbf{Turn Events} & \\
       \includegraphics[trim=.5cm 0cm .5cm 0cm,clip,width=1.4in]{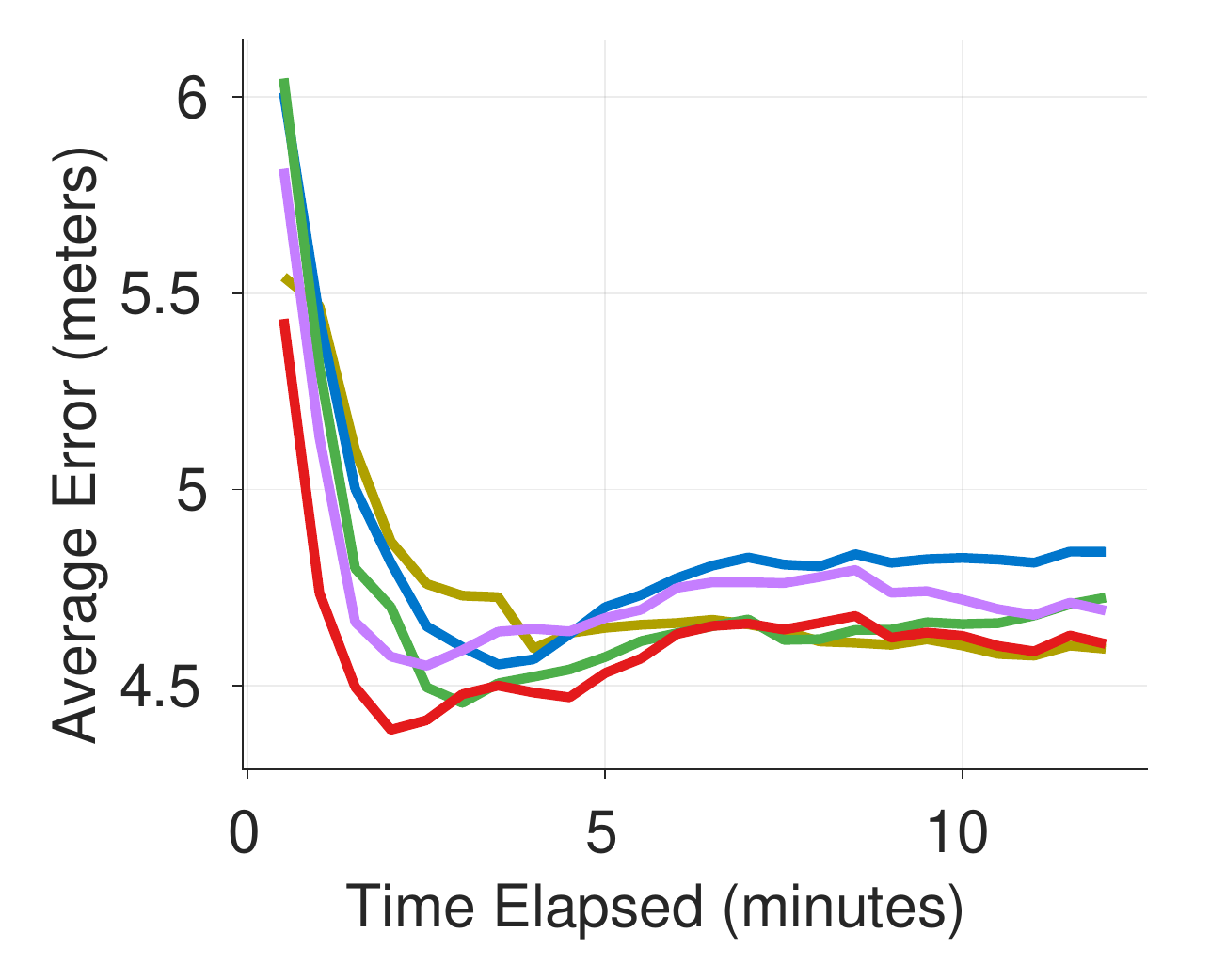}&
    \includegraphics[trim=.5cm 0cm .5cm 0cm,clip,width=1.4in]{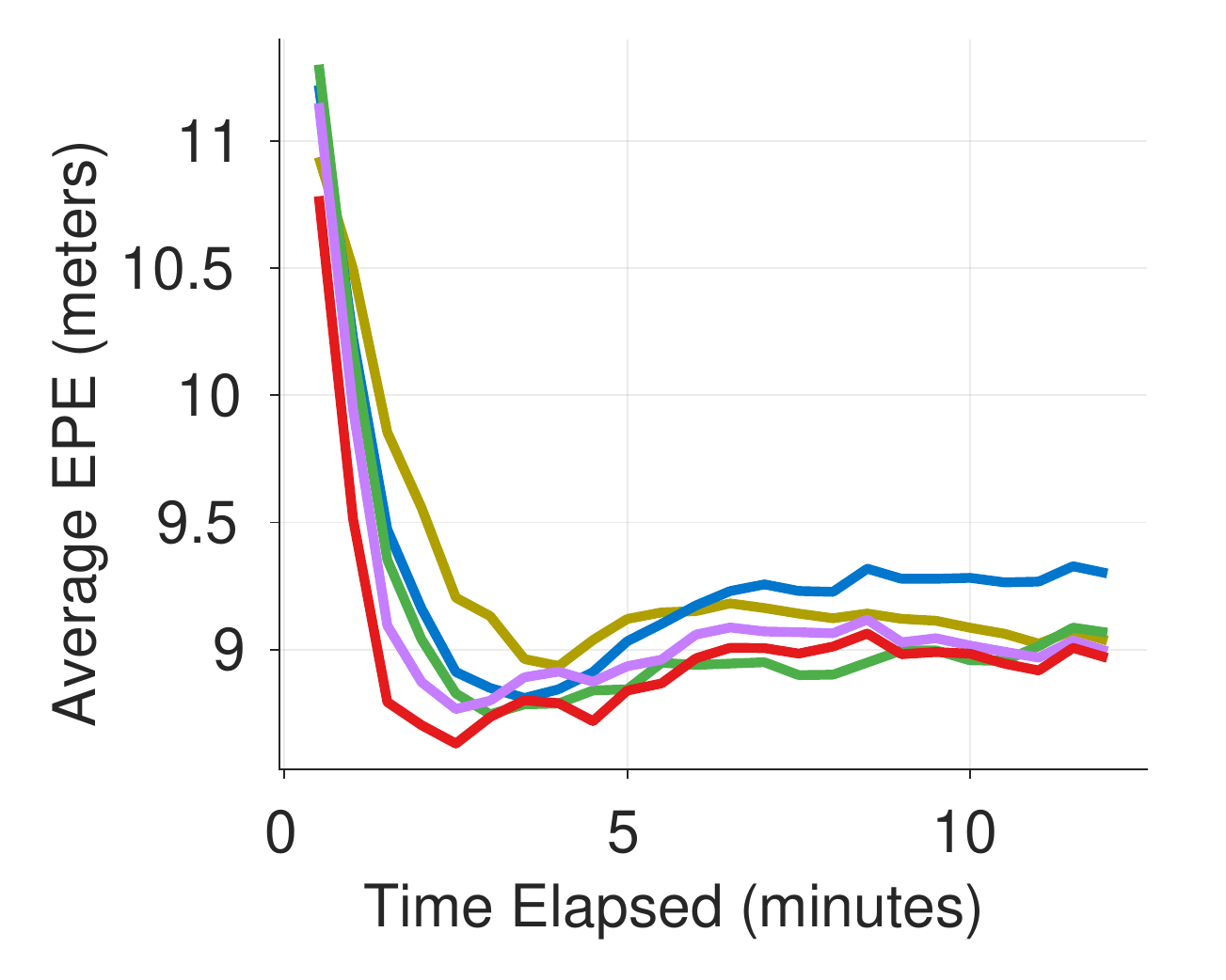}&
    \includegraphics[trim=.5cm 0cm .5cm 0cm,clip,width=1.4in]{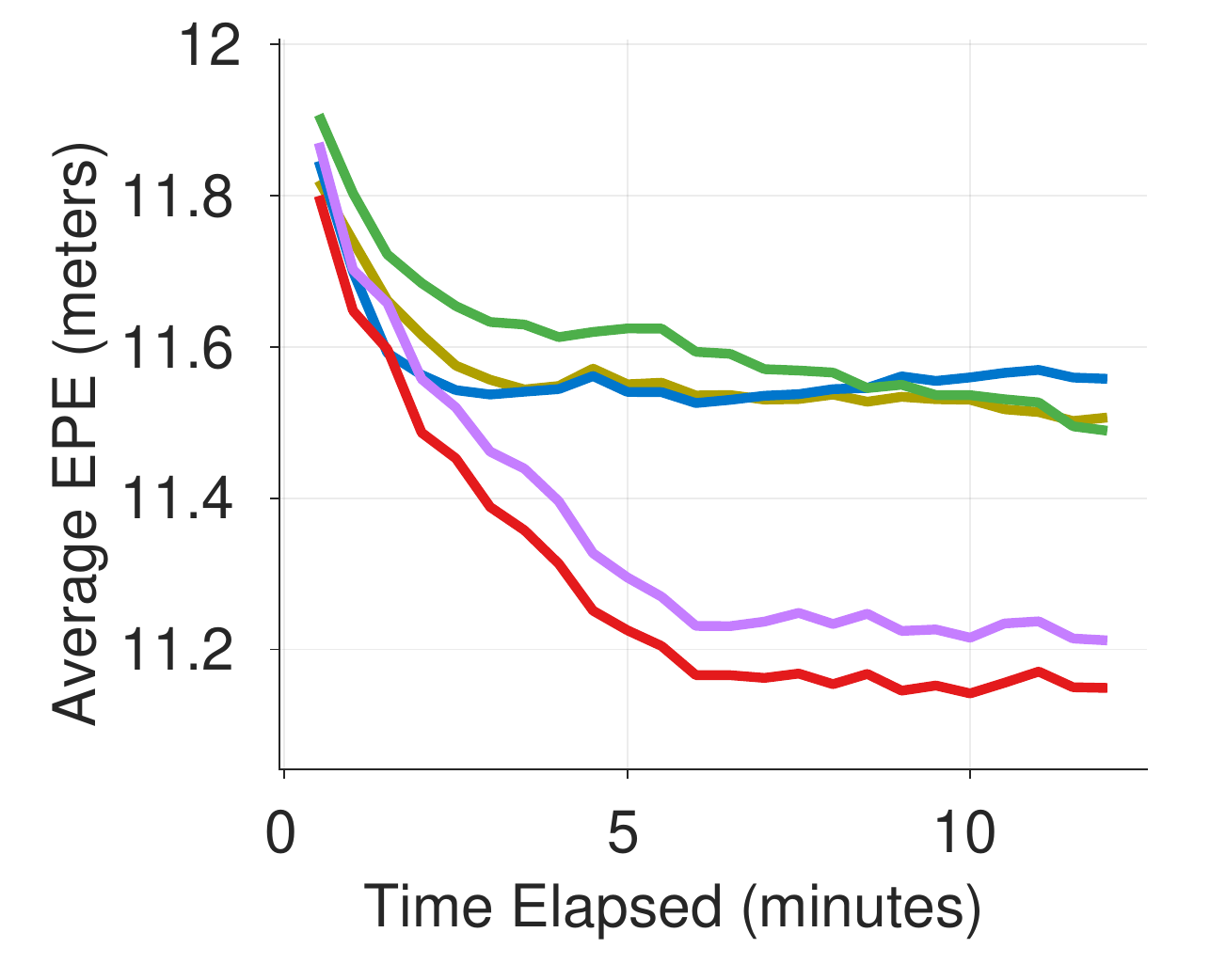} &
        \raisebox{-0.25cm}{\includegraphics[trim=15cm 0cm 0cm 3cm,clip,width=0.9in]{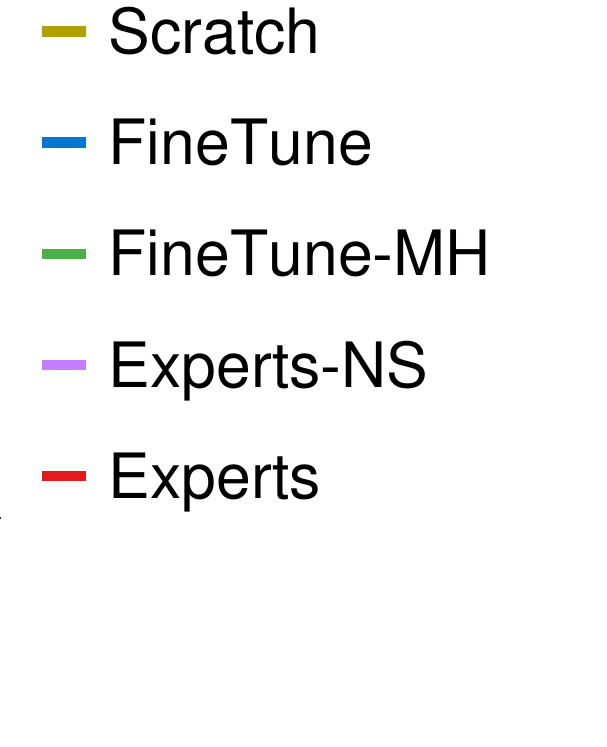}}
       \end{tabular}
          \caption{\textbf{Adaptation to new users.} The learned dynamics models are evaluated as more data becomes available for personalization (starting the onset of the experiment, time 0 on the X-axis). Results are shown for a 20 second prediction error.}~\label{fig:allaexy}
          \vspace{-0.3in}
       \end{figure}

Based on our analysis of blind user dynamics, we expect personalization to be essential in predicting future user behavior. As shown in Fig.~\ref{fig:allaexy}, several adaptation schemes are benchmarked to understand issues in personalizing the dynamics model. A person-specific RNN baseline involves learning a single model (\emph{i.e.,} $\hat{f}_{\theta_{M+1}}$) from \textbf{scratch}. We also analyze the de-facto standard in transfer learning of \textbf{fine-tuning} a pre-trained model. In the experiments, we consider two variants of fine-tuning. In one variant, initial pre-training is done in a person-agnostic manner ({\em i.e.}, a model learned with all previous users' data) and weights are updated for the new user's motion dynamics. In the second variant, the pre-training is done with multiple network heads~\cite{osband2016deep,li2017learning} (a separate head for each previous user). These approaches can learn to leverage shared motion patterns across different users and are the most widespread approaches for transfer learning. However, relying on the fine-tuning mechanism to learn and adapt a general hidden representation can be prone to optimization issues, in particular when data is limited at the onset of adaptation.

Fig.~\ref{fig:allaexy} depicts the results for the adaptation experiments, where a large reduction in long-term prediction error across all the personalization techniques can be observed. Due to the inherent variability, we find all models perform similarly at the onset of the experiment as it begins after an initial batch of new user's data. For this challenging prediction task, we can see a significant reduction in prediction errors once the entire adaptation set has been observed by the models. For instance, adaptation translates to a 2.23m ($19.7\%$) reduction in 20 second end-point prediction error for the fine-tuning adaptation baseline that is pre-trained with multiple separate network heads (\textbf{FineTune-MH}), 1.92m ($17.1\%$) reduction for the standard fine-tuning baseline (\textbf{FineTune}), and 1.81m ($16.8\%$) for the experts model. For a complete analysis, we also show results of a weighted experts model without the jointly trained new user-specific dynamics model as one of the experts (\textbf{Experts-NS}). Overall, the success of the improved long-term user behavior prediction is directly due to user-specific consistency in reactions, as the models are required to generalize to disjoint part of the navigation in our evaluation procedure.

\begin{figure}[!t]
   \centering
   \begin{tabular}{ccc}
  \multicolumn{3}{c}{\includegraphics[trim=0cm 18cm 13cm 0cm,clip,width=4in]{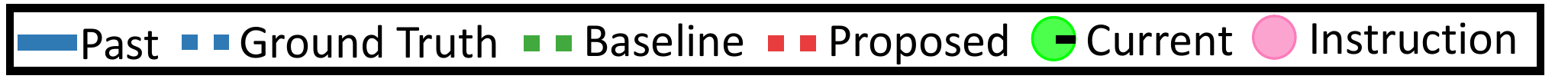}}
  \\
   \includegraphics[width=1.7in]{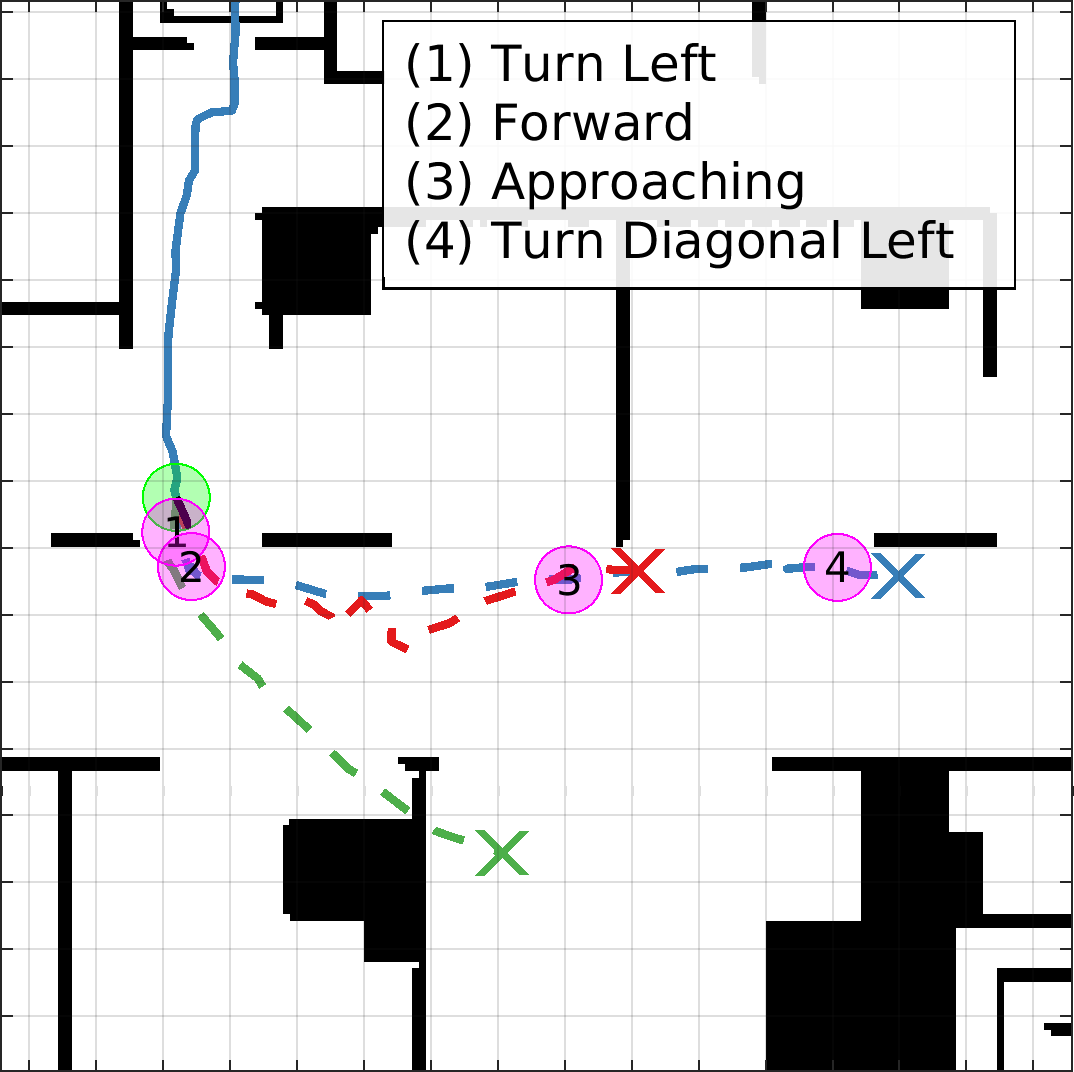} &
   \includegraphics[width=1.7in]{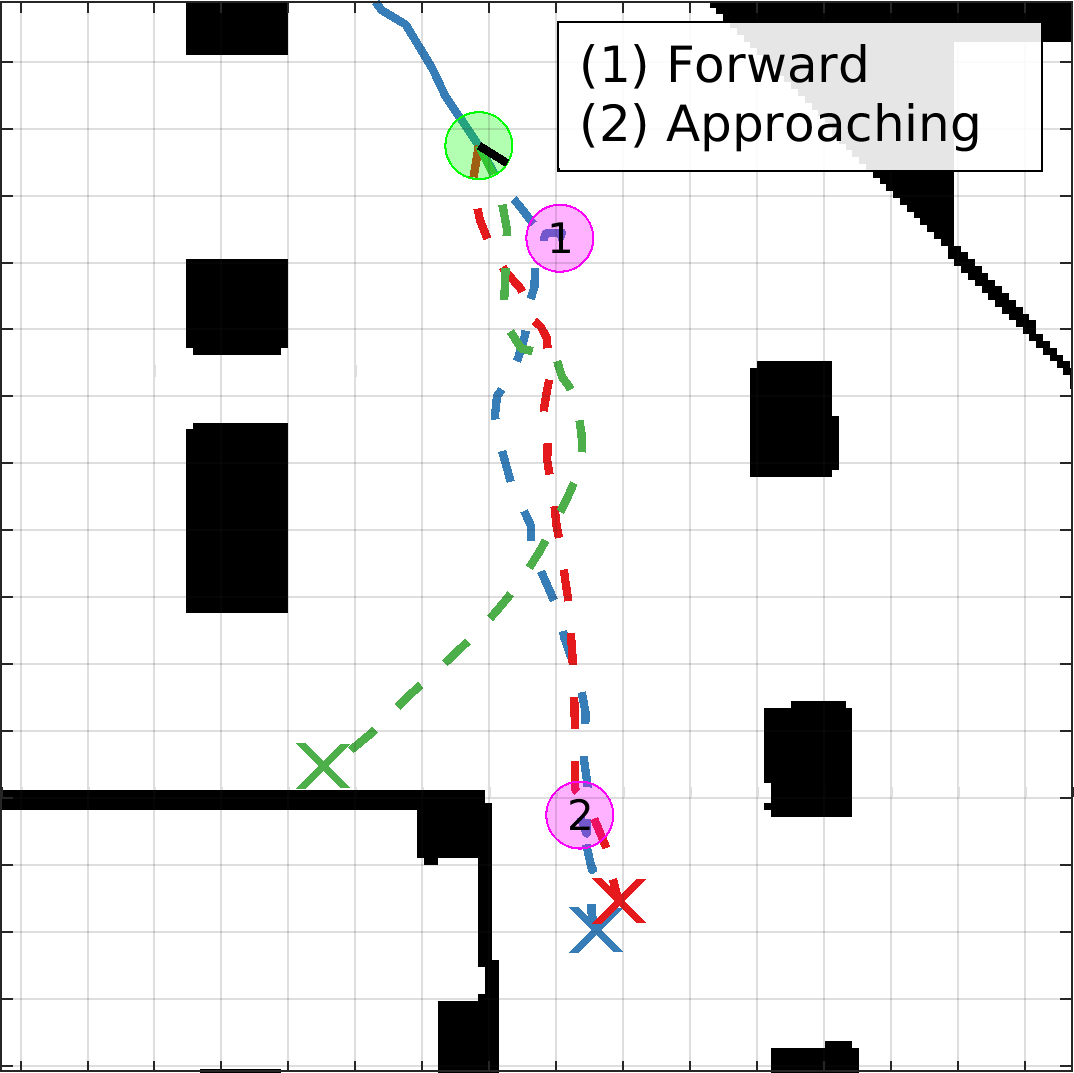}  &
    \includegraphics[width=1.7in]{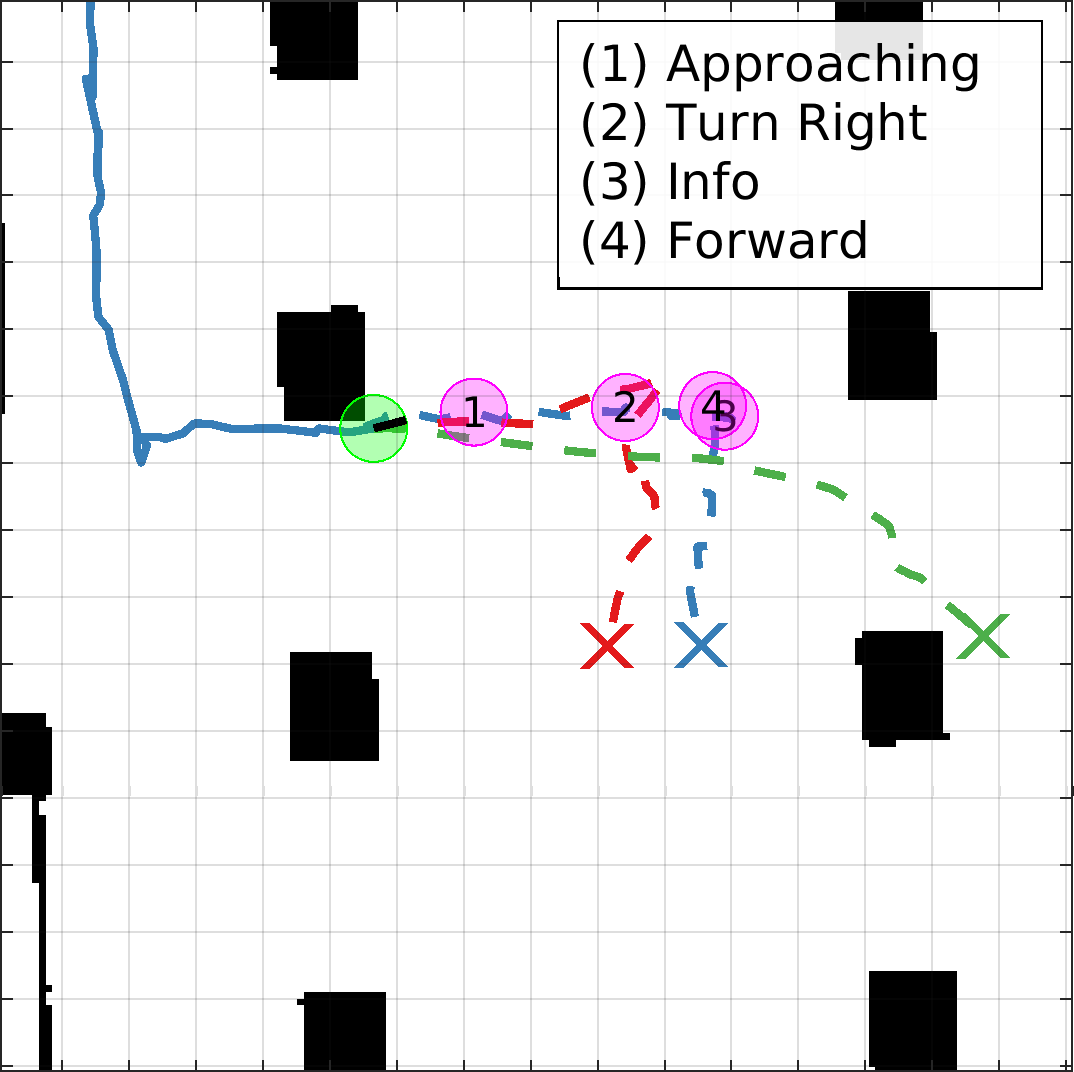} \\
    (a) & (b) & (c)
    \\
      \includegraphics[width=1.7in]{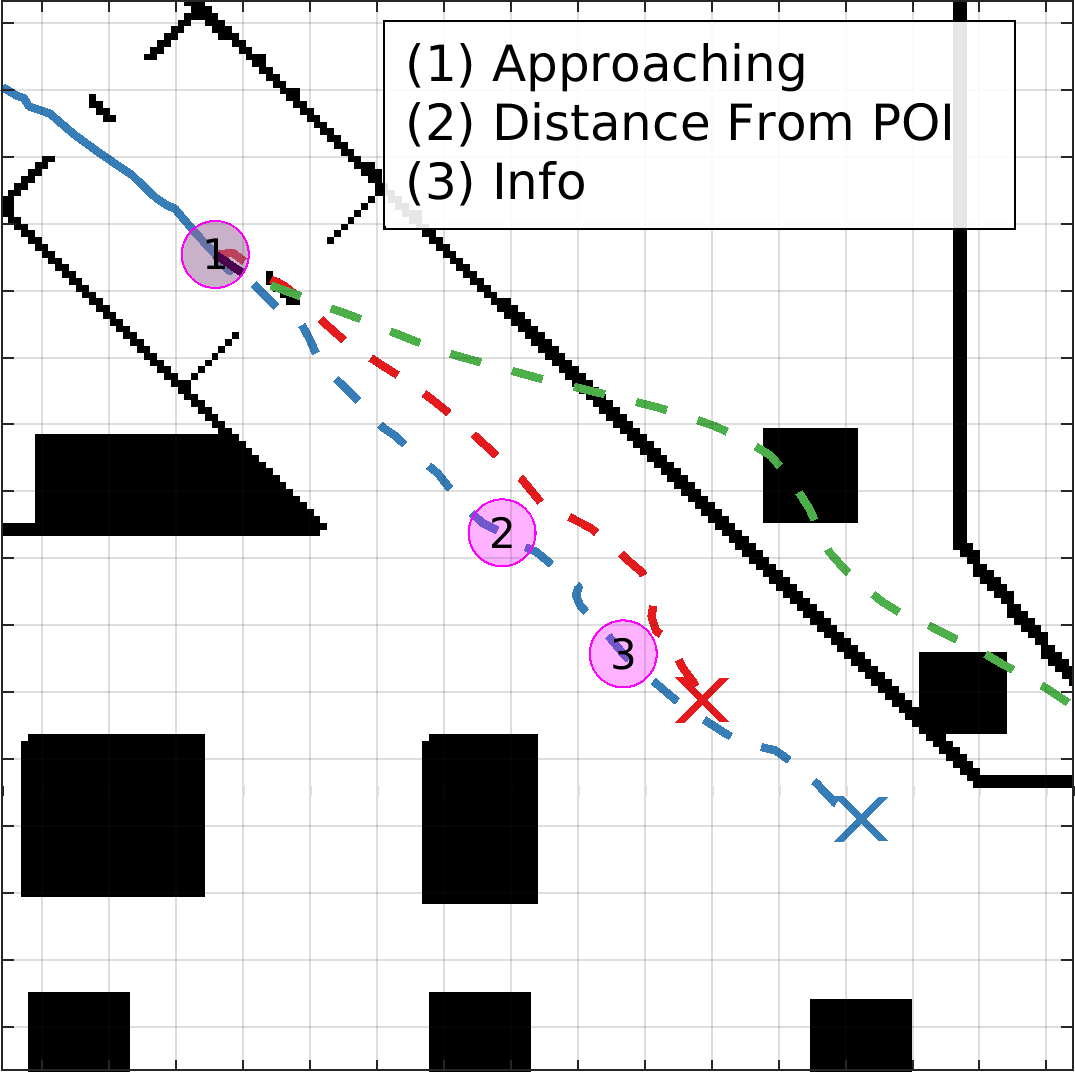} &
   \includegraphics[width=1.7in]{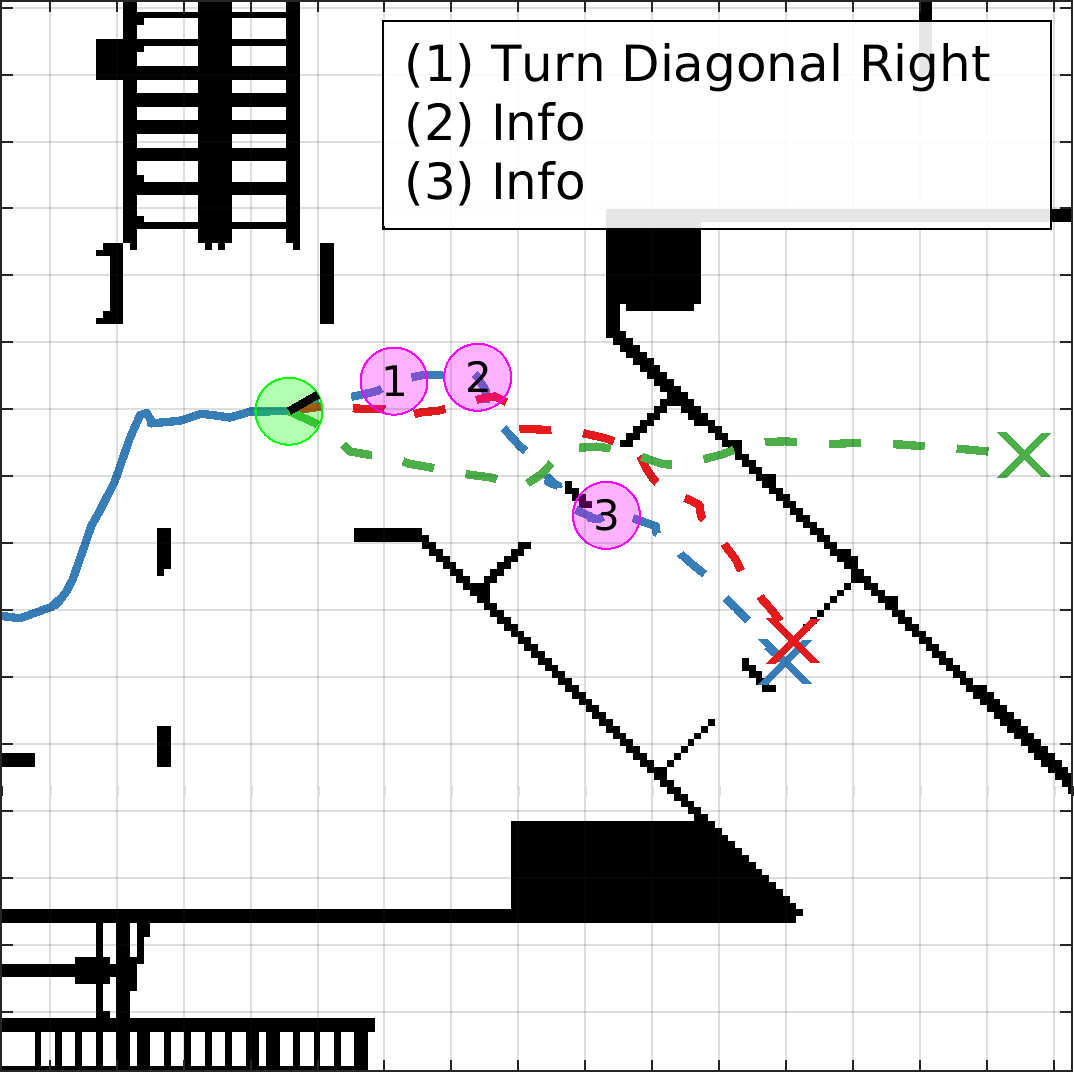}  &
    \includegraphics[width=1.7in]{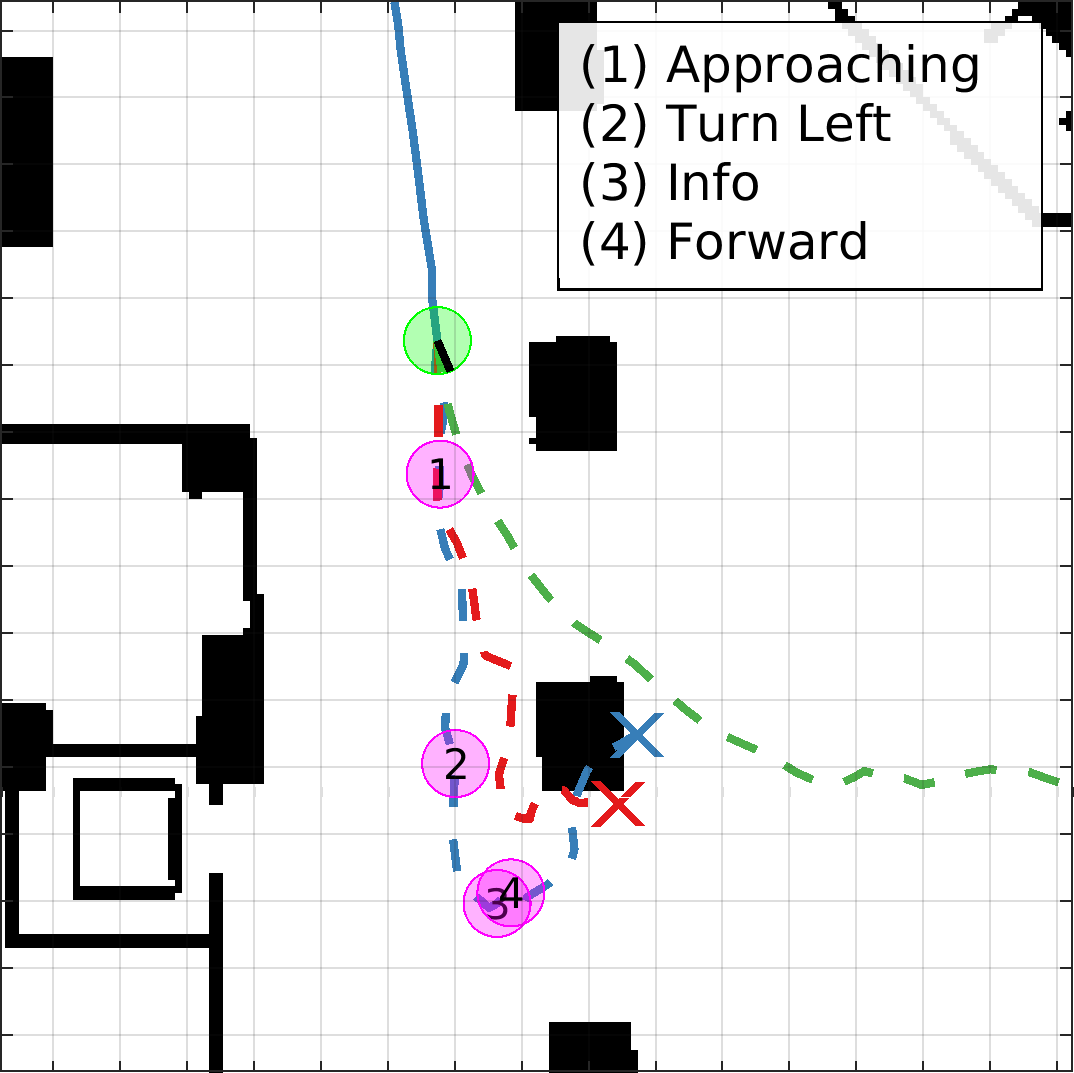} \\
    (d) & (e) & (f)
    \end{tabular}
   \caption{\textbf{Visualization of long-term prediction results.} The weighted experts personalization scheme is compared with a fine-tuning adaptation baseline. Instruction points from the policy are only shown along the ground truth future trajectory. }~\label{fig:expers}
 \end{figure}

When contrasting to the model that is trained from scratch in Fig.~\ref{fig:allaexy}, we can see how both fine-tuning baselines are able to leverage previously observed user data. Multi-head training is found to improve learning of a generalizable representation which can be better transferred across different users. However, despite ablative experiments with model size and learning rates, the model shows slower adaptation performance compared to the experts model (which is shown to achieve the same performance rates in half the time or less), and only minimal improvement during turns compared to the model that is trained from scratch. While turn events are rare, they result in the most inter-user variability. This performance issue is a significant limitation as navigation errors occur most frequently during turns in the dataset. In this case, we find that the model is unable to fully account for personal variability, and instead results in a more generic or averaged motion model. This can be seen when visualizing the predicted trajectories in Fig.~\ref{fig:expers} (the appendix contains more examples). Specifically, the model incorrectly estimates user's reaction and fails to account for individual veering behavior (Fig.~\ref{fig:expers}(c)). Although adaptation through fine-tuning a shared-weights model has many benefits (\emph{e.g.,} computational efficiency), it comes with a trade-off when adapting to large changes in the input-output variable relationship when observing a new user.

The experts model shows the best performance in terms of data-efficiency, often converging in under 2 minutes with minimal `over-fitting' to the user behavior thereafter (\emph{i.e.,} the observed phenomenon where additionally observed data increases testing error). By leveraging outputs from a set of specialized dynamics models, this adaptation method can quickly identify which previously observed models most match the current user's dynamics. This adaptation process is therefore found to be more efficient when compared to single model fine-tuning. In terms of prediction results, we find that unlike the other adaptation baselines this architecture is able to balance both short and long-term prediction accuracy across different types of instructional events and environmental scenarios, as can be seen by inspecting the $T=5$ and $T=10$ seconds prediction plots in the appendix. 

   \begin{figure}[!t]
   \centering
   \begin{tabular}{ccc}
      \includegraphics[width=1.7in]{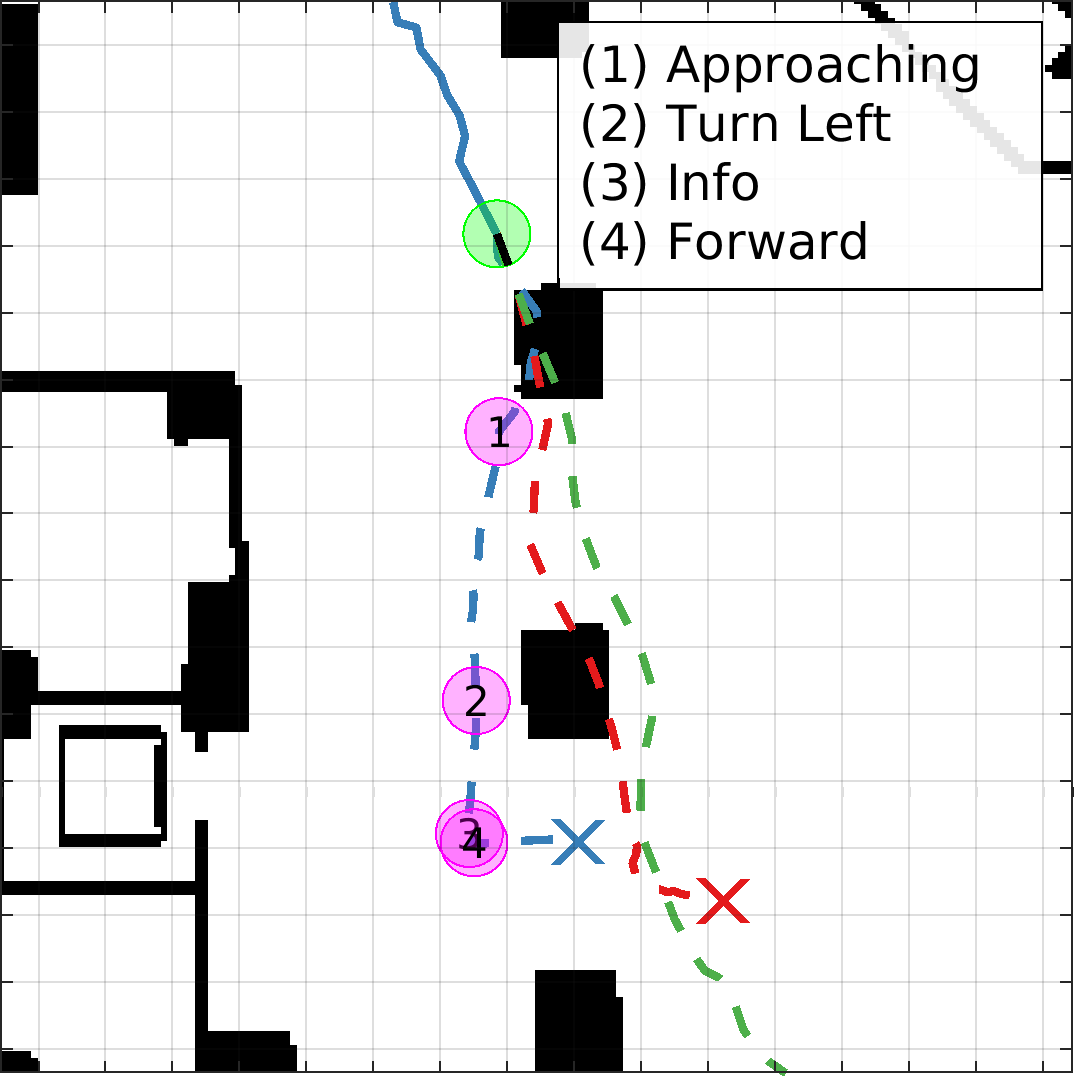} &
   \includegraphics[width=1.7in]{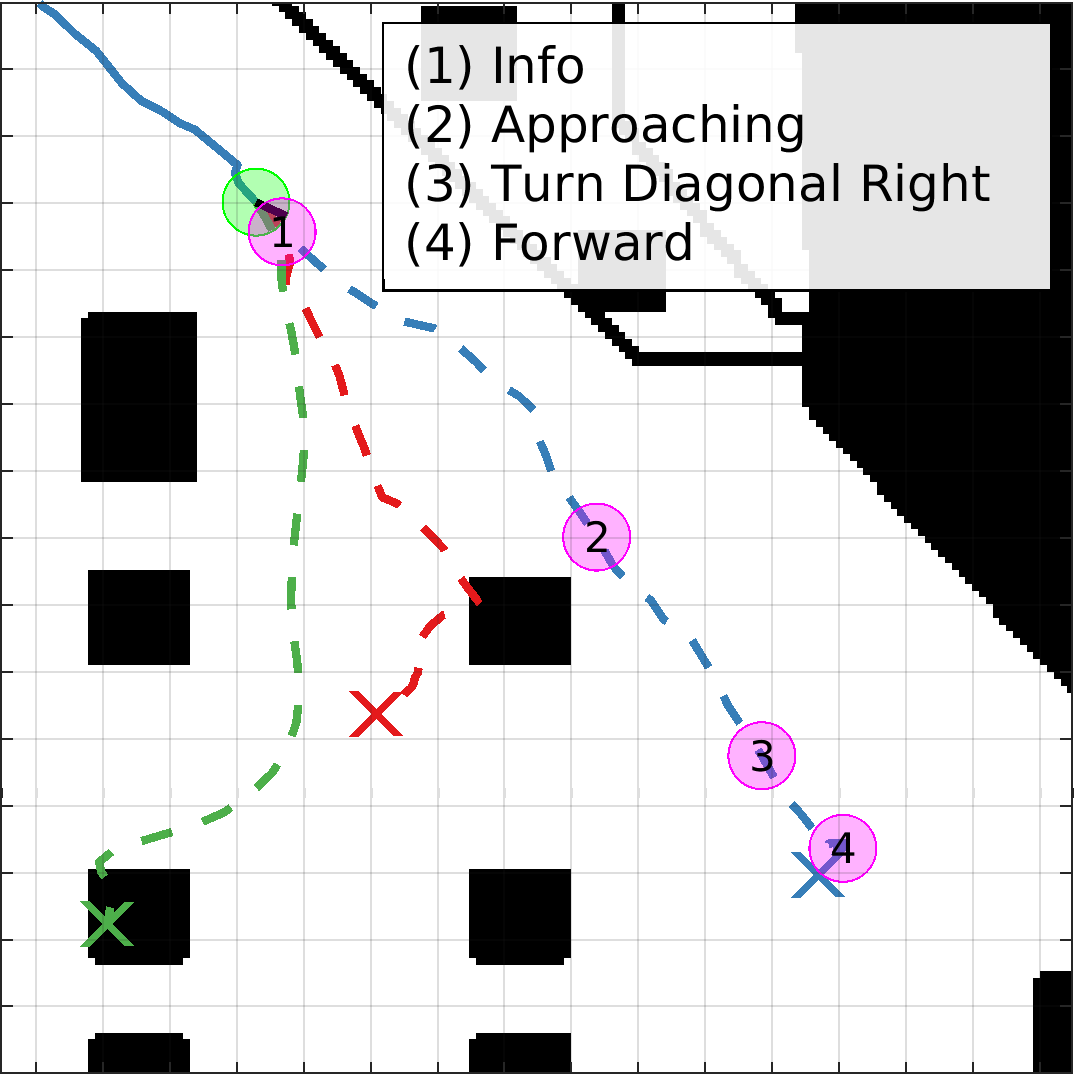}  &
    \includegraphics[width=1.7in]{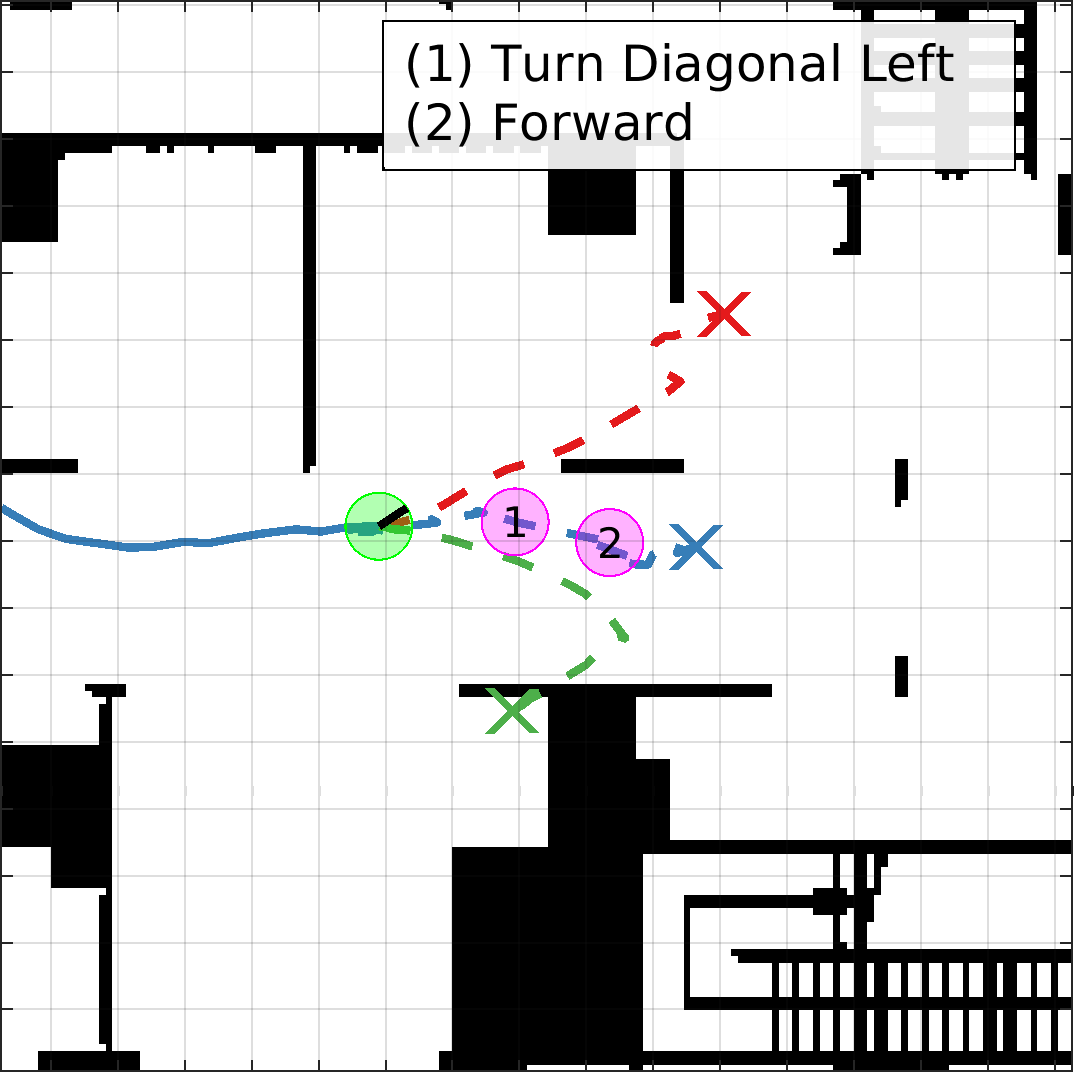}
   \\
   (a) & (b) & (c)
    \end{tabular}
   \caption{\textbf{Challenging cases.} Even under challenging cases, such as around pillars and significant veering, the proposed approach results in more reasonable predictions by better capturing long-term motion patterns of the specific user. }~\label{fig:failasupp}
 \end{figure}

Isolating turn events in the evaluation (Fig.~\ref{fig:allaexy}-right) demonstrates that the proposed experts model outperforms other baselines at both adaptation (lower errors at the onset of learning) and convergence. Since the performance is averaged over thousands of events, the results are statistically significant ($p<.001$). The results are also statistically significant up to minute 3 over the entire dataset Fig.~\ref{fig:allaexy}-left ($p<.01$). These findings suggest that employing the proposed model could mean the difference between a successful turn or a early/late turning and consequently slower arrival to destination. While we expect the model to eventually prefer the new user's personal model (hence matching performance with the user-specific model that is trained from scratch), we find that the model learns to generalize better through the majority voting, even when ample data is available.



We note that at times, predictions may pass through walls or obstacles. Despite the high accuracy indoor navigation system used in the study, there is still an inherent localization error which leads to such instances in the training data. Since observed transitions sometimes go through walls (especially thin walls or wall corners), the trained models also predict such behavior. Nonetheless, we find that the models do learn a notion of interactivity between obstacles and blind users. For instance, certain mobility strategies by users such as tracing along poles and walls can be observed (\emph{e.g.}, Fig.~\ref{fig:expers}(f)). 

Fig.~\ref{fig:failasupp} shows challenging prediction scenarios, where the prediction model overestimates a veering or incorrectly predicts interaction of the user with obstacles. Although minimal veering can lead to large differences over the prediction horizon, the proposed model still produces plausible predictions. For instance, in Fig.~\ref{fig:failasupp}(c) the model considers the orientation of the current user to assume veering to the left, which is reasonable. Also, the response to a `turn diagonally to the left' instruction and the interaction with the wall are a reasonable scenario.

\begin{figure}[!t]
         \centering
         \includegraphics[trim=0cm 0cm 6cm 0.1cm,clip,width=2.4in]{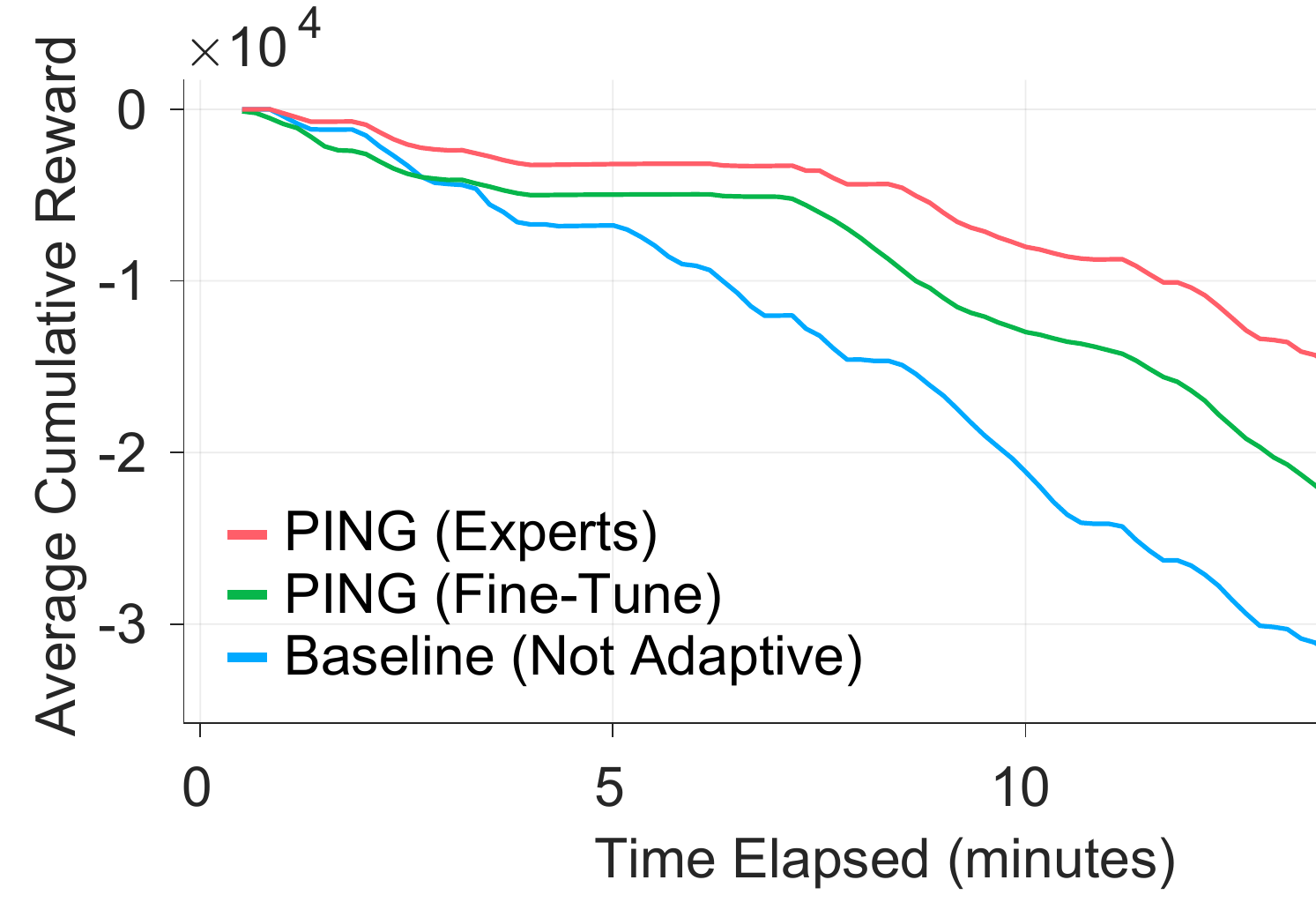}
        \caption{\textbf{Navigation success results with PING.} }~\label{fig:mpcres}
         \vspace{-0.3in}
    \end{figure}

\subsection{Evaluation of Personalized Instruction Generation} 
 
Towards evaluating a complete adaptive system on diverse walkers, we create simulated walkers in the same floor layout as the real-world studies. The simulated users are informed by the statistics of the real-world study, and used to create a larger variety of walking types in the same settings. Hence, we can test PING extensively before the final deployment in a safe manner. We emphasize appropriate timing of instructions during turns, so that walking parameters are randomized over reaction time for instructions, speed change during turns, and amount of veering from the planned path. All experiments begin with 30 seconds of data collection with the static baseline policy of~\cite{navcog3}. The results, averaged over 100 random trails in Fig.~\ref{fig:mpcres}, show that using the experts model for adapting the system on walkers with greater diversity translates into longer periods of the user staying on the path and timely arrival to the destination. Specifically, we observe a $49\%$ reduction in instances of large deviation (over 1.5m) compared to the non-adaptive baseline.

%% file: appendix.tex
\appendix
 \clearpage
  \section{Appendix}
 \subsection{Model Training Details}
Hyper-parameter settings were determined for each model type by a grid search over layer size $\in \{10,25,50,100,200\}$, depth $\in \{1,2,3,4\}$, and learning rate $\in \{10^{0},10^{-1},\dots,10^{-5}\}$. All training objectives used the MSE between model's prediction and observed (reparameterized) transitions. For a person-specific model, a single layer LSTM with 50 hidden units was used, followed by a fully connected layer. The person-agnostic fine-tuning baseline model (learned over data from all users) was found to benefit from increased model capacity, hence we used two stacked LSTM layers of 100 hidden units each. All models were trained with Stochastic Gradient Descent (SGD) with momentum.

\subsection{Reparameterization for Efficient State Regression}

The parameterization of the state space can significantly impact the training of a predictor, in particular when data is limited. If a `left turn' instruction results in similar dynamic changes, the model can more easily learn a basic understanding of what a left turn is, regardless of the user position or initial direction. For instance, instead of regressing future orientation and position values of the user which can result in overfitting or poor convergence with limited data, we may regress velocity values, {\em i.e.}, the difference in $s^H$, $\dot{s}^H = [\dot{x}, \dot{y}, \dot{\alpha}] = [\frac{\Delta x}{\Delta t},\frac{\Delta y}{\Delta t},\frac{\Delta \alpha}{\Delta t}]$ for more meaningful and general learning of user behavior across environments. Yet, we still encountered issues in predicting such state differences, especially around the critical non-linear components of the navigation. 

We found it useful to modify the state space to better accommodate learning user behavior during such events by converting all parameters in $\dot{s}^H$ to polar coordinates. The idea is that a `turn left' action would result in a similar signal to be regressed, regardless of the initial direction of the user. While similar turn types ({\em e.g.}, `turn left') result in similar values of $\dot{\alpha}$ (in radians per second), this is not the case for $[\dot{x}, \dot{y}]$ throughout a turn, which depend on the initial velocity direction and magnitude. This implies that learning the mapping from actions to changes in the state is more challenging, {\em e.g.}, the dynamics during a 90-degree turn are more complex. In our experiments, this resulted in models which poorly behaved around turns. 

For instance, in Fig.~1 in the main paper, the changes in the x and y positions over an entire turn will change depending on the starting heading of the user. Changing this heading will result in very different velocity dynamics values in these components. To simplify training, we convert $[\dot{x}, \dot{y}]$ to polar coordinates, $[\rho,\beta]$ (linear velocity magnitude and orientation). We emphasize the difference between $\dot{\alpha}$, which is the user's heading, and $\beta$, which is the direction of the user velocity vector (the two are independent). Specifically, the user may rotate in place (changing $\dot{\alpha}$ while $\beta$ is 0), or step to the right (changing $\beta$, but not $\dot{\alpha}$). 

$\beta$ encodes a motion direction, and therefore will have different values in the same type of turn. To ensure that the model learns a basic understanding of turn types (without having to show it examples at all initial user motion directions), we predict the change in $\beta$ instead, $\dot{\beta}$, so that \begin{equation}
\hat{s}^H = [\rho,\dot{\beta}, \dot{\alpha}]
\end{equation}
With this parameterization the coordinates are invariant to the original directionality of both heading and velocity, but only vary with the turn amount and turn direction. This straightforward modification was found to impact the learning performance of the regressor during turns.

\subsection{Additional Model Evaluation}
\begin{figure}[!t]
   \centering
   \begin{tabular}{cc}
    \includegraphics[trim=4.3cm 10cm 5.2cm 10cm,clip,width=2.5in]{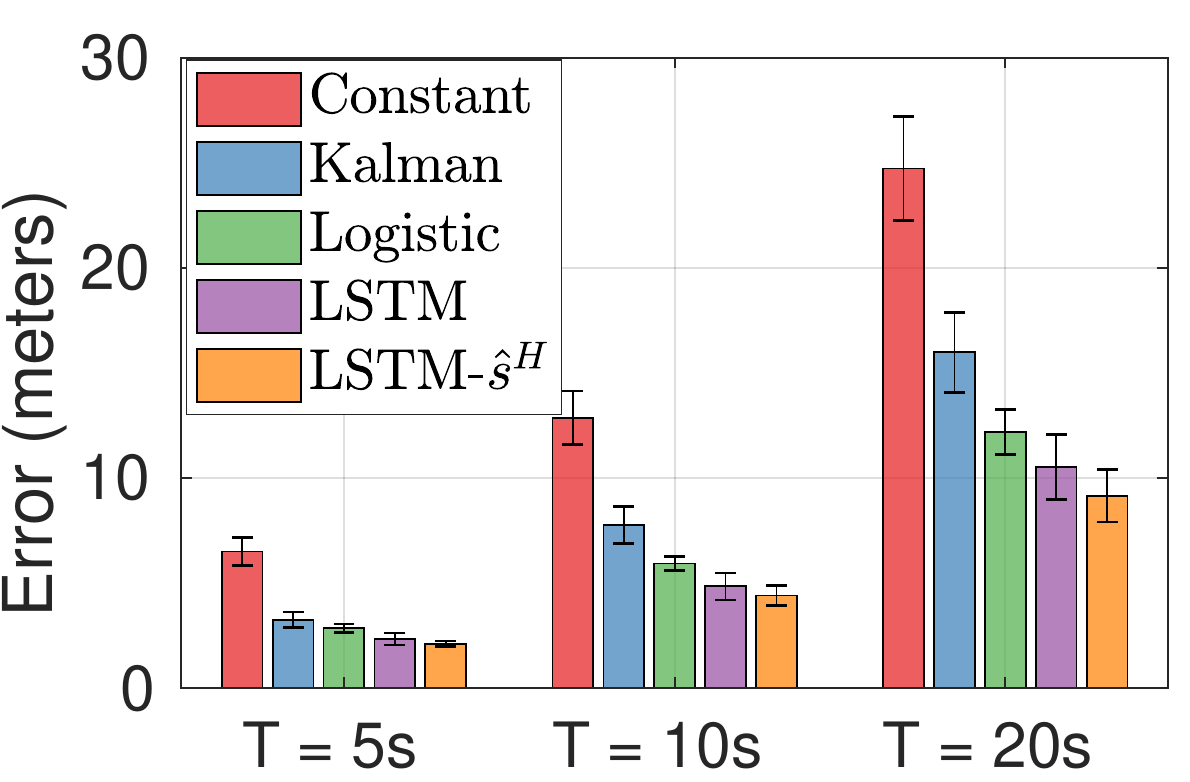}
    &
  \includegraphics[trim=4.3cm 10cm 5.2cm 10cm,clip,width=2.5in]{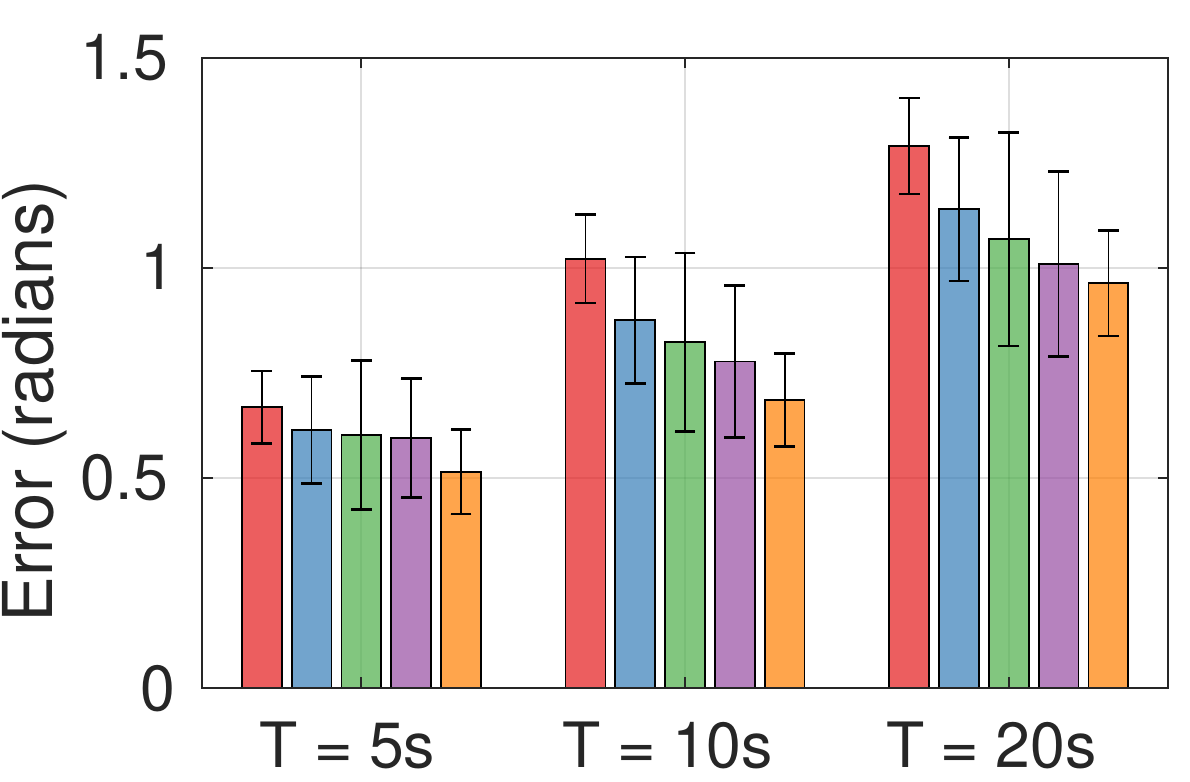}
  \\
      (a) Average end-point error 
      &
  (b) Average end-point orientation error
     \end{tabular}
   \caption{\textbf{improvement due to the proposed reparameterization.} With each model, we predict 5, 10, and 20 seconds into the future, and average the end-point displacement and orientation errors over all instances. All models, besides LSTM$-\hat{s}^H$, regress velocity values, \emph{i.e.,} difference in state instead of raw state. }~\label{fig:dynamicseval}
 \end{figure}

We preform ablative evaluation to benchmark different approaches in prediction over the entire dataset. Training and evaluation is performed across all users to determine impact of different model choices on general motion modeling. By comparing different prediction approaches, including a simple constant velocity model, Kalman filter~\cite{kalman1960new}, Logistic regression~\cite{freedman2009statistical}, we can better understand the impact of our model choice on the end-user's experience. Fig.~\ref{fig:dynamicseval} shows the benchmarking of different model choices for user motion prediction. The models were \textbf{all trained to predict velocity measurements}, as commonly done~\cite{NNdynamics}, besides LSTM-$\hat{s}^H$. 

The analysis shows discriminative training to be useful for handling of highly non-linear trajectory components, where users may reach full stops for multiple seconds or encounter obstacles and veer in different directions. We notice how reparameterization of the regression output space ($\hat{s}^H$ instead of $\dot{s}^H$~\cite{NNdynamics} for the other baselines) is beneficial for better prediction, both in end-point prediction of the trajectory-error drops significantly from 10.5 to 9.1 meters on average end-point prediction at $T=20$ seconds prediction, as well as orientation prediction. The difficulty in learning a rotational and initial-state invariance from the data automatically, in particular without samples at all initial orientations, can be easily alleviated with an appropriate parameterization of the output space.

\subsection{Prediction Evaluation}
\begin{figure}[!t]
   \centering
   \begin{tabular}{ccccc}
 \multicolumn{5}{c}{\fbox{\includegraphics[trim=0cm 0cm 0cm 0cm,clip,width=4in]{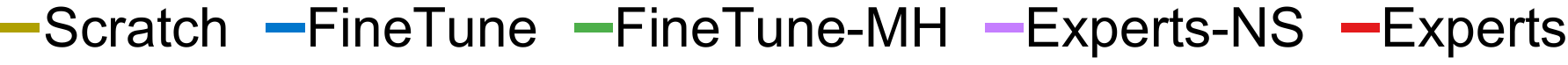}}} \\
 \\
 \multirow{1}{*}{\rotatebox[origin=c]{90}{\parbox[c]{3.5cm}{\centering Prediction T=20s}}}  & \textbf{All Events} & \textbf{Turn Events} & \textbf{EPE - All Events} & \textbf{EPE - Turn Events} \\
     & 
     \includegraphics[trim=0cm 0cm 0cm 0cm,clip,width=1.1in]{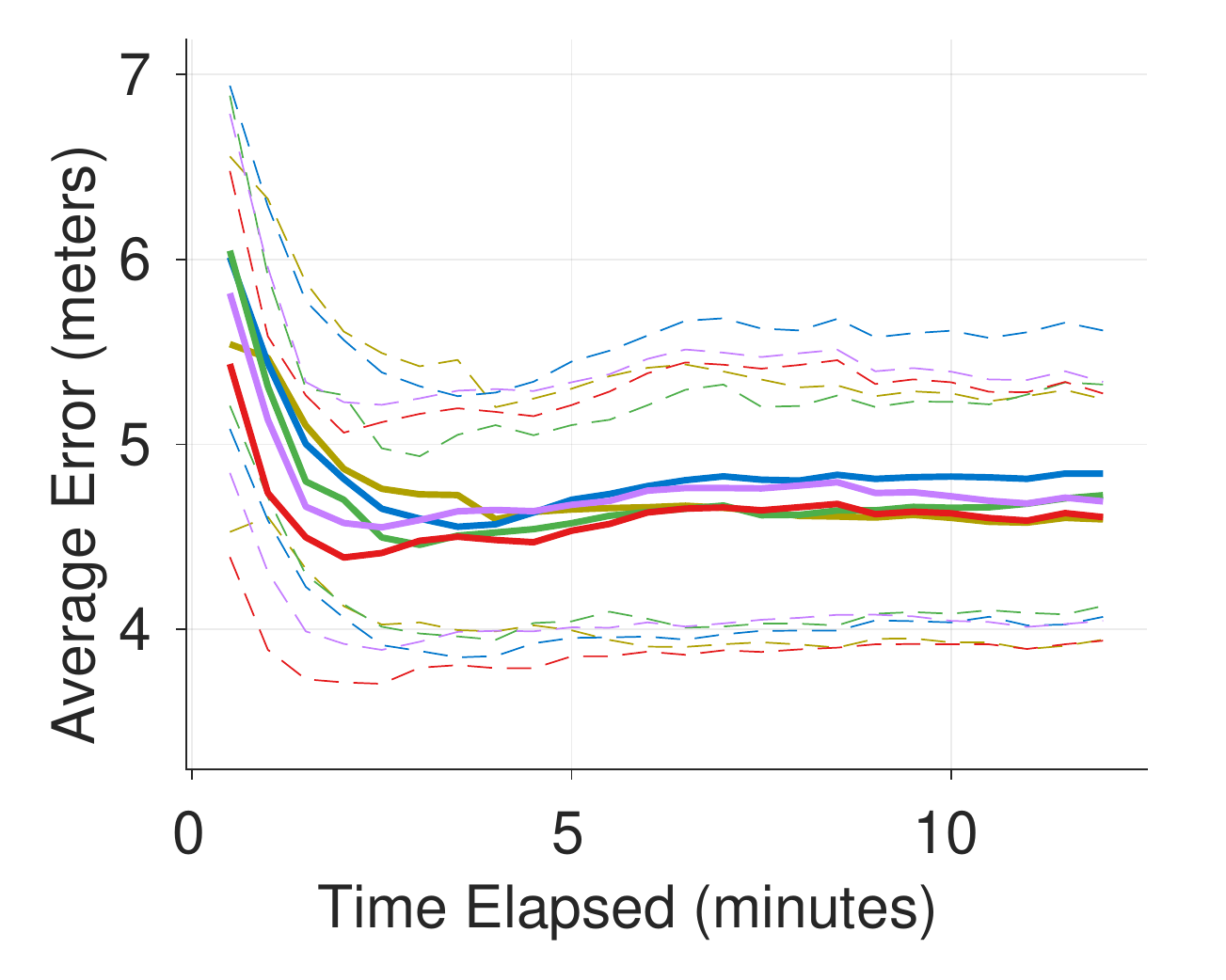}&
     \includegraphics[trim=0cm 0cm 0cm 0cm,clip,width=1.1in]{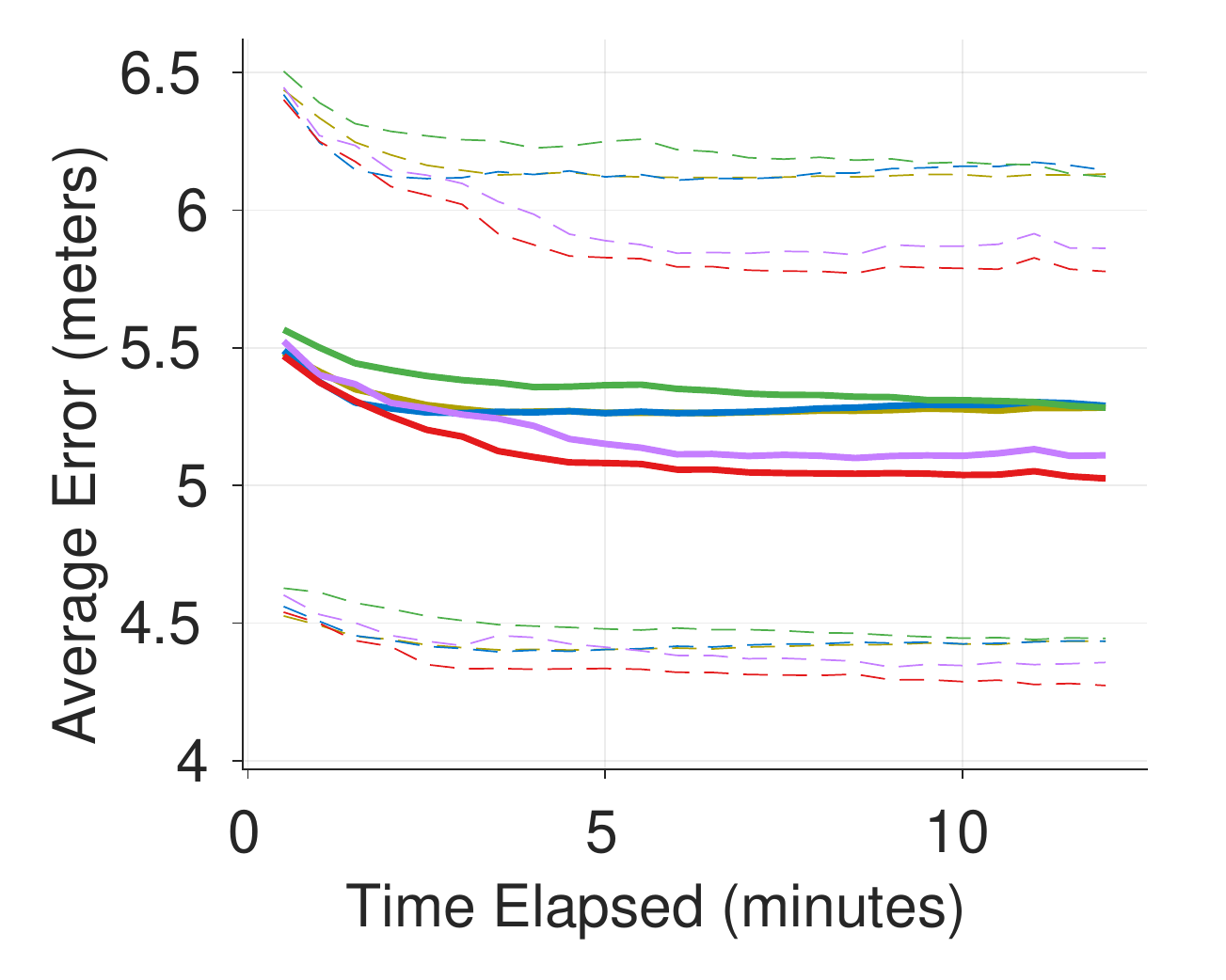}&
     \includegraphics[trim=0cm 0cm 0cm 0cm,clip,width=1.1in]{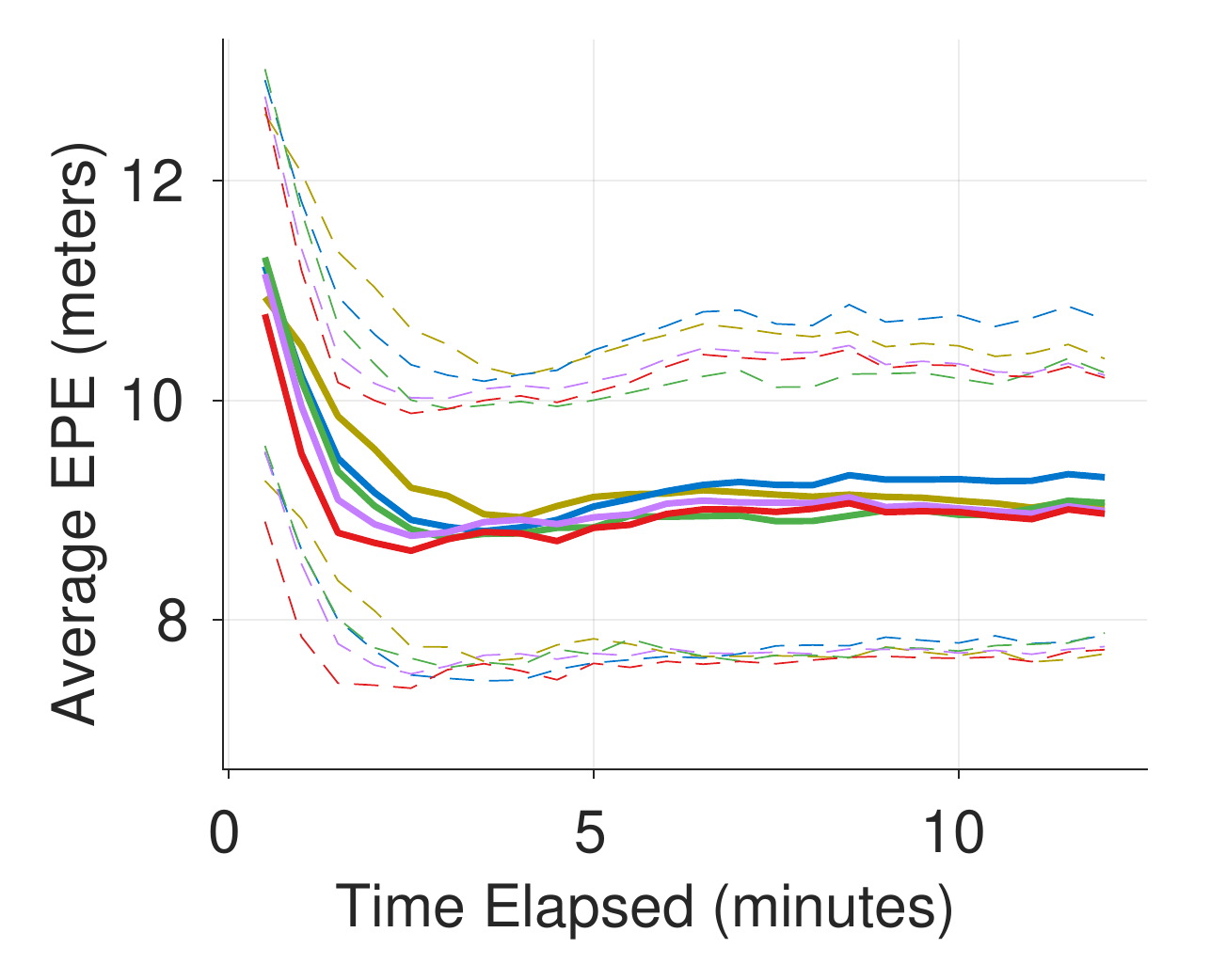}&
     \includegraphics[trim=0cm 0cm 0cm 0cm,clip,width=1.1in]{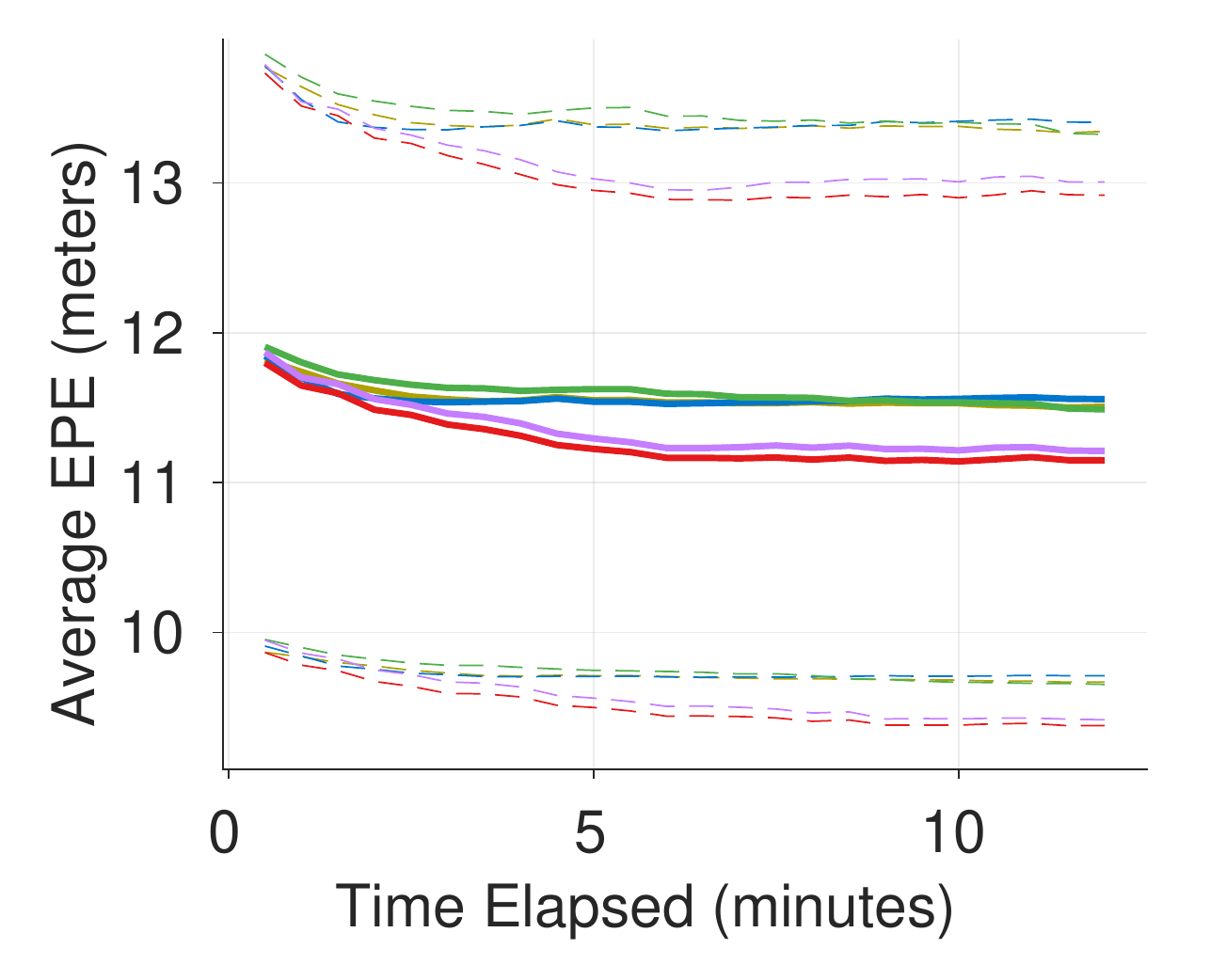}\\
\multirow{1}{*}{\rotatebox[origin=c]{90}{\parbox[c]{3.5cm}{\centering Prediction T=10s}}} & (a) & (b) & (c) & (d) \\
      &
          \includegraphics[trim=0cm 0cm 0cm 0cm,clip,width=1.1in]{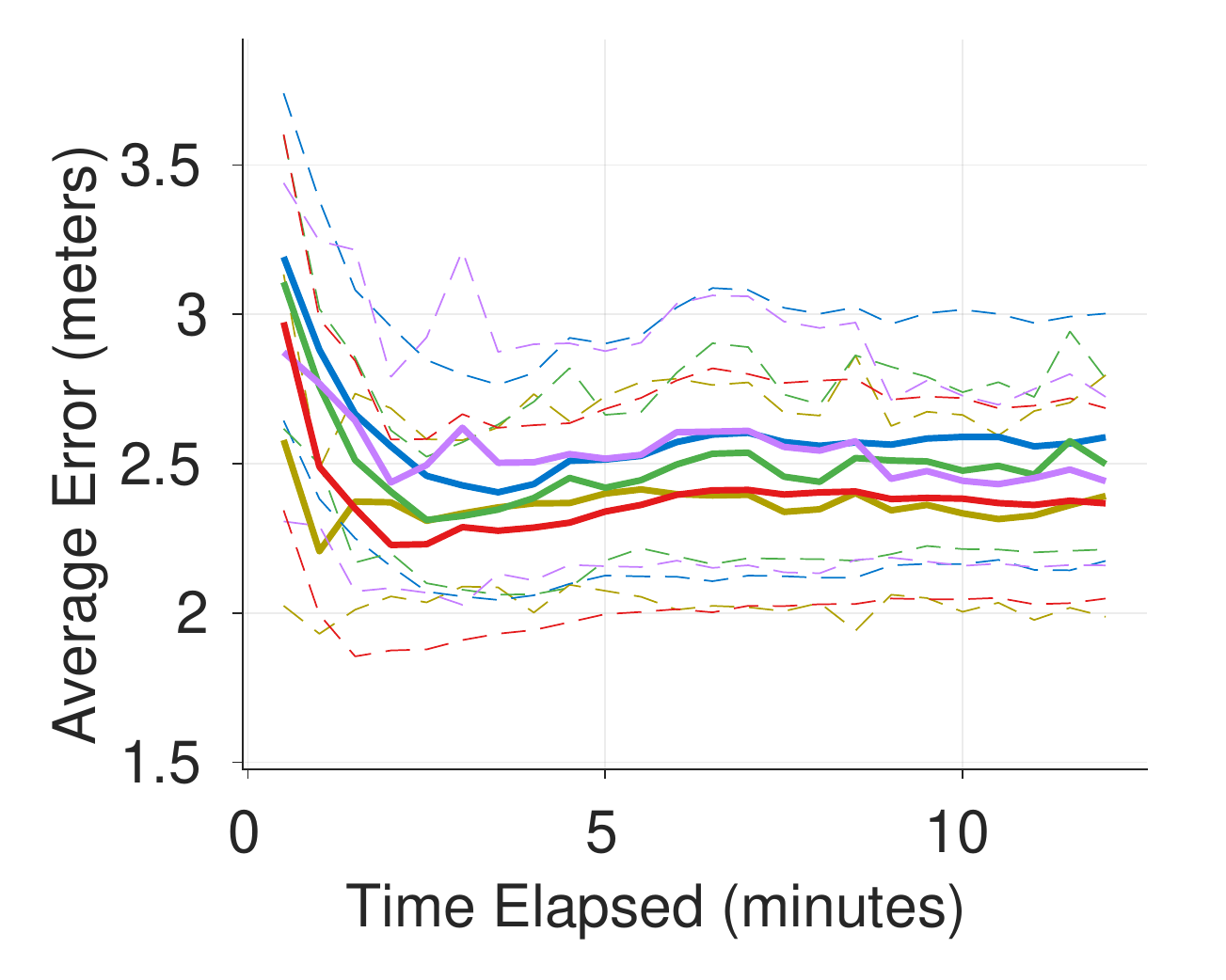}&
     \includegraphics[trim=0cm 0cm 0cm 0cm,clip,width=1.1in]{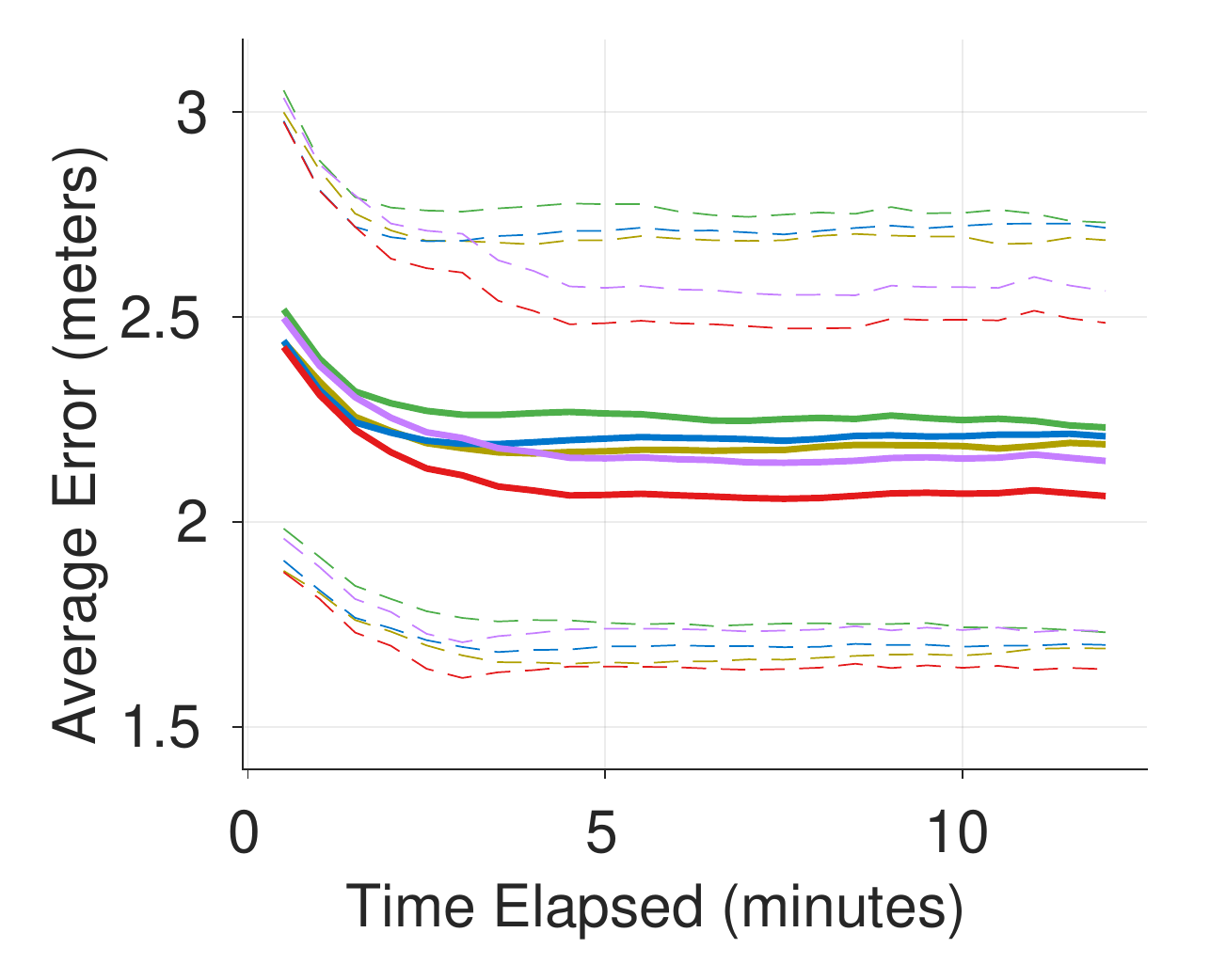}&
     \includegraphics[trim=0cm 0cm 0cm 0cm,clip,width=1.1in]{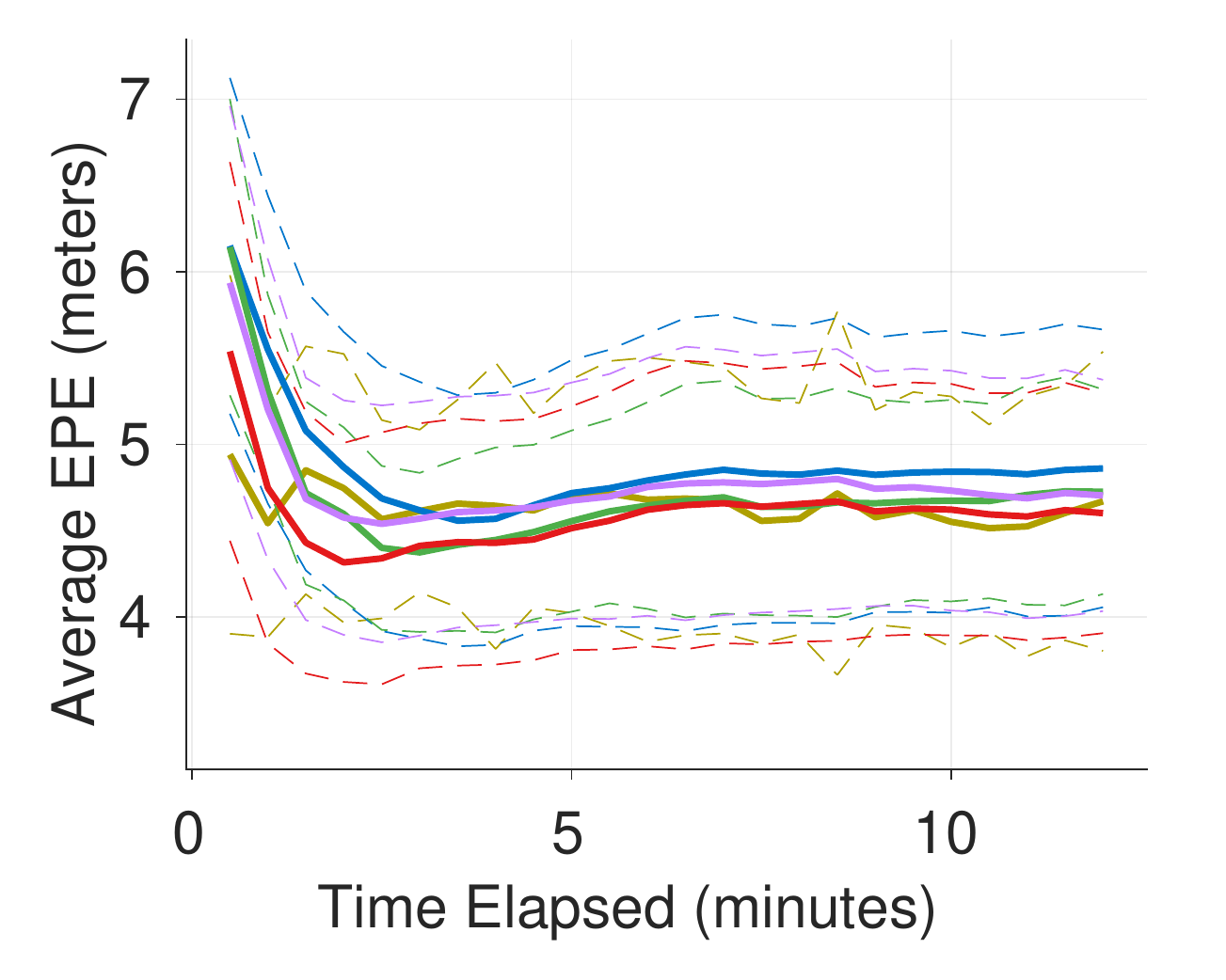}&
     \includegraphics[trim=0cm 0cm 0cm 0cm,clip,width=1.1in]{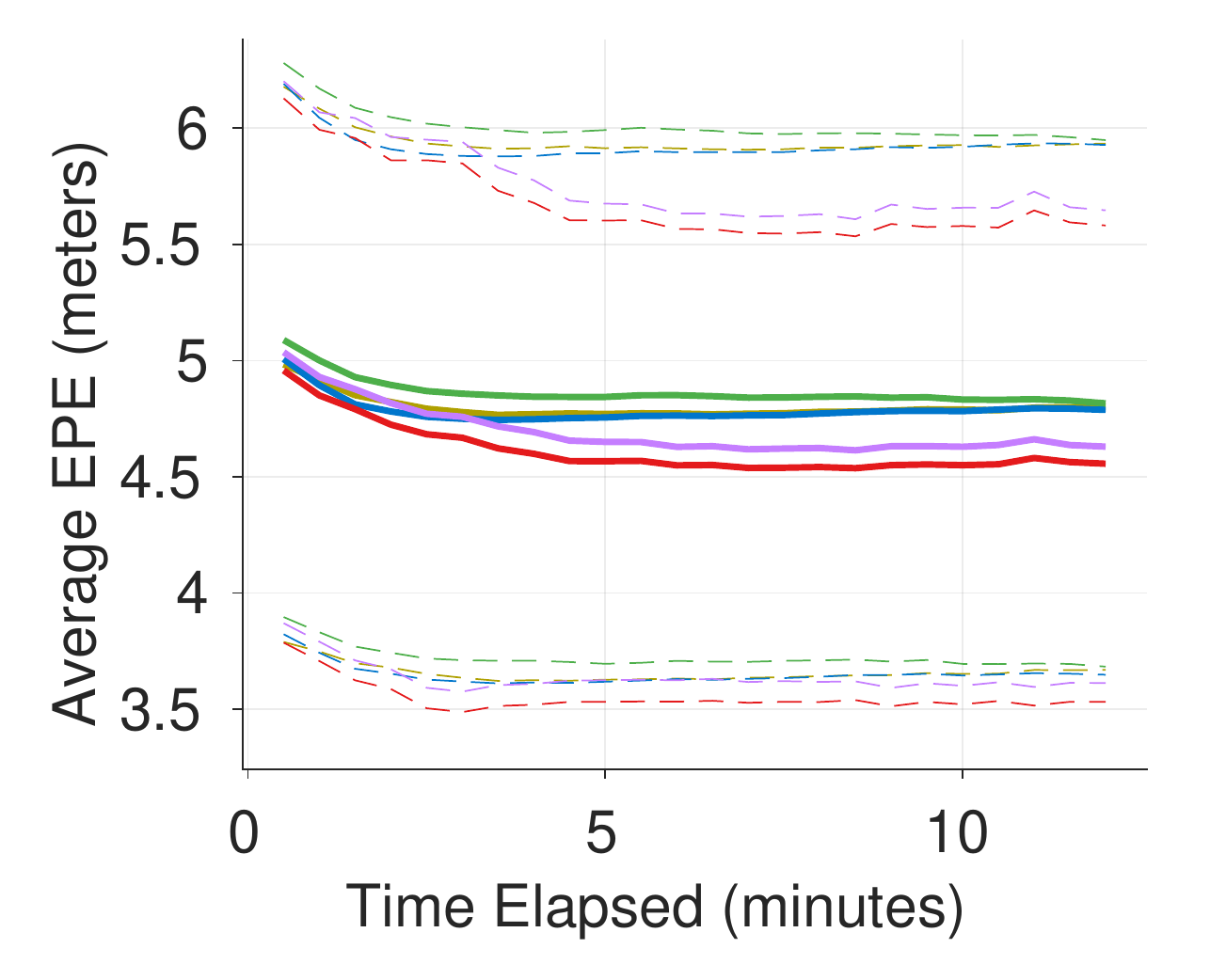}\\
   \multirow{1}{*}{\rotatebox[origin=c]{90}{\parbox[c]{3.5cm}{\centering Prediction T=5s}}}     & (e) & (f) & (g) & (h) \\
 &
    \includegraphics[trim=0cm 0cm 0cm 0cm,clip,width=1.1in]{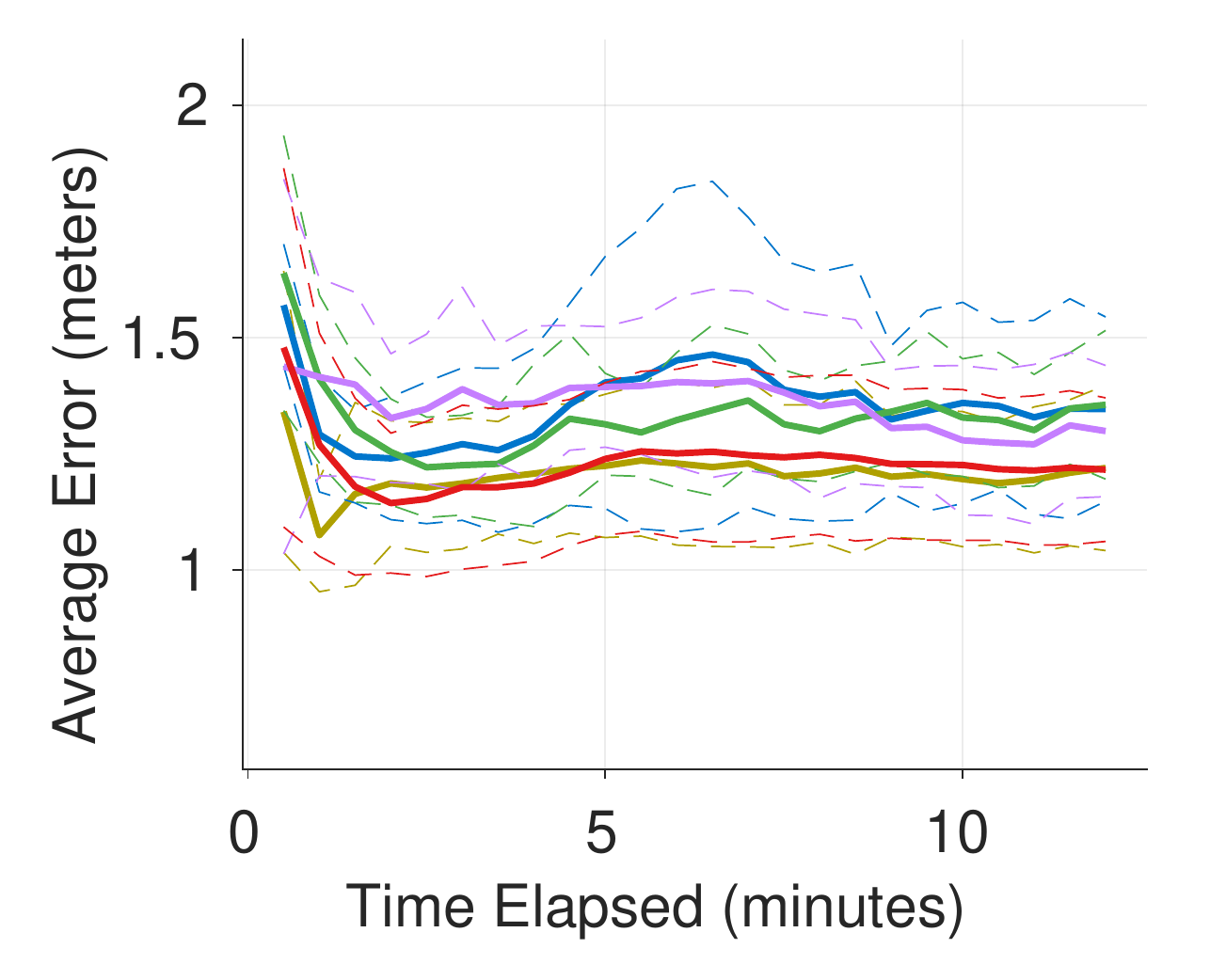}&
     \includegraphics[trim=0cm 0cm 0cm 0cm,clip,width=1.1in]{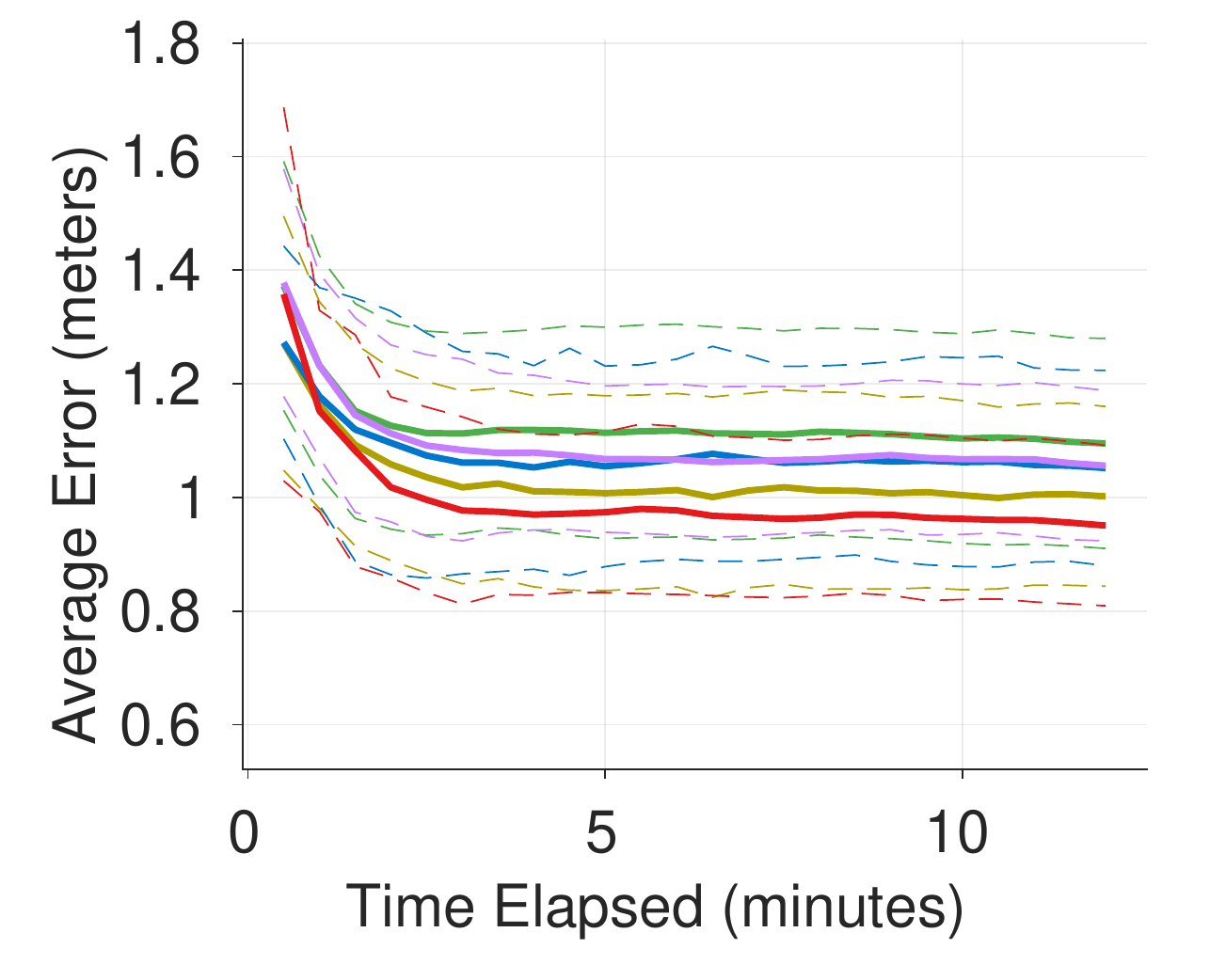}&
     \includegraphics[trim=0cm 0cm 0cm 0cm,clip,width=1.1in]{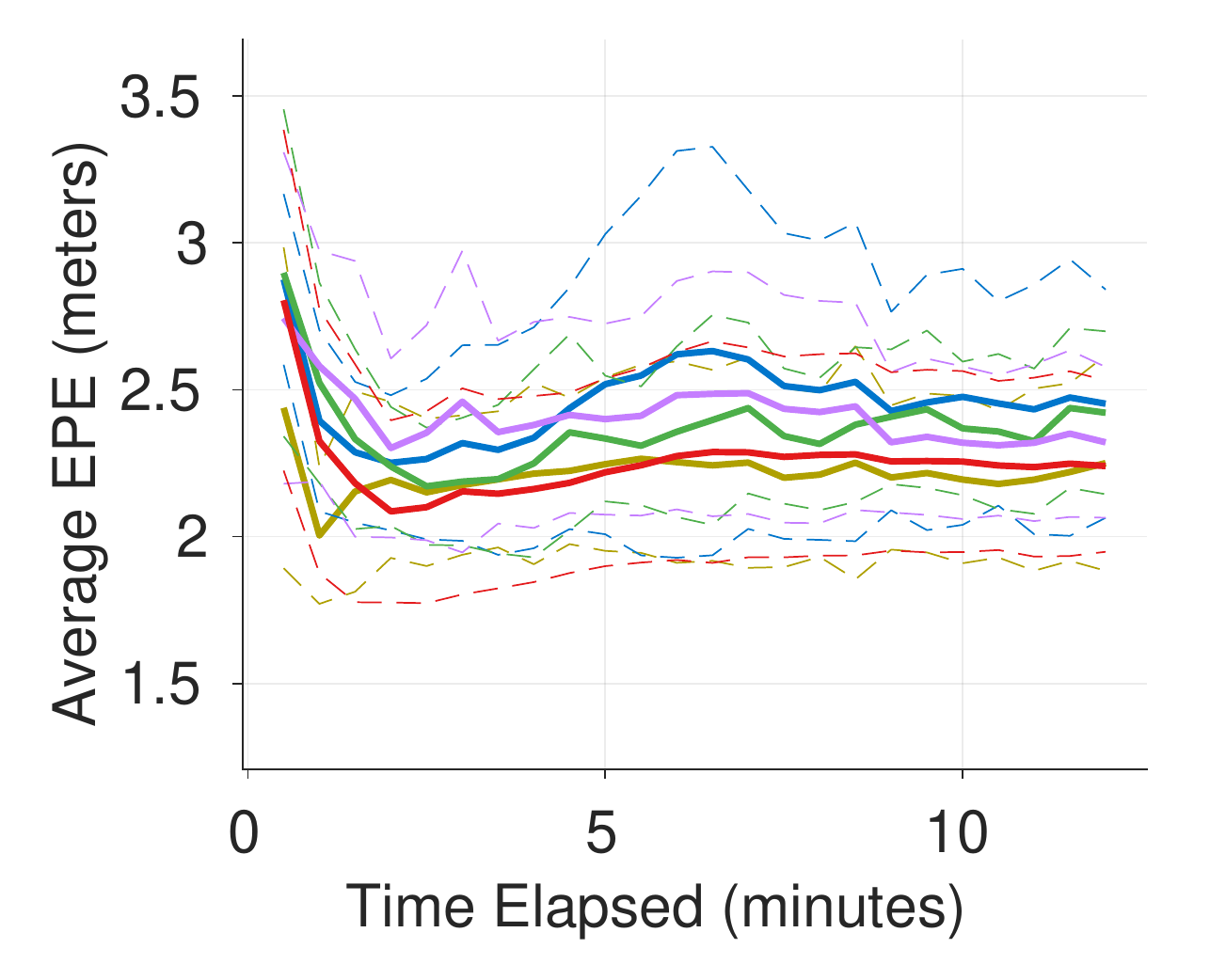}&
     \includegraphics[trim=0cm 0cm 0cm 0cm,clip,width=1.1in]{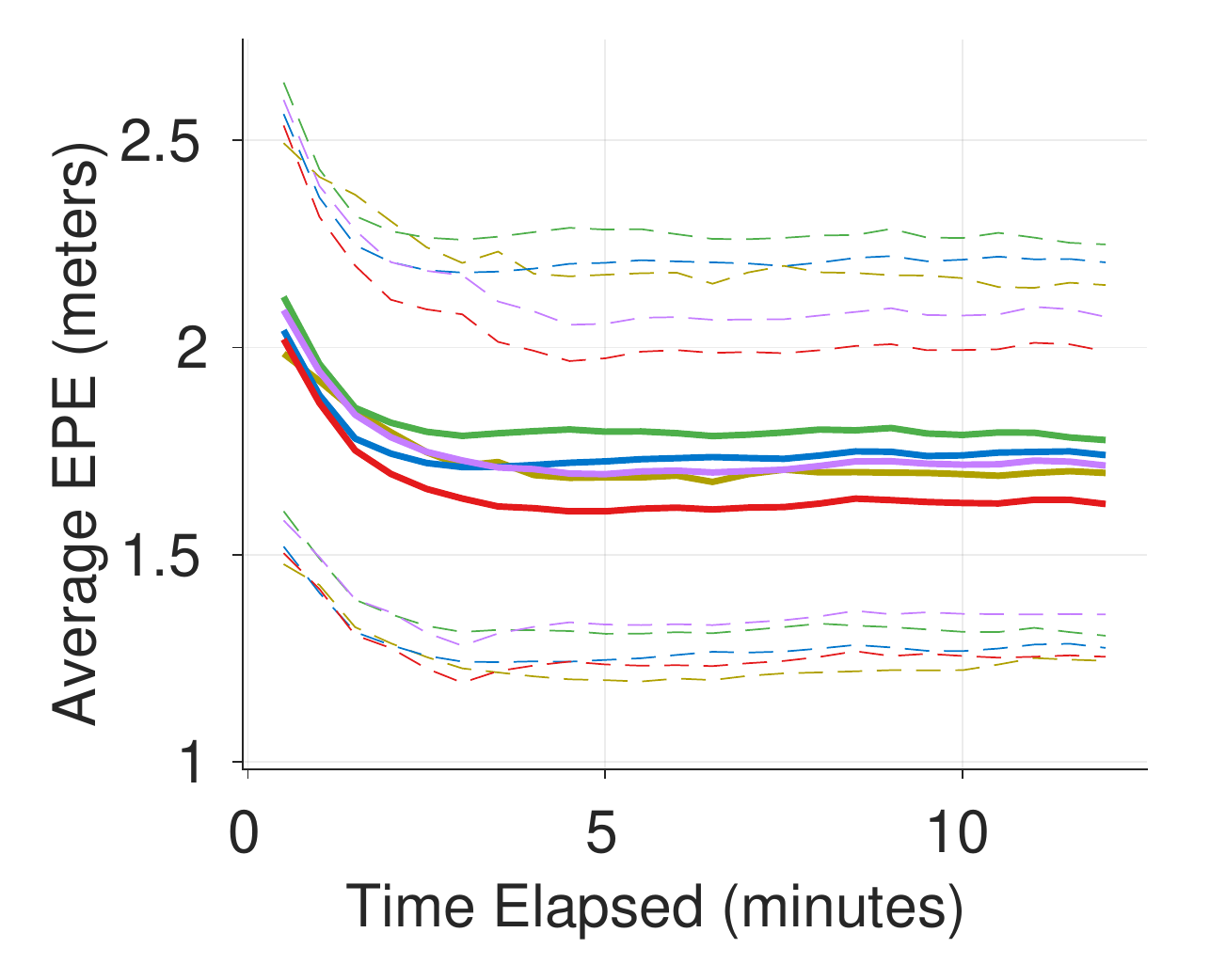}\\
       & (i) & (j) & (k) & (l) 
     \end{tabular}
   \caption{\textbf{Adaptation to new users.} As more data becomes available from the onset of the experiment, we plot the long-term prediction errors (a-d for $T=20$, e-h for $T=10$, and i-l for $T=5$ seconds into the future), of average displacement error along the trajectory and end-point displacement error (EPE). Dashed lines are standard deviation.}~\label{fig:allaexysupp}
 \end{figure}

Fig.~\ref{fig:allaexysupp} shows summarizing plots for adaptation evaluation of the dynamics model to new users. As more data becomes available from the adaptation set (time into the experiment) the model better predicts the future trajectory of the user. In addition to the $T=20$ second prediction results for end-point error in the paper, we also show shorter term prediction results for $T=5$ and $T=10$ seconds. For evaluation, we employ average displacement error (in meters) across all points along the predicted and actual trajectories and the end-point displacement error between the trajectory end-points.

\clearpage

\begin{figure}[!t]
   \centering
   \begin{tabular}{ccccc}
 \multicolumn{5}{c}{\fbox{\includegraphics[trim=0cm 0cm 0cm 0cm,clip,width=4in]{fig/legfinal.pdf}}} 
 \\
 \\
  \multicolumn{5}{c}{\textbf{Entire Dataset}}
 \\
     \includegraphics[trim=0cm 0cm 0cm 0cm,clip,width=.8in]{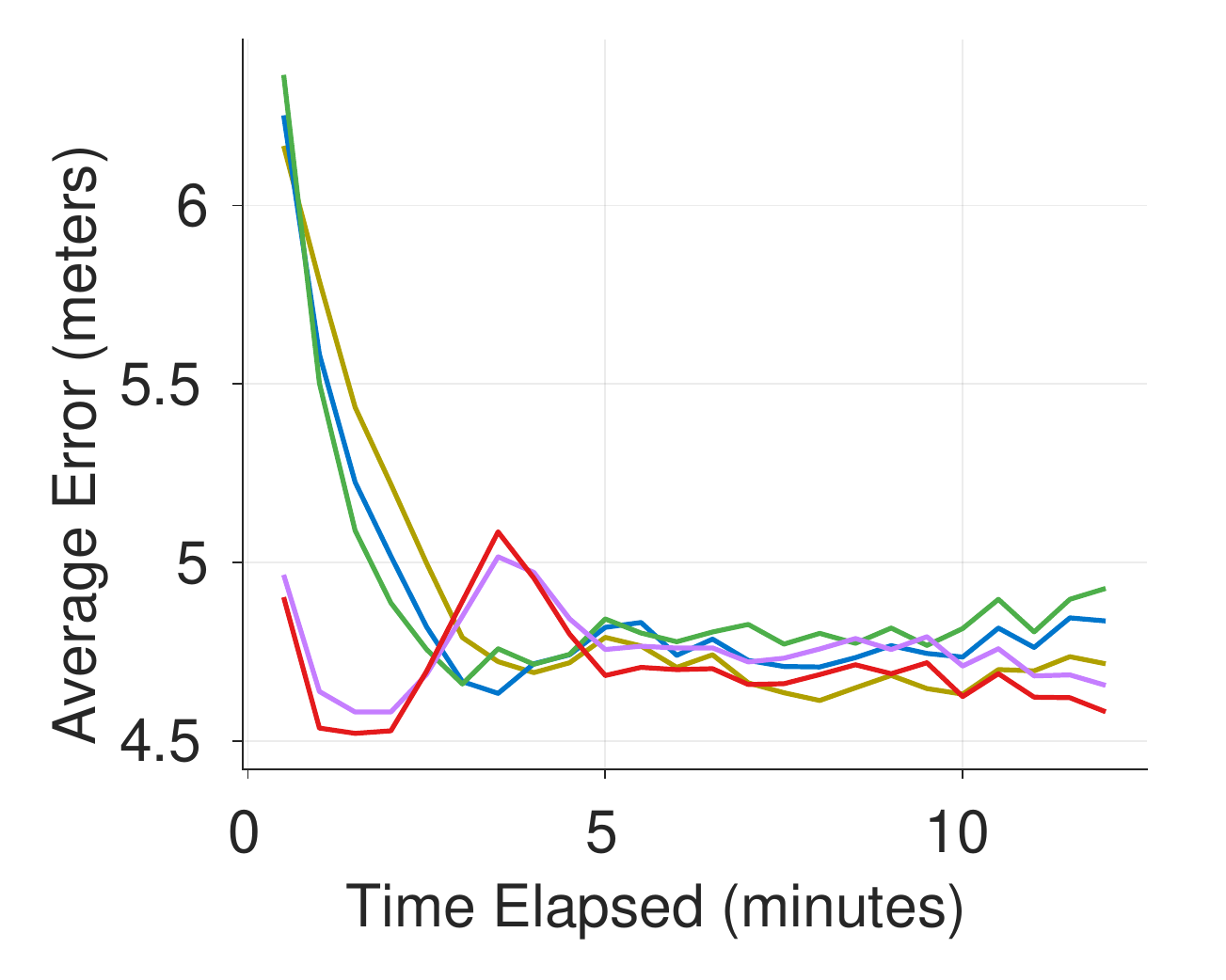}&
      \includegraphics[trim=0cm 0cm 0cm 0cm,clip,width=.8in]{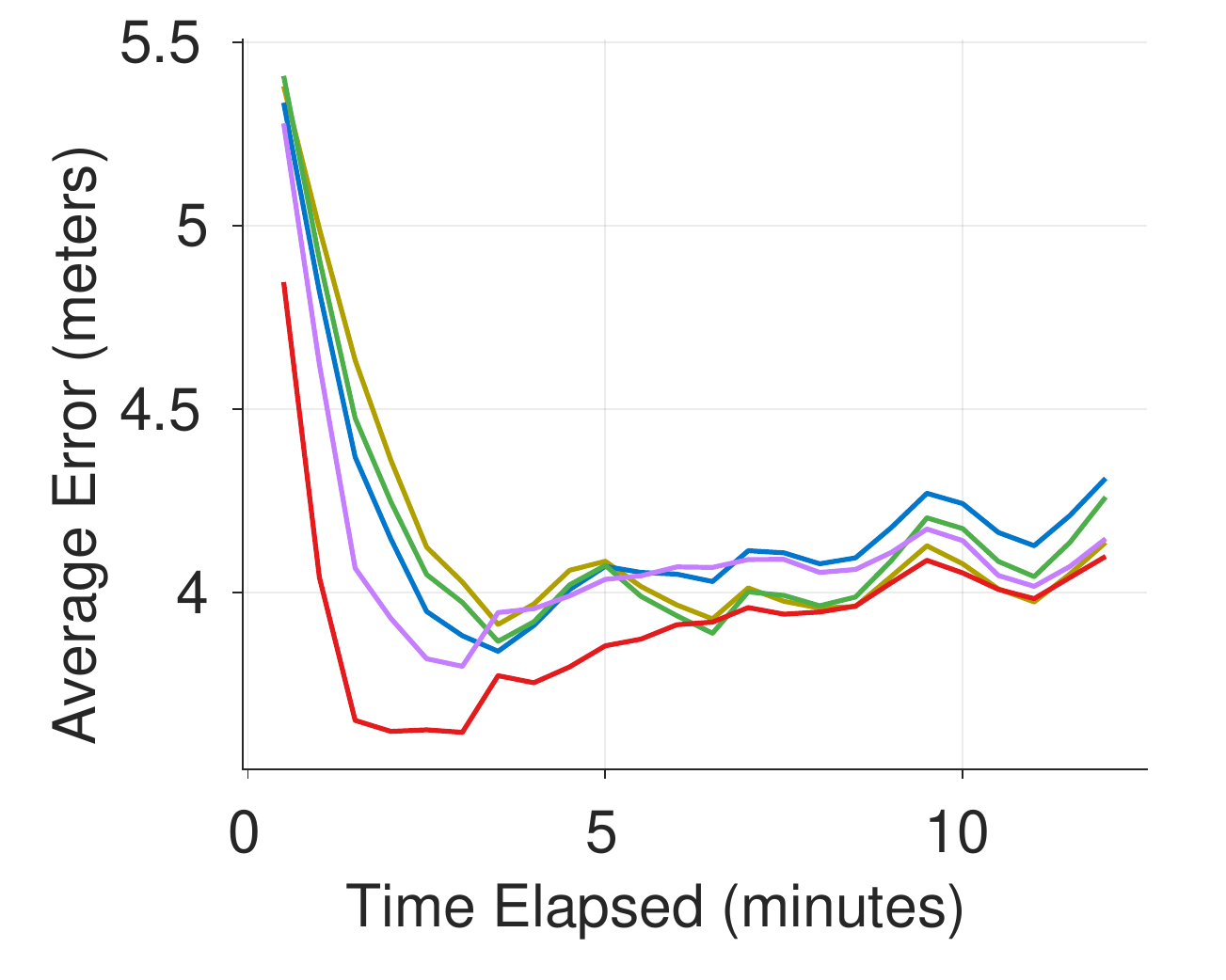}&
      \includegraphics[trim=0cm 0cm 0cm 0cm,clip,width=.8in]{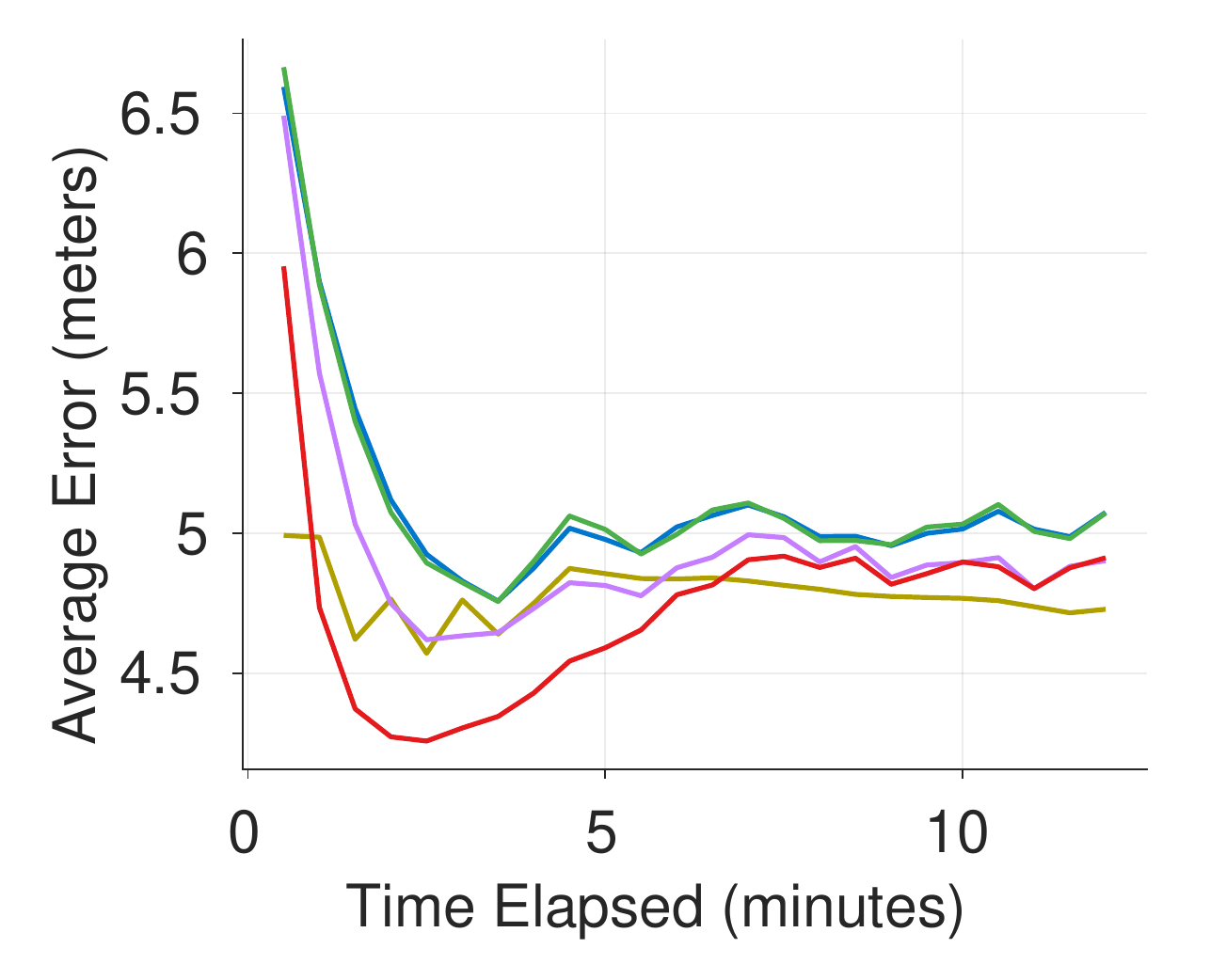}&
      \includegraphics[trim=0cm 0cm 0cm 0cm,clip,width=.8in]{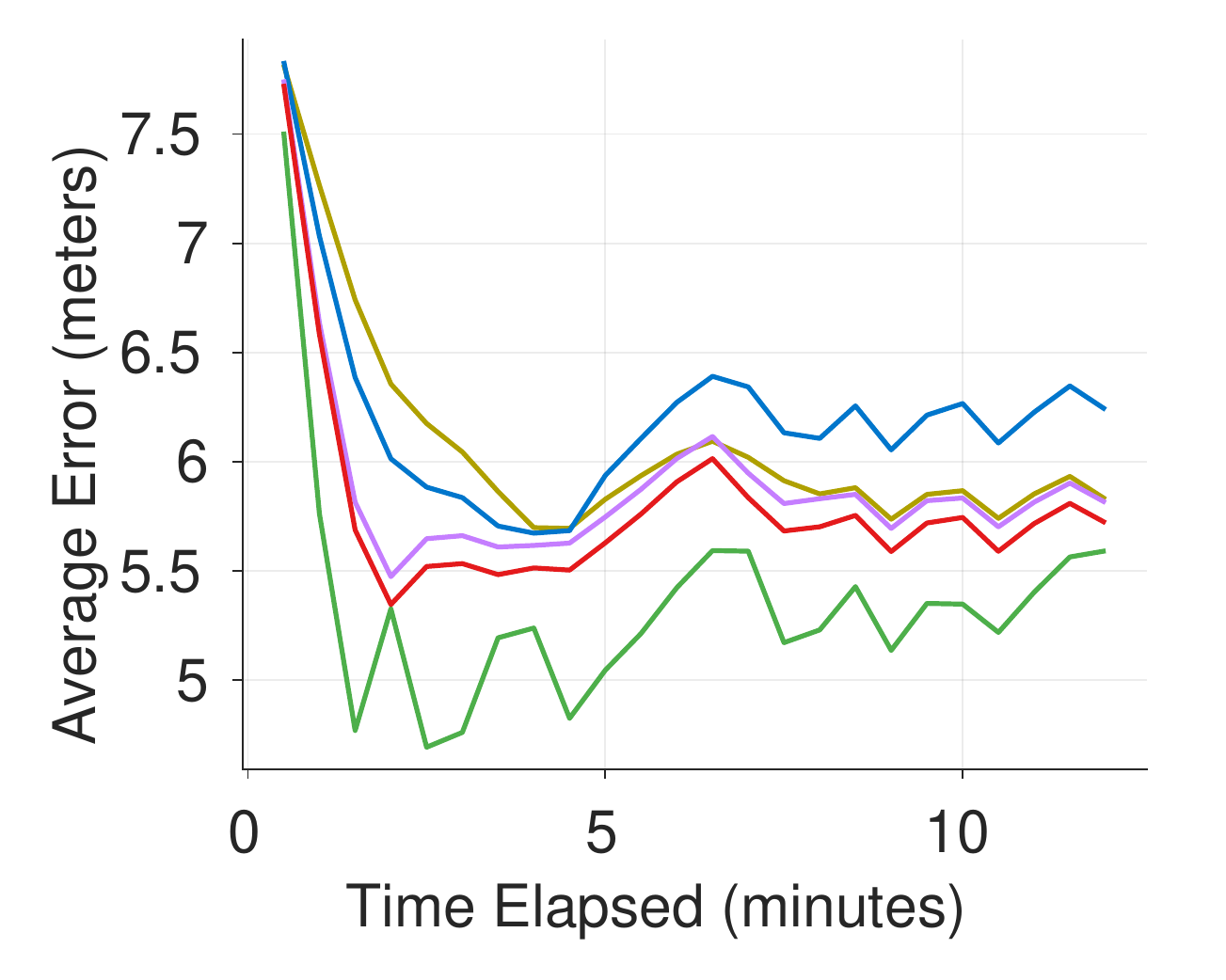}&
      \includegraphics[trim=0cm 0cm 0cm 0cm,clip,width=.8in]{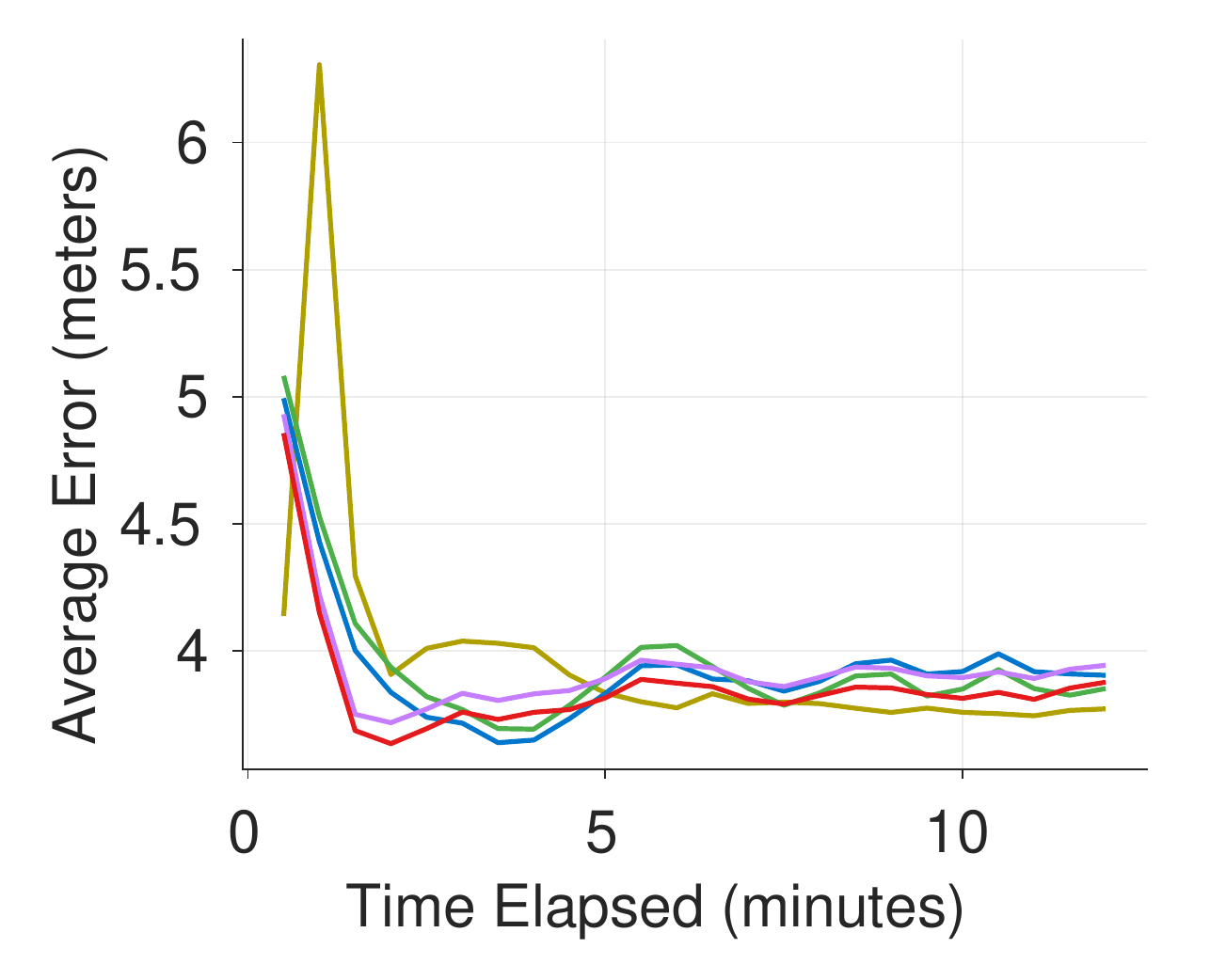}
        \\
            \includegraphics[trim=0cm 0cm 0cm 0cm,clip,width=.8in]{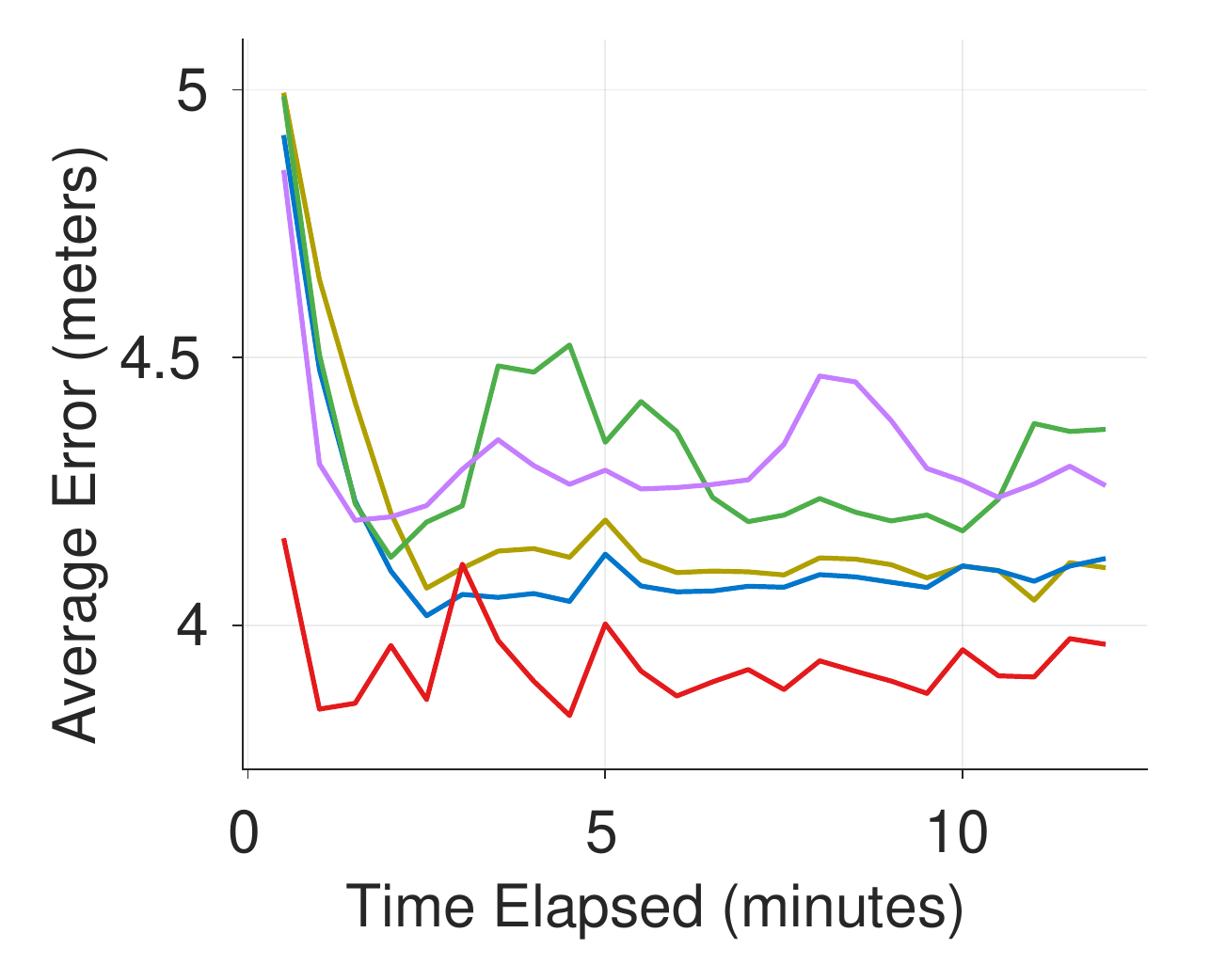}&
      \includegraphics[trim=0cm 0cm 0cm 0cm,clip,width=.8in]{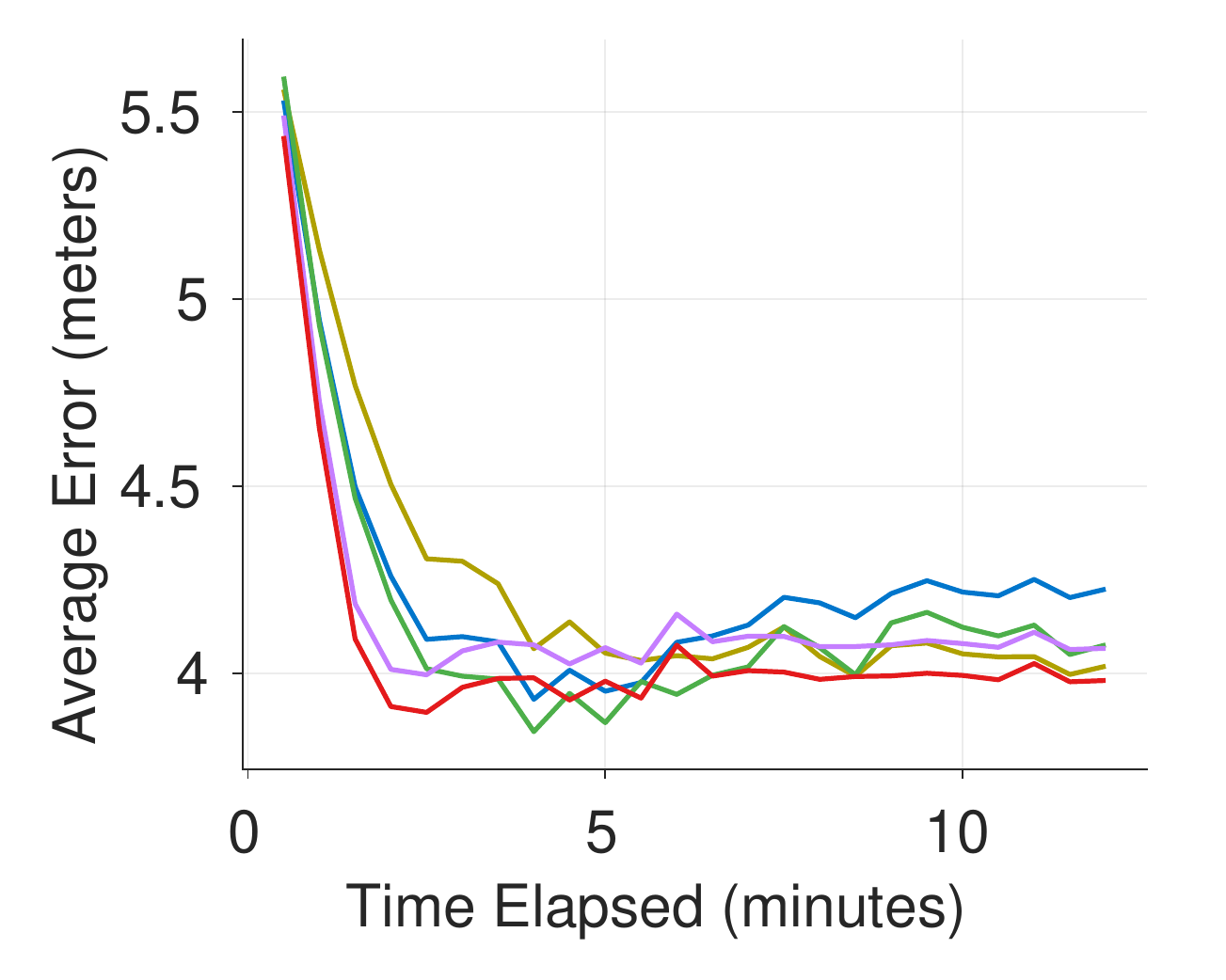}&
      \includegraphics[trim=0cm 0cm 0cm 0cm,clip,width=.8in]{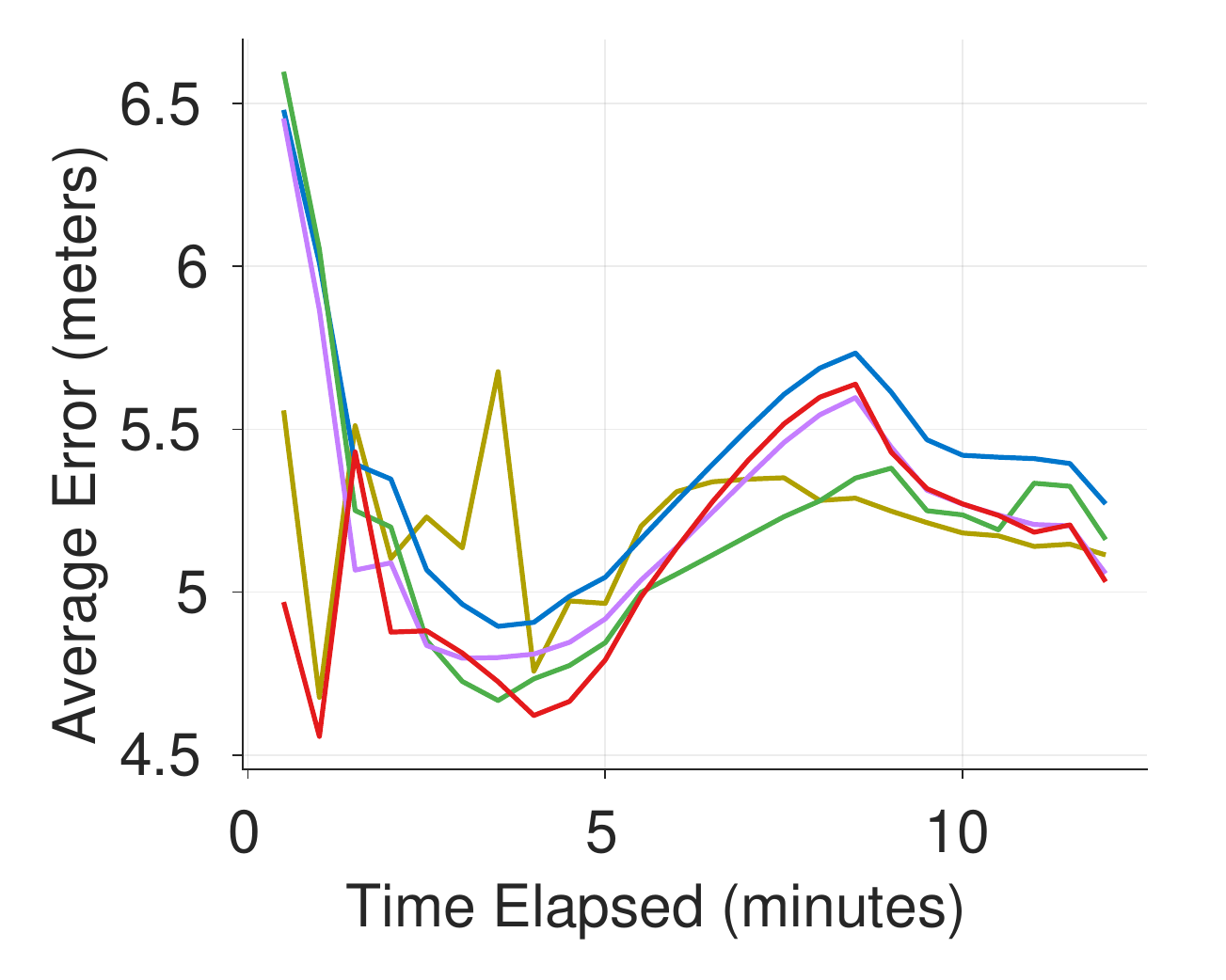}&
      \includegraphics[trim=0cm 0cm 0cm 0cm,clip,width=.8in]{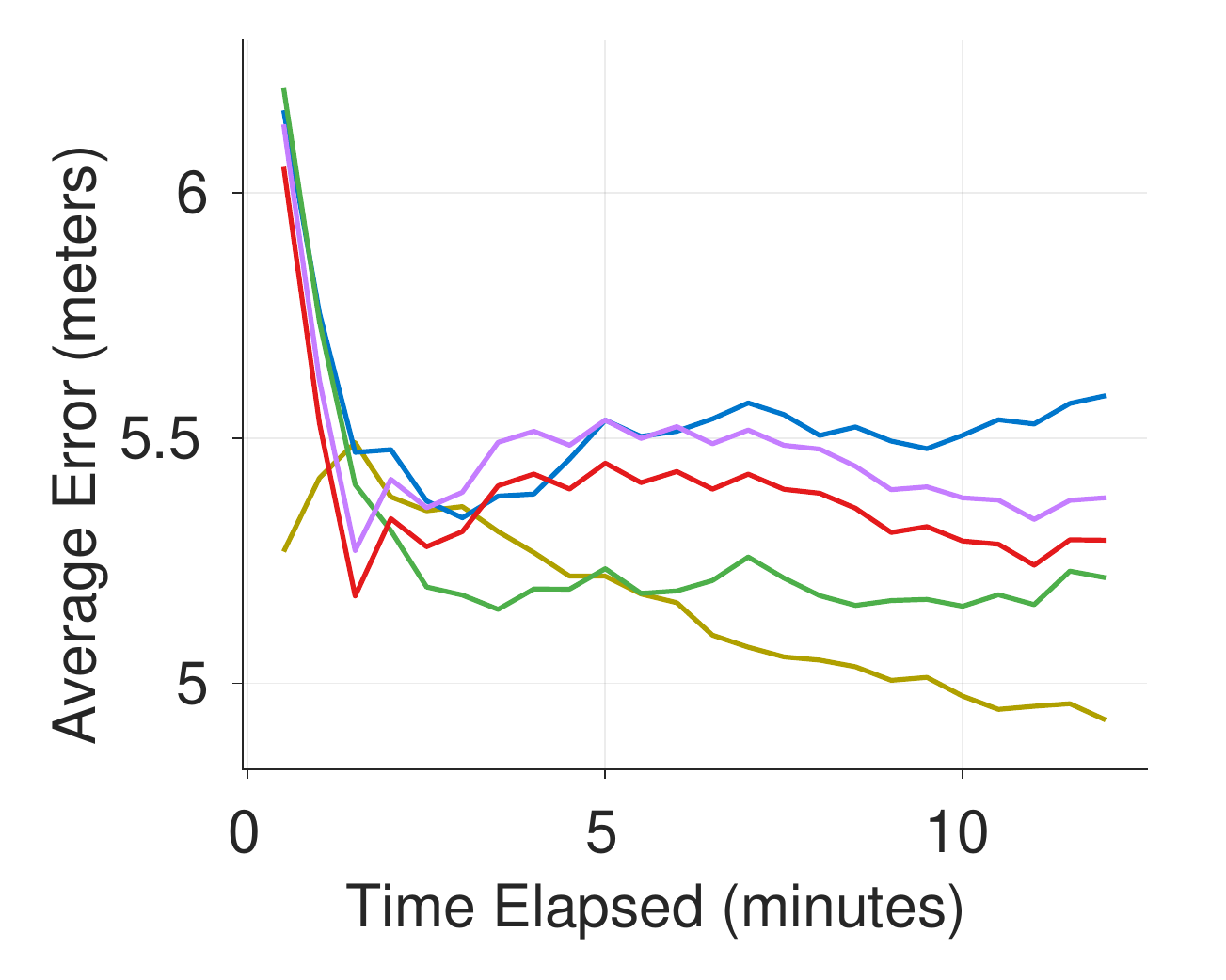}&
        \\
         \multicolumn{5}{c}{\textbf{Turn Events}}
         \\
                      \includegraphics[trim=0cm 0cm 0cm 0cm,clip,width=.8in]{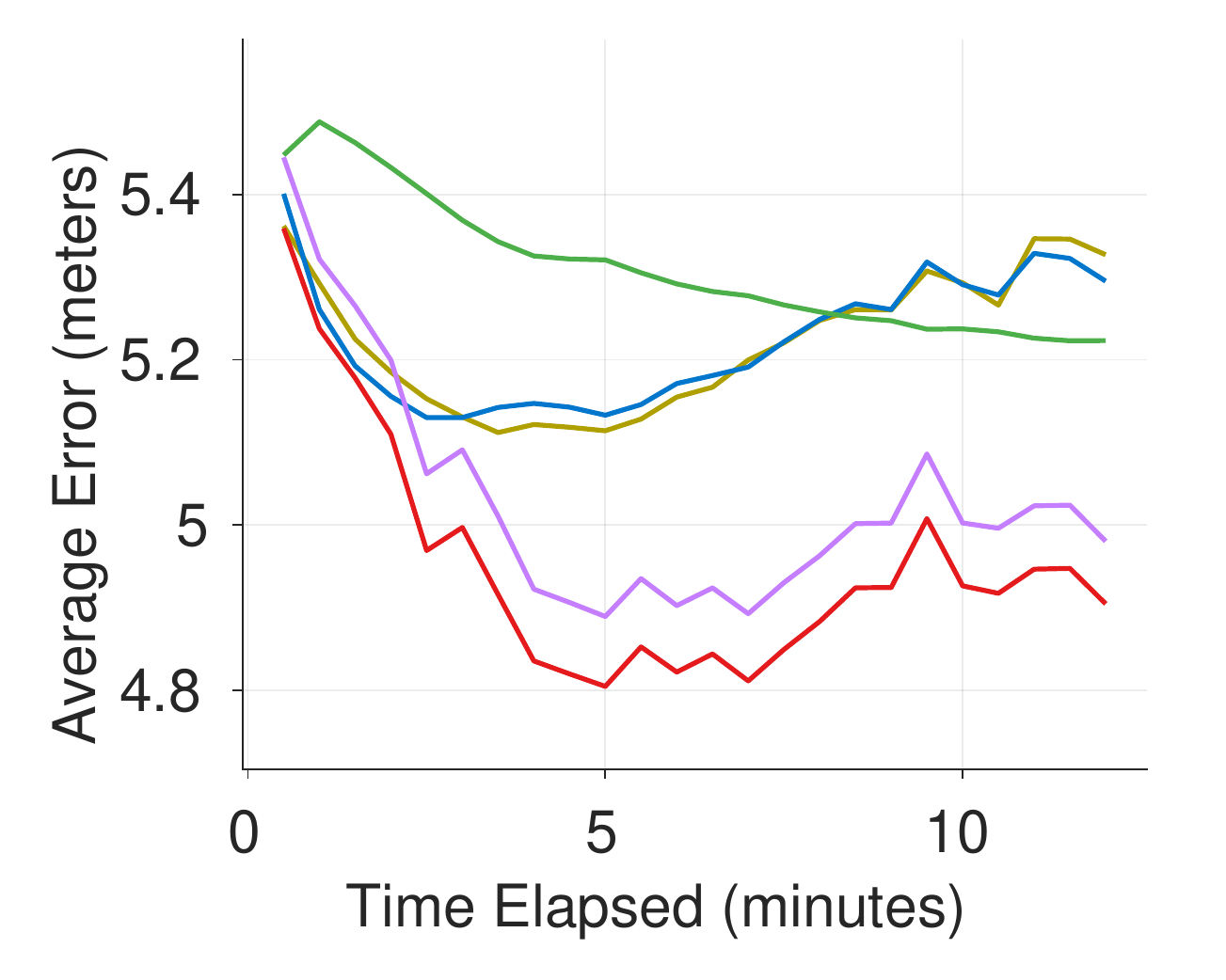}&
      \includegraphics[trim=0cm 0cm 0cm 0cm,clip,width=.8in]{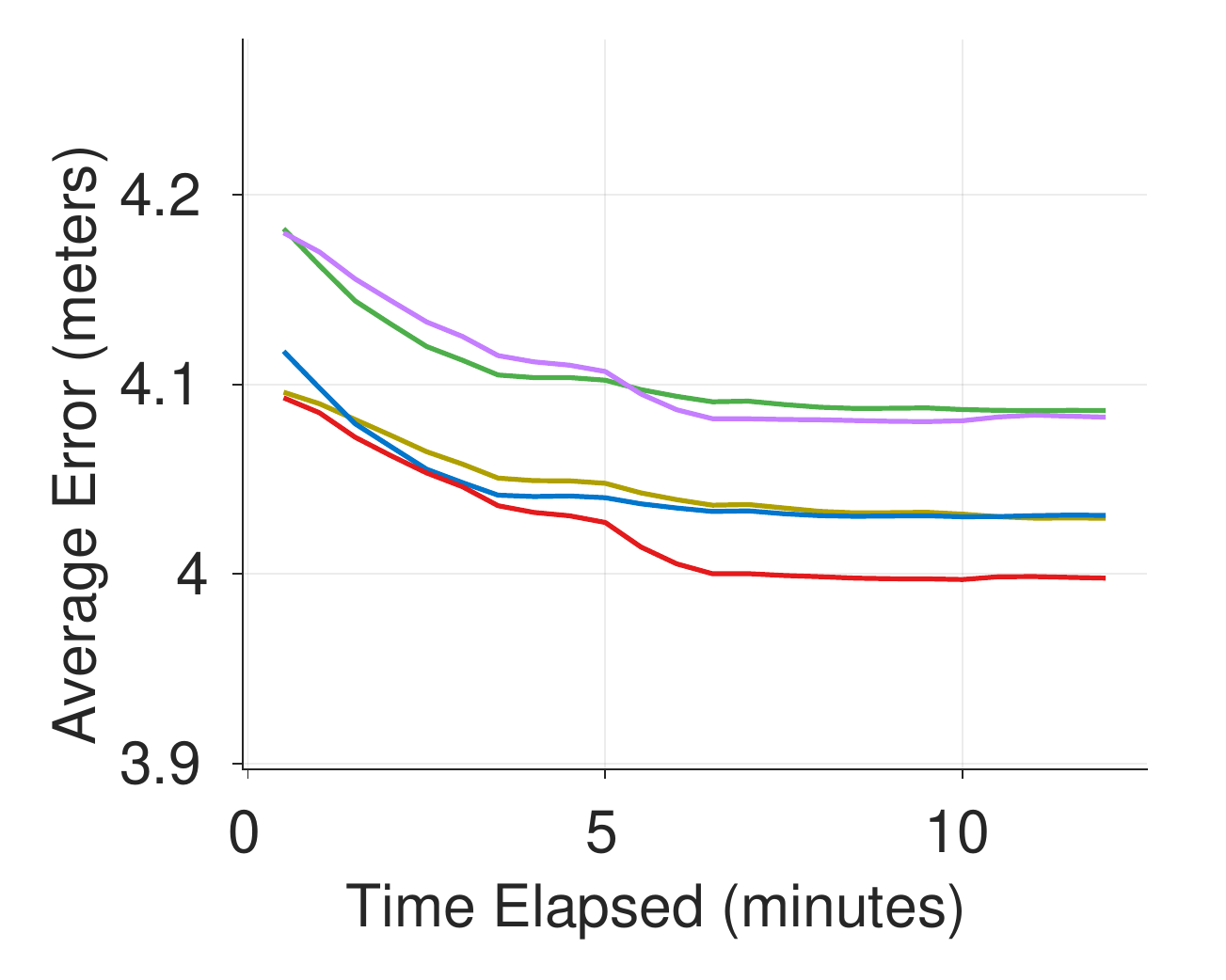}&
      \includegraphics[trim=0cm 0cm 0cm 0cm,clip,width=.8in]{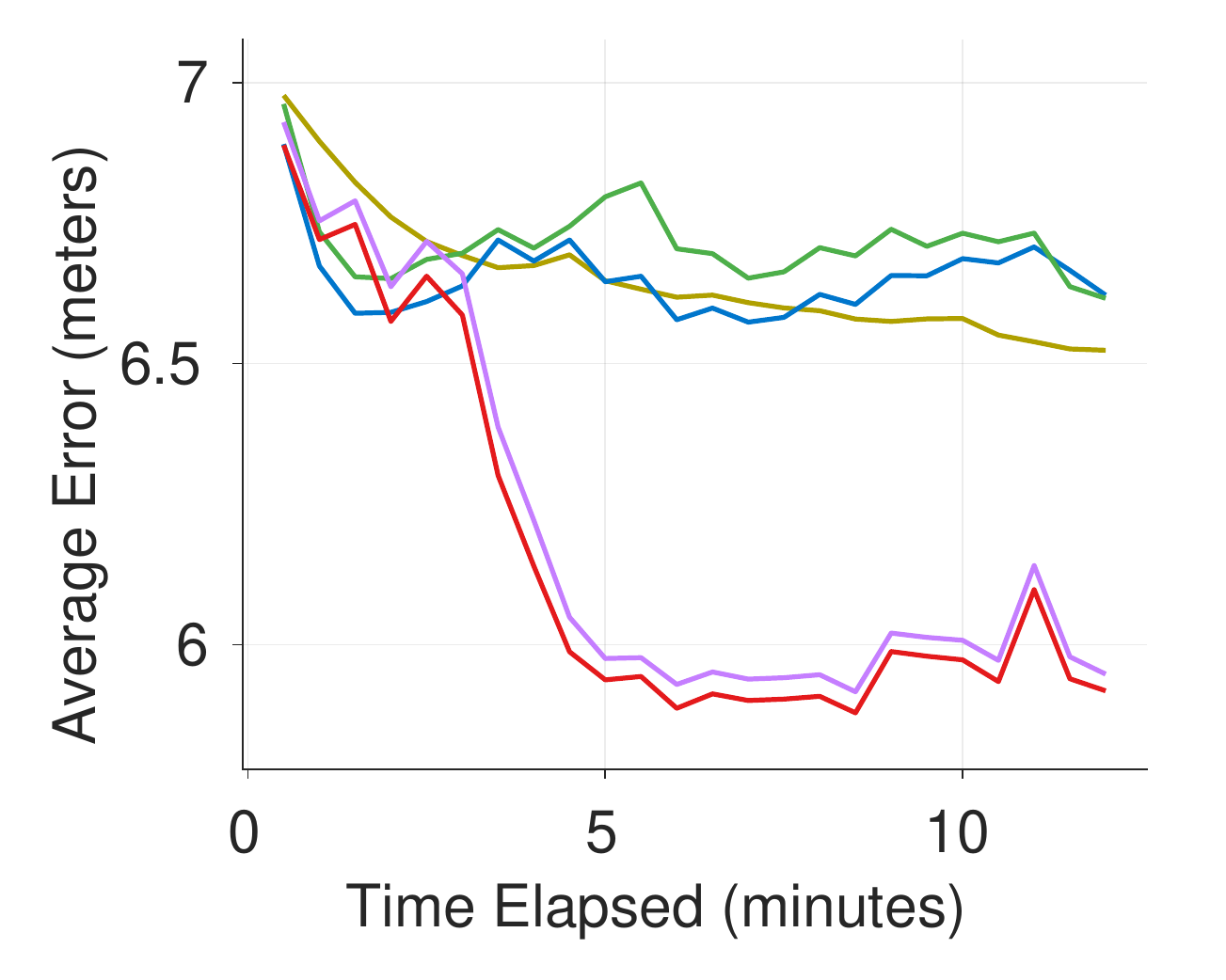}&
      \includegraphics[trim=0cm 0cm 0cm 0cm,clip,width=.8in]{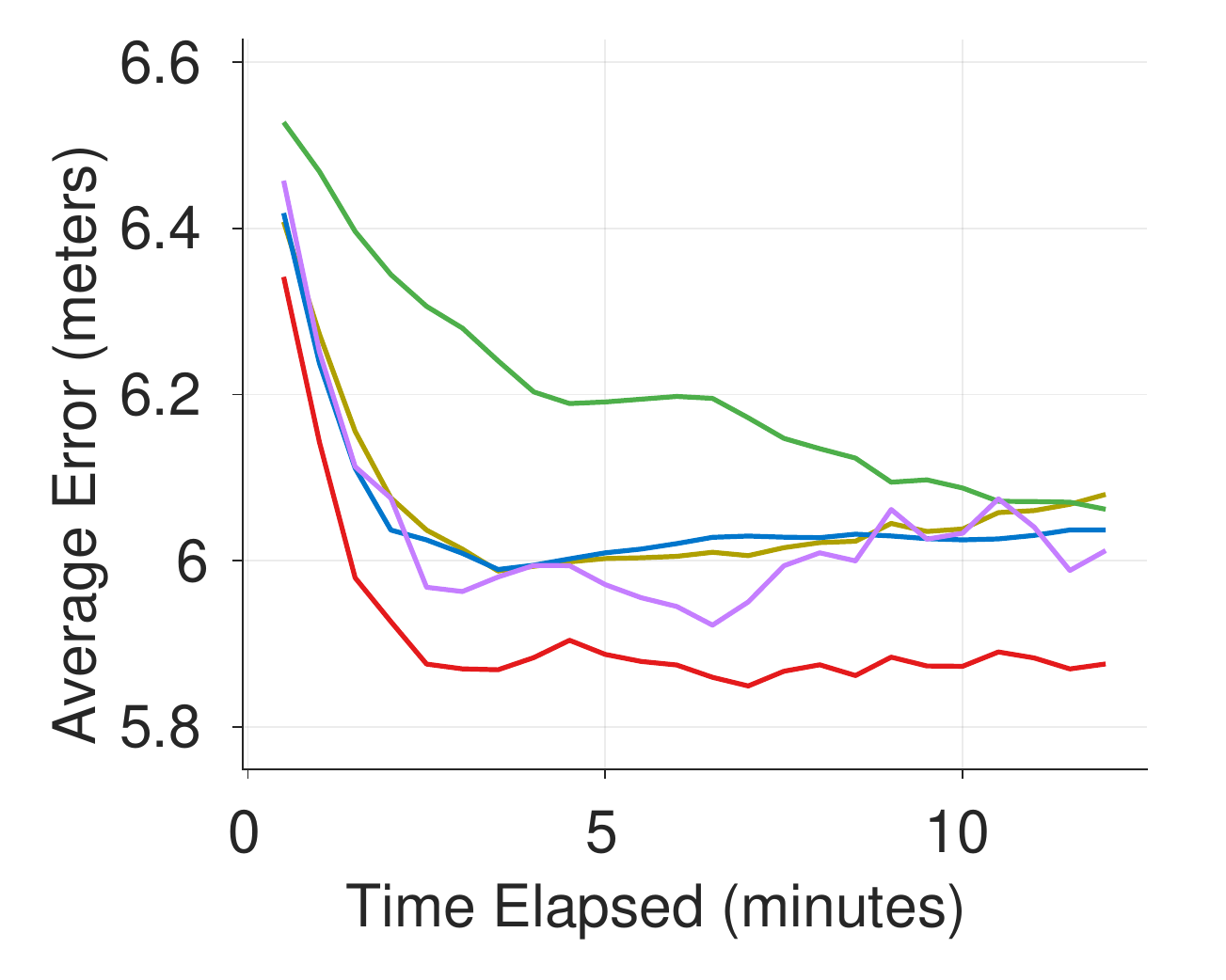}&
      \includegraphics[trim=0cm 0cm 0cm 0cm,clip,width=.8in]{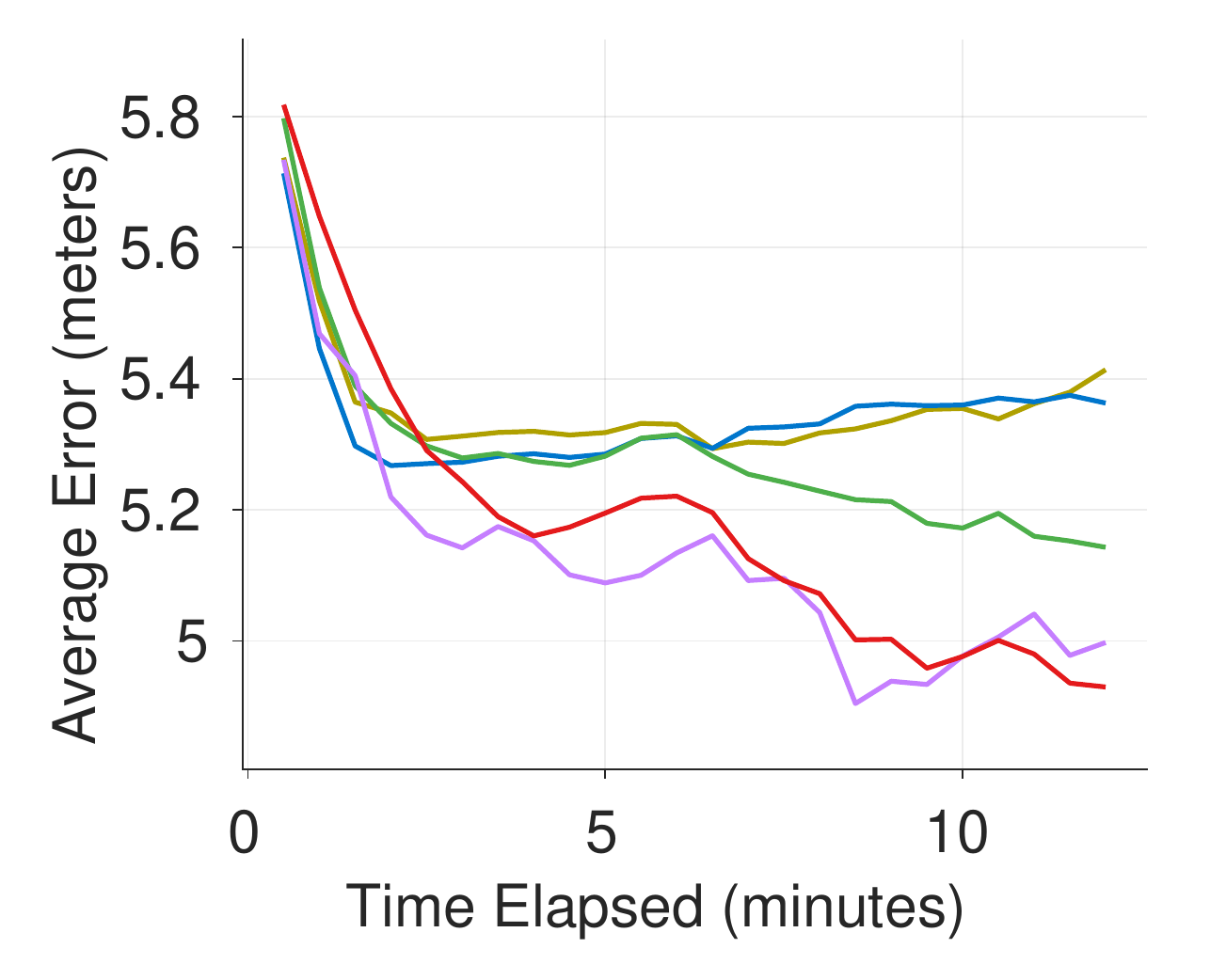}
        \\
            \includegraphics[trim=0cm 0cm 0cm 0cm,clip,width=.8in]{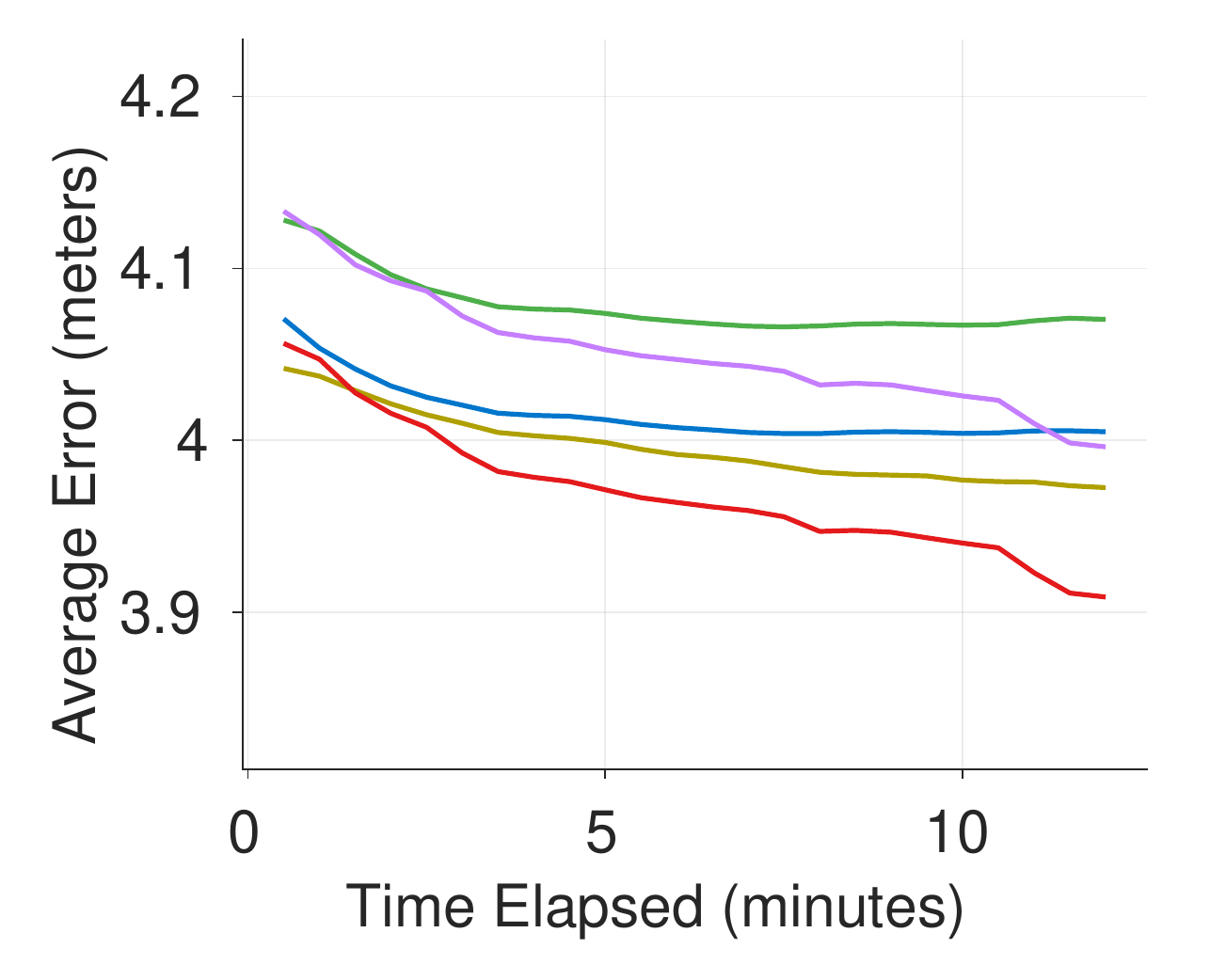}&
      \includegraphics[trim=0cm 0cm 0cm 0cm,clip,width=.8in]{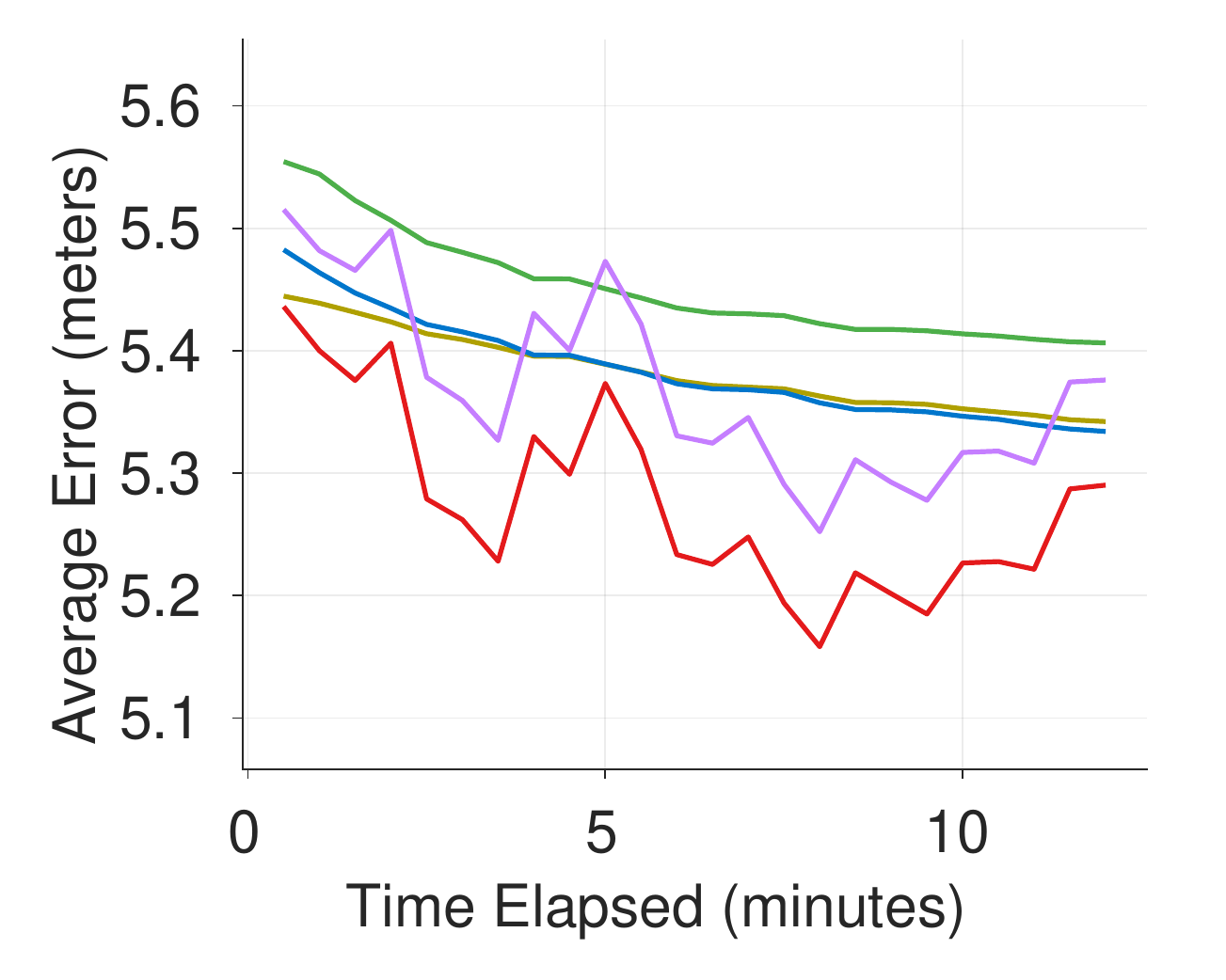}&
      \includegraphics[trim=0cm 0cm 0cm 0cm,clip,width=.8in]{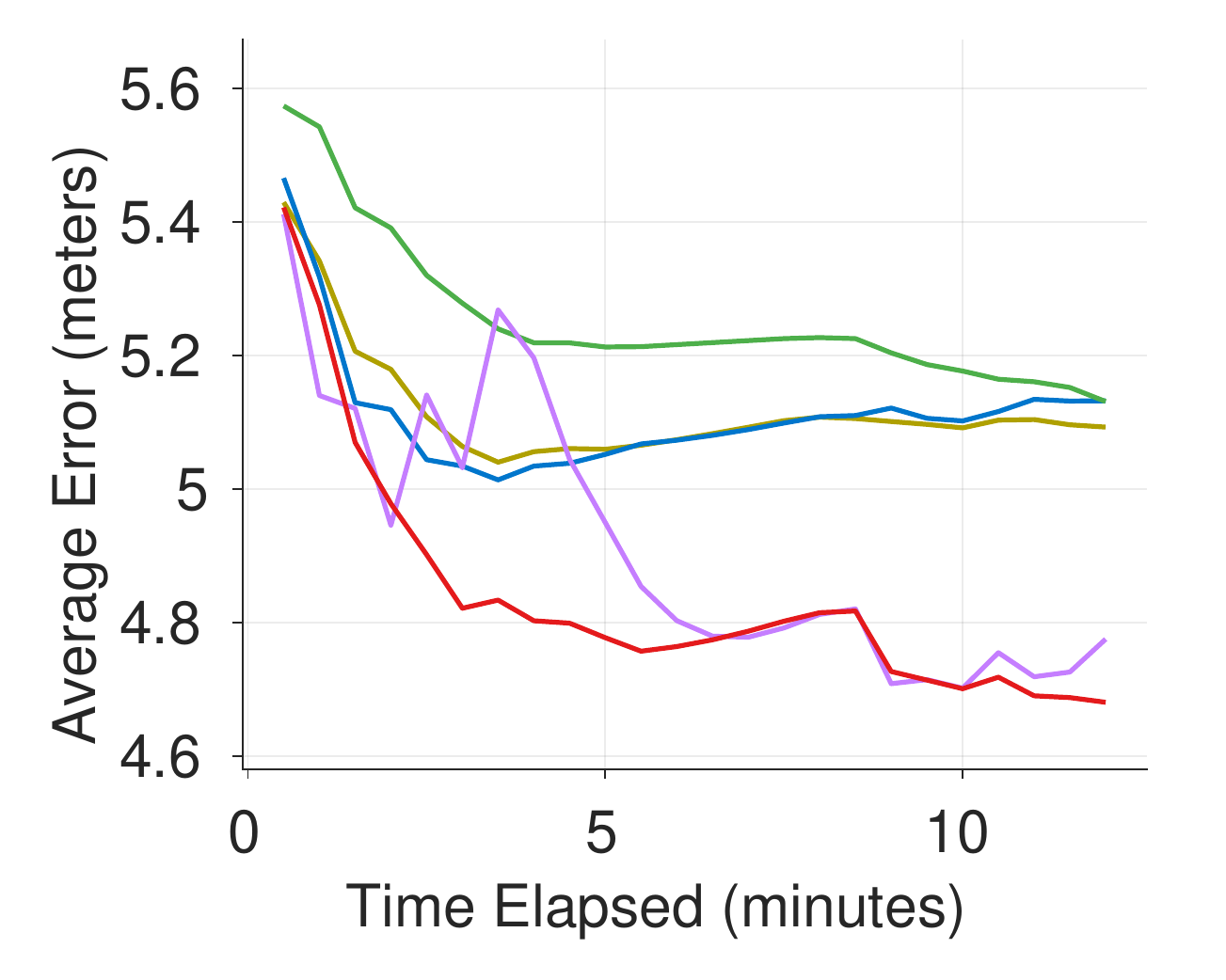}&
      \includegraphics[trim=0cm 0cm 0cm 0cm,clip,width=.8in]{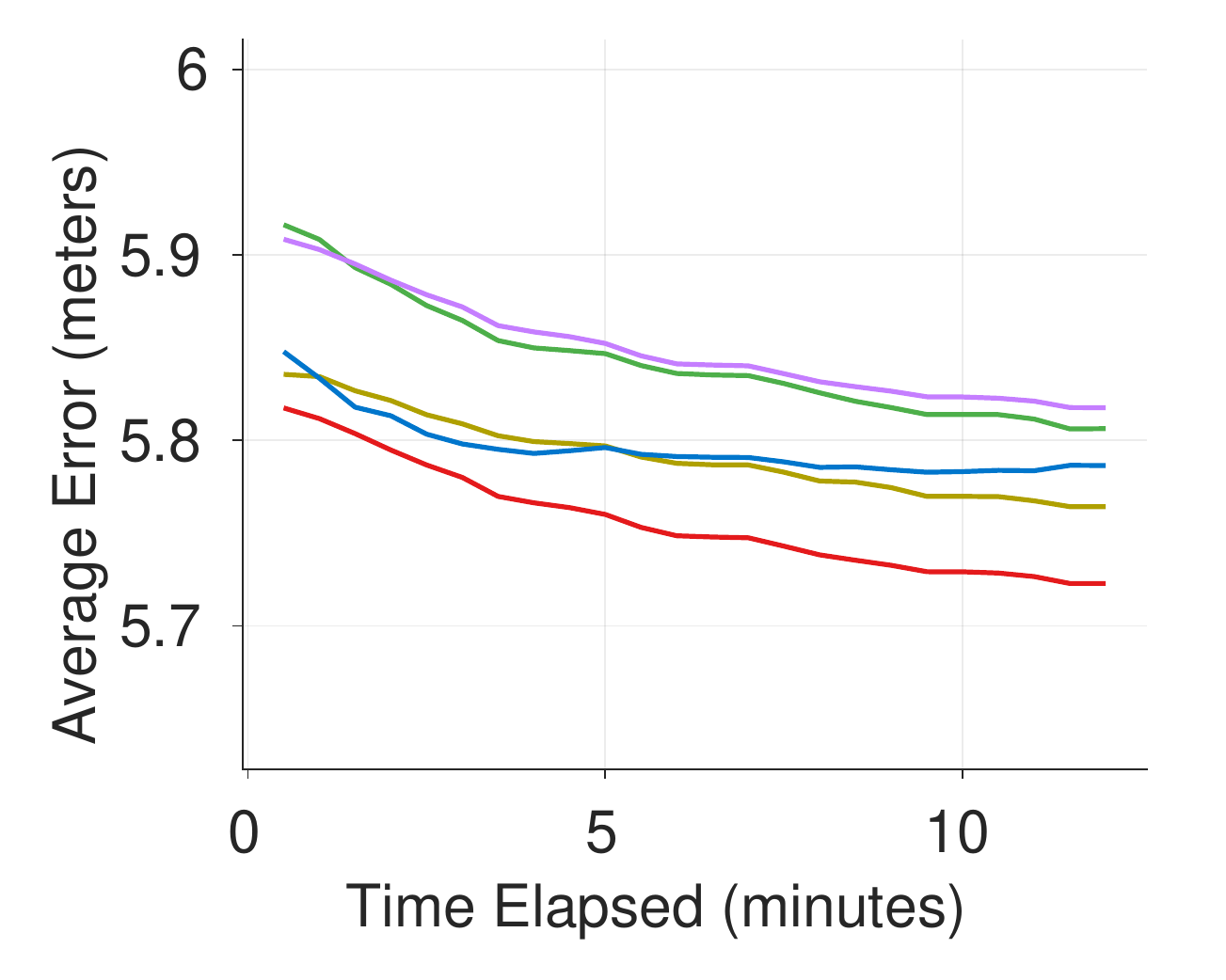}&
     \end{tabular}
   \caption{\textbf{Adaptation to new users.} Results for individual users.}~\label{fig:allaexysuppind}
 \end{figure}

\subsection{Dataset Analysis}

User reactions to the interface ultimately determine the success of the navigation task. To motivate the proposed approach for accommodating individual differences in blind navigation, we include a data-driven analysis of this critical issue in system design. The analysis reveals the impact of personalization on navigation success and end-user experience. To analyze per-user variability in the data, we will plot speed measures on an individual basis within each instruction type (the measures we are ultimately trying to model and predict in the main paper). We study average speed over a time window of 6 seconds following an instruction onset to capture user motion (linear speed in meters per second and angular speed in radians per second, often averaged across all events in the dataset). We also report average speed change following an instruction~\cite{variabilityohnbar}.
    
  Statistical testing is performed to ensure statistical significance in user behavior variability. For each test, we set the participant id as the independent variable (for studying inter-user variability). When reporting findings, we fix a motion measure (\emph{e.g.,} average angular speed) as the dependent variable. In all experiments normality is first verified using a Shapiro-Wilk test, and followed by ANOVA under normality or a t-test otherwise, with an alpha level of $.05$.

\subsubsection{Turn Events - User Variability Analysis}

Turning events require a detailed analysis, as they exhibit the most significant variability, take multiple seconds to perform, and are crucial to the overall navigation task~\cite{ahmetovic2016navcog2}. The majority of navigation errors in the dataset occur during turns. The figures show the spectrum of variability among users.


   \begin{figure}[H]
  \centering
  \begin{tabular}{ccc}
 \multicolumn{3}{c}{\textbf{Large Turns}}
\\
   \includegraphics[width=1.5in]{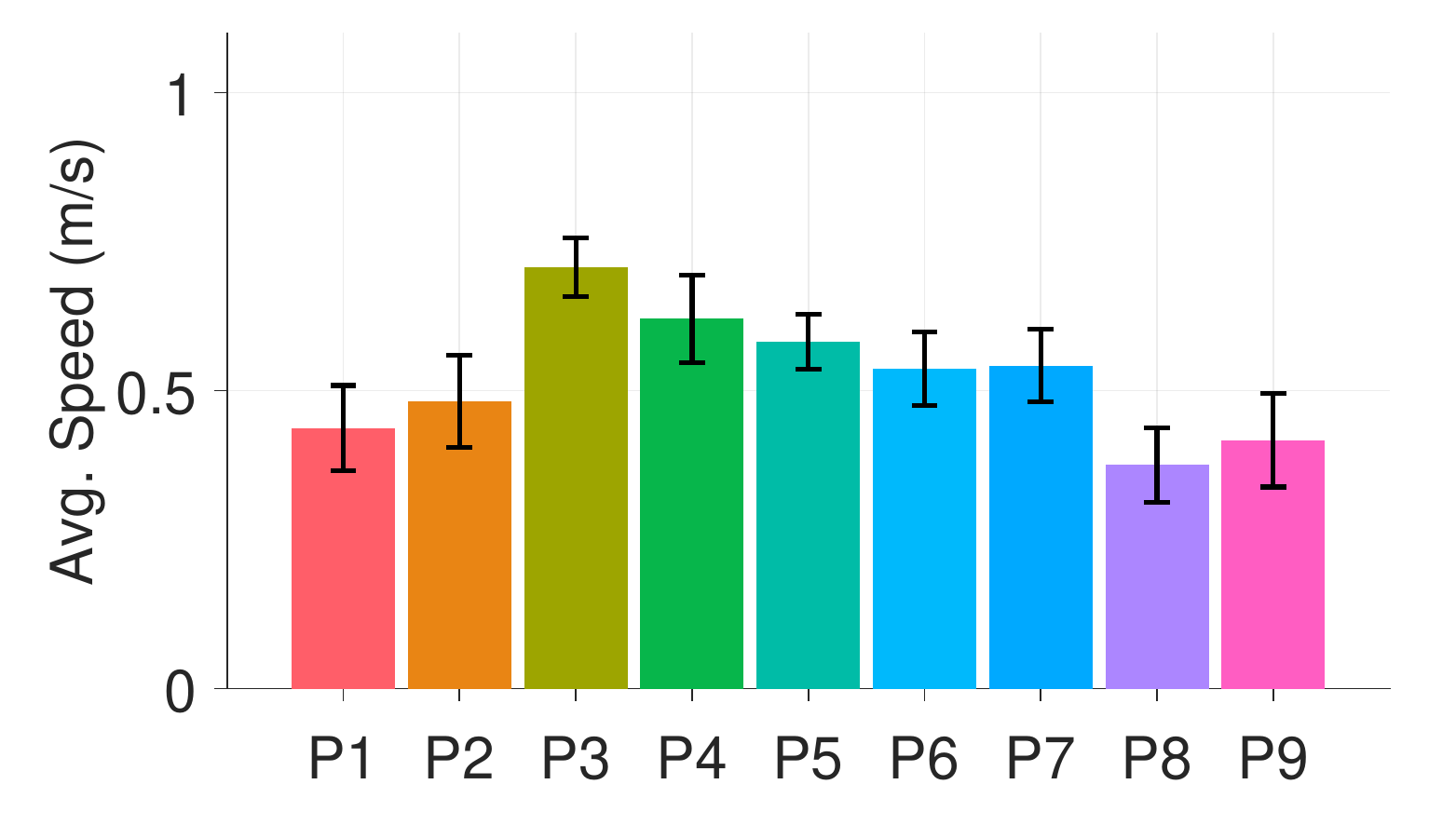} 
   &
   \includegraphics[width=1.5in]{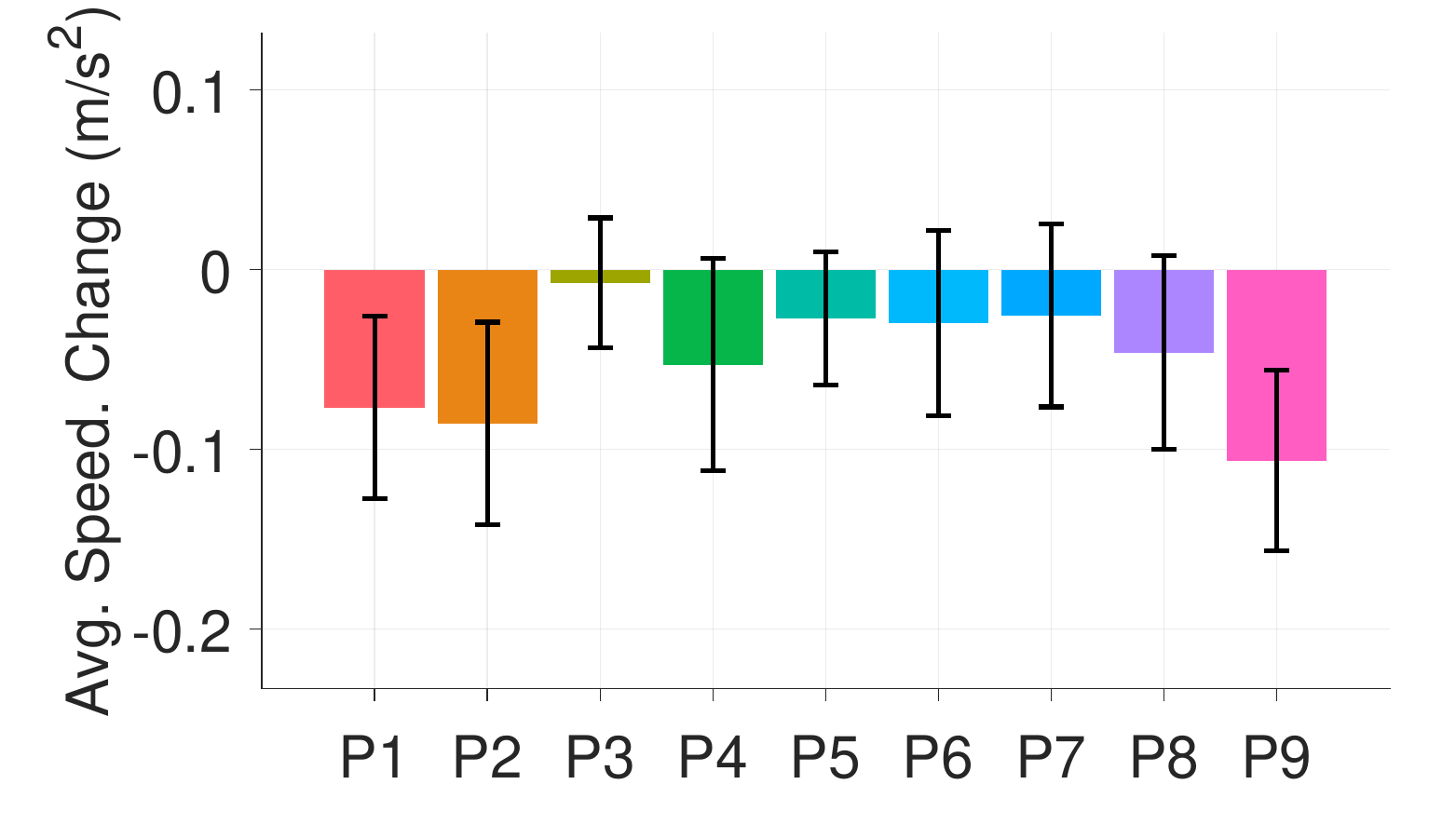} 
   &
   \includegraphics[width=1.5in]{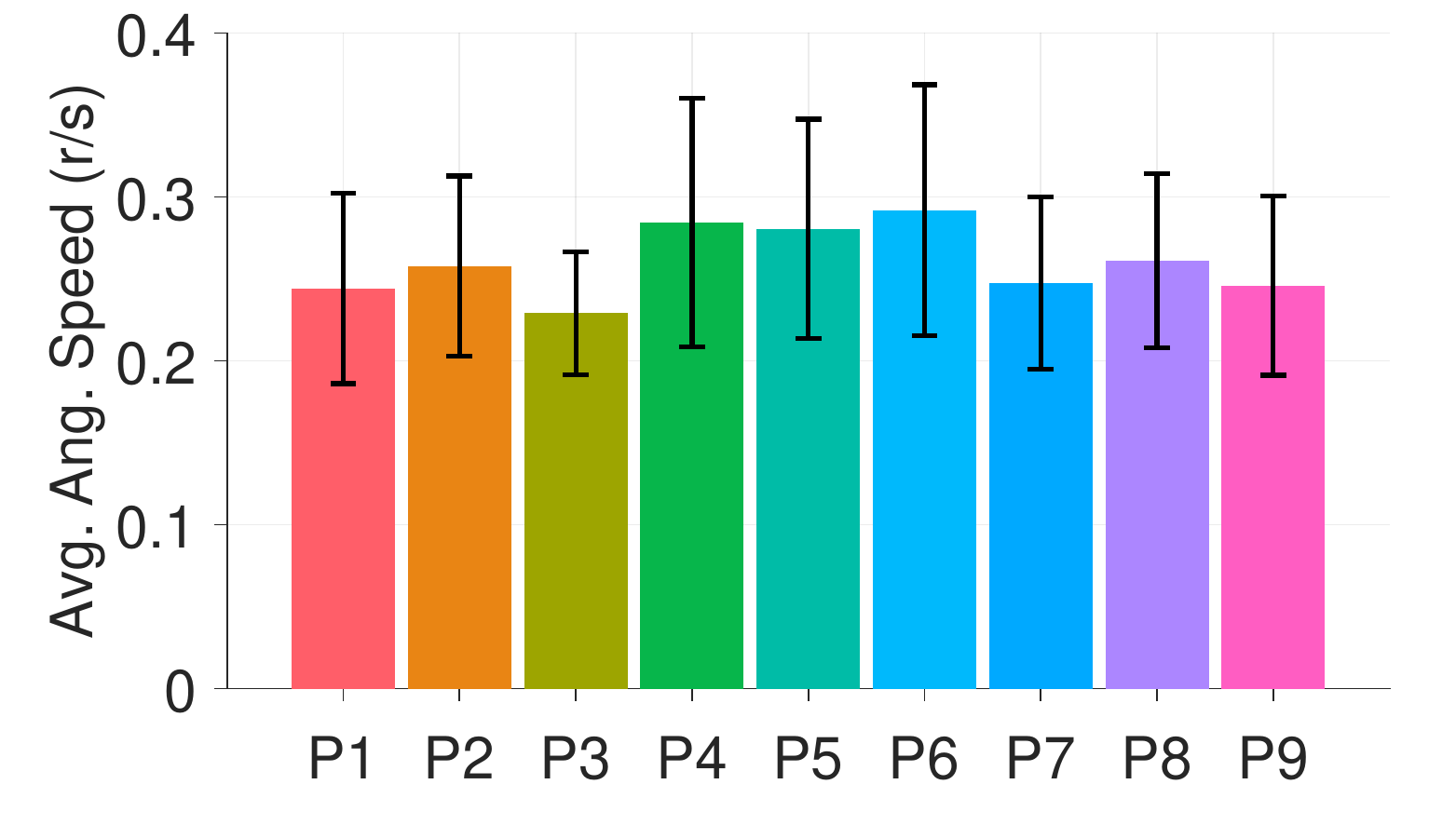} 
 \\
 (a) & (b) & (c) \\
 \multicolumn{3}{c}{\textbf{Small Turns}}
   \\
   \includegraphics[width=1.5in]{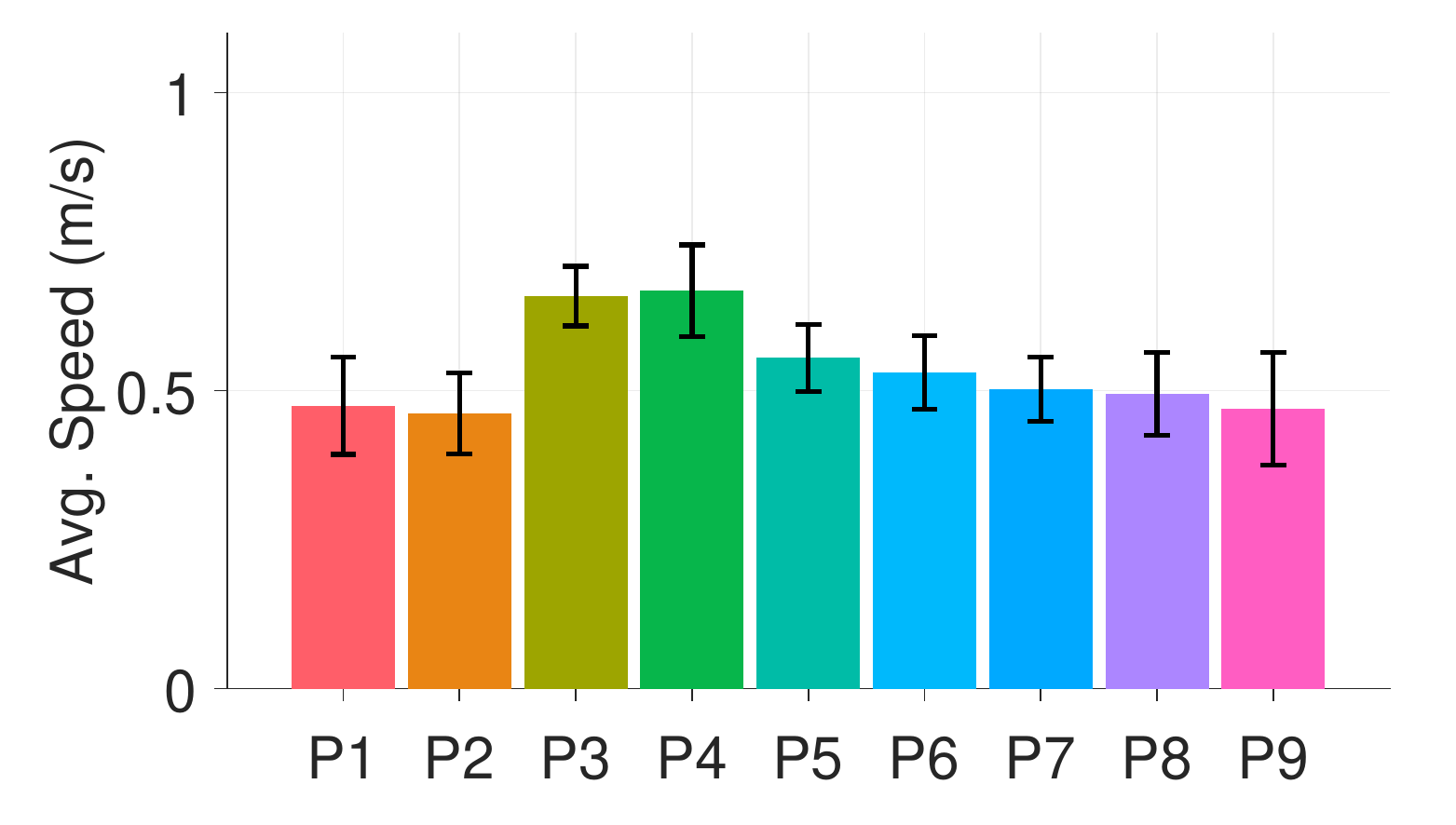} 
   &
   \includegraphics[width=1.5in]{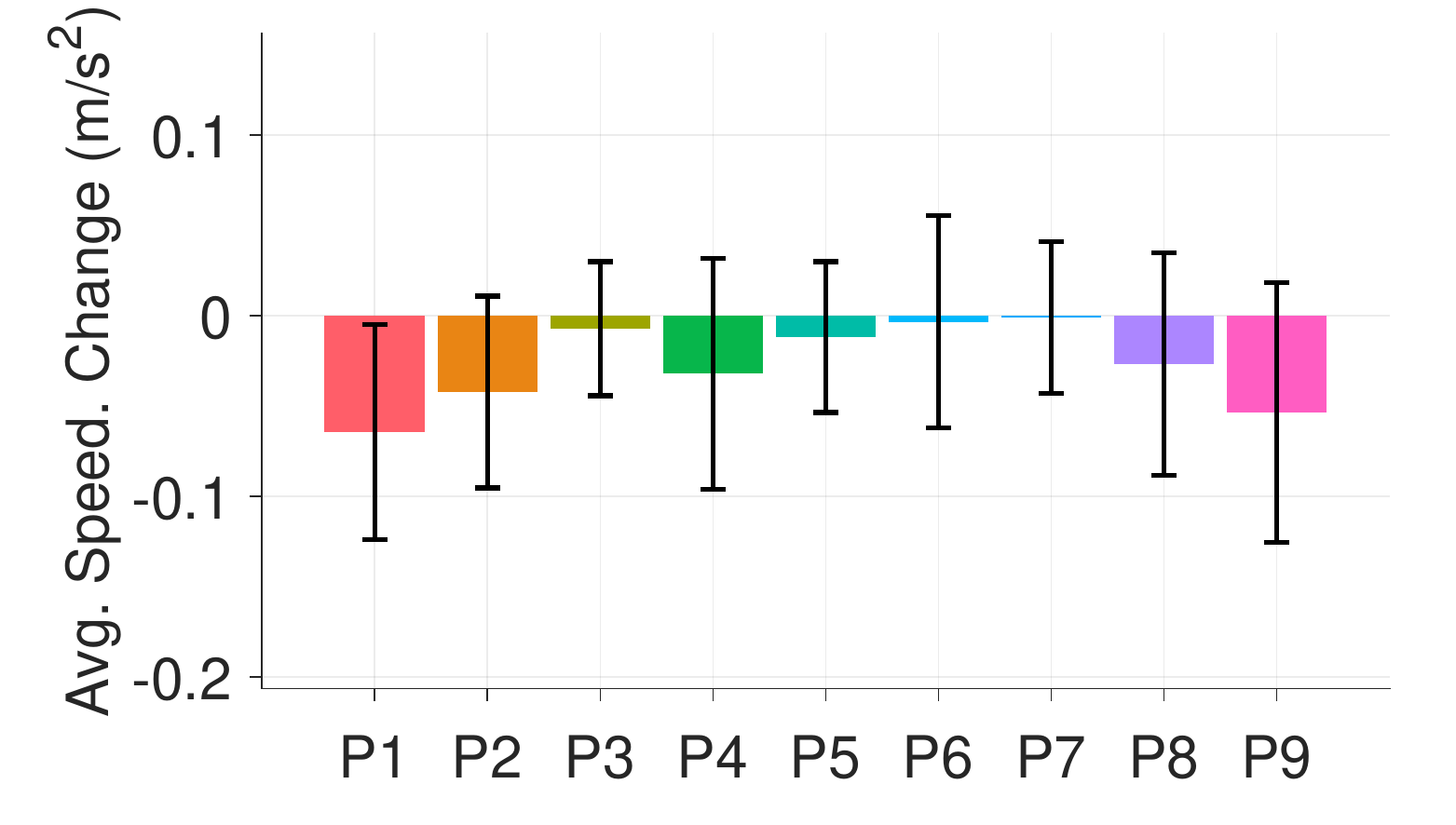} 
   &
   \includegraphics[width=1.5in]{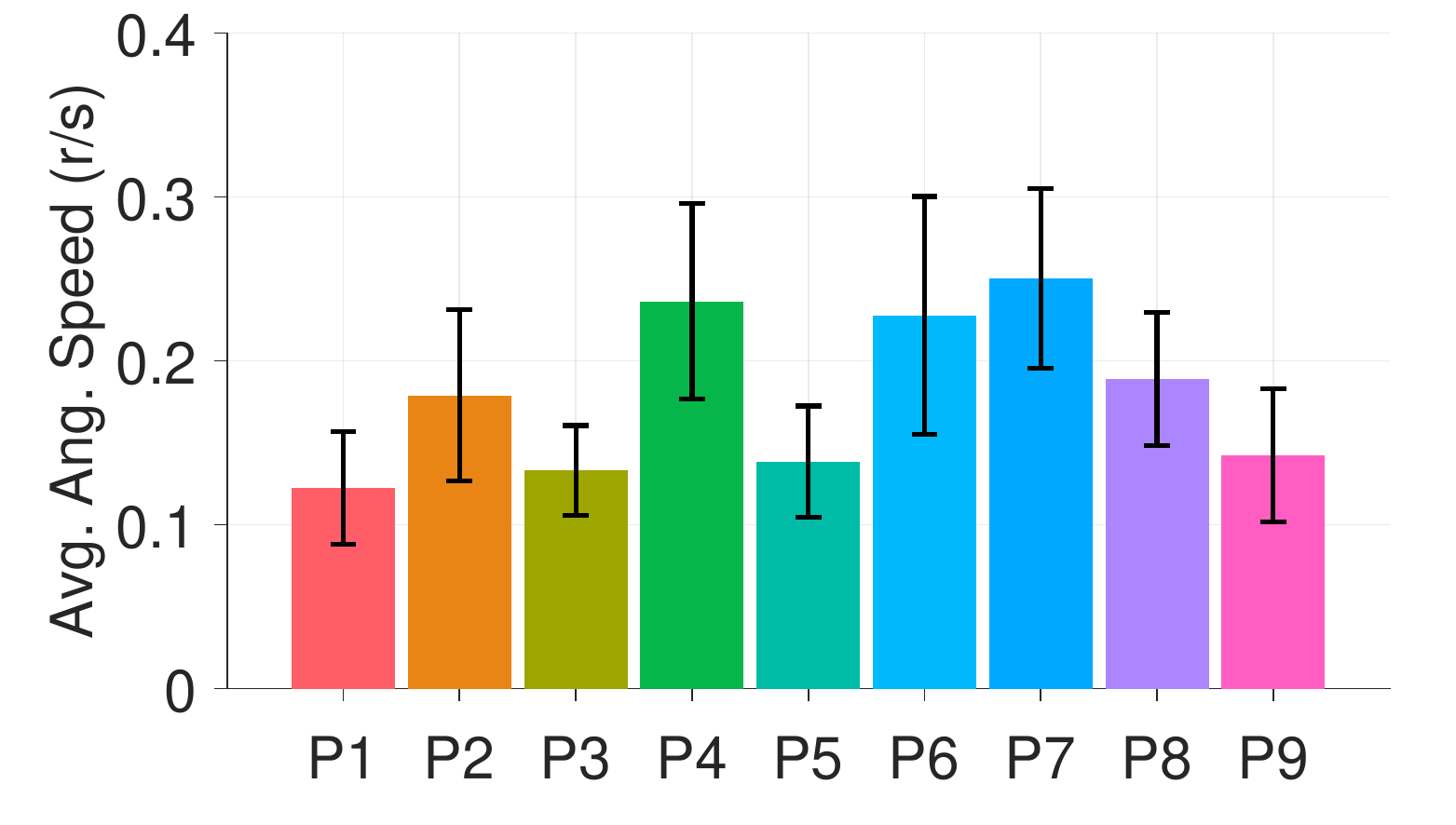} 
   \\
   (d) & (e) & (f)
    \end{tabular}
   \caption{\textbf{Impact of instructions on user's speed.} Comparison of (a) average speed (over a 6 second window) following the onset of instructional guidance, (b) speed change following the onset of instructional guidance (difference between end speed and starting speed divided by the time change of 6 second window), and (c) angular speed following different instructions, following \textbf{`large turn'} (90 degrees) and \textbf{`small turn'} (less than 60 degrees) instructions. Black bars show standard deviation. }~\label{fig:varturnssmallsupp}
  \end{figure}
  
   \begin{figure}[t]
  \centering
  \begin{tabular}{cc}
\textbf{Distance Statistics - Turns} &
\\
\includegraphics[width=3in,trim={0cm 0 0cm 0},clip]{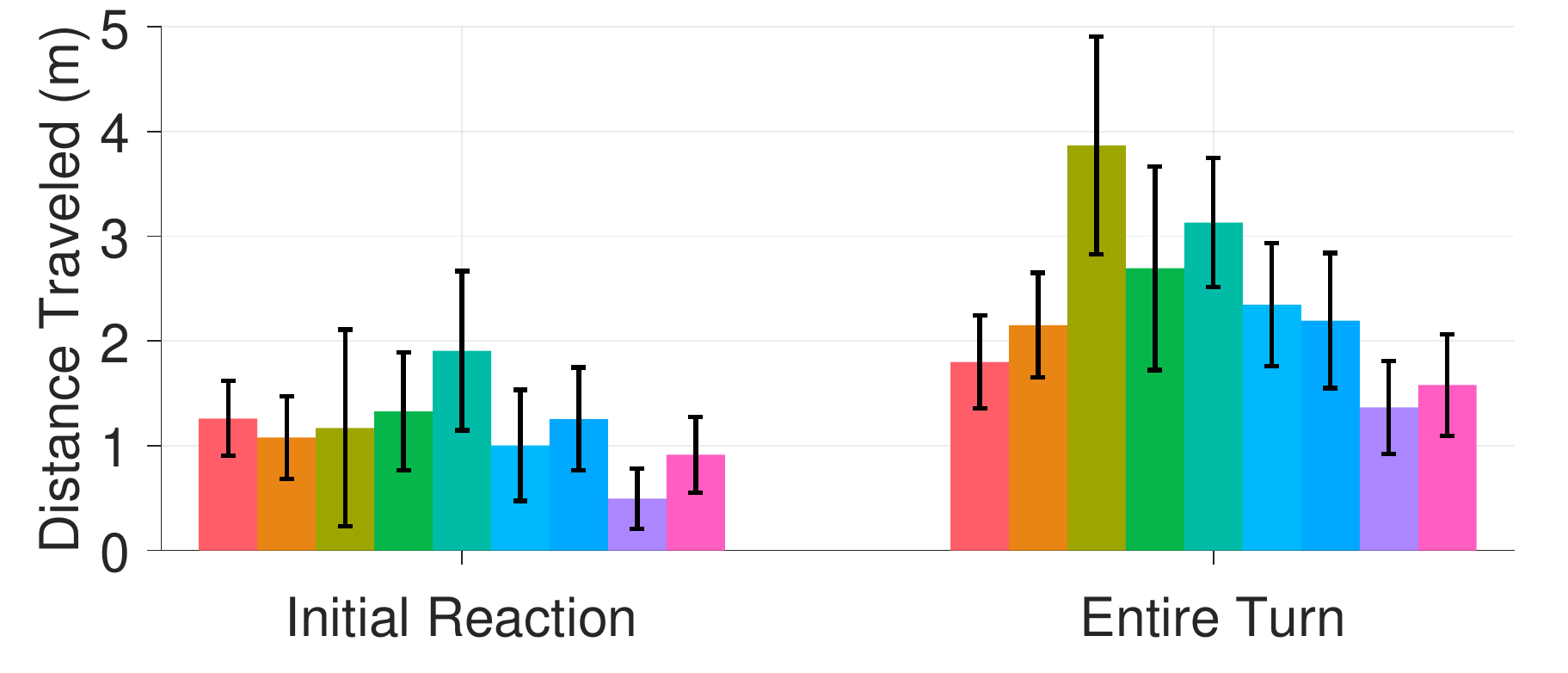} &    \includegraphics[width=0.2in,trim={0cm 2cm 13cm 0},clip]{fig/legp.pdf}
    \end{tabular}
   \caption{To better understand the relationship between user motion, instruction timing, and navigation errors, we plot the distance traveled by users during two components of a turning task, (a) initial reaction (between instruction onset to turning motion onset), and (b) entire turn task (instruction onset to turning motion completion). We can see how adaptive timing of instructions can prevent navigation errors in situations of open space or multiple corridors, as the distance traveled during the entire turning task (or even the initial reaction before turning begins) can span several meters.}~\label{fig:distturns}
  \end{figure}

Variability in linear speed is significant both for large turns ($F(8,130) = 5.7, p<.001$) and small turns ($F(8,106) = 2.63, p=.011$). We find that the guide-dog user P4 is generally faster than other participants, but cane users (\emph{e.g.,} P3) can maintain a similar speed during turns, so that even when testing without the guide-dog user variability among participants is still significant ($p<.001$). Since reactions to `turn' notifications involve the largest reduction in speed among all instruction types, the statistical difference remains in average speed change ($p<.001$ for both large and small turns among users). The large spectrum includes P3, who tends to maintain speed at turns with only a slight overall reduction, and P1, P2, and P9, who reduce their speed significantly. Beyond just using a velocity-based timing of instructions, a user-specific model of motion reaction to `turn' notifications can potentially reduce the chance of missing a turn (especially considering inherent localization errors).    

\textbf{Significant Variability in Angular Speed During Small Turns.} In the case of large turns, we do not find the angular speed variability to be statistically significant among users ($p=.576$), yet it is significant for small turns ($p=.003$). This can be explained by the physically easier task of gauging orientation during a sharp turn (90 degrees) compared to diagonal or slighter turns. While all users were instructed and trained of the differences between turn types and their extent before the study onset, variations in small turning could easily result in over or under-turning, and consequently veering off the planned path. As a concrete example, we find that a diagonal turn in an open space area along the route caused P7, P4, and P5 to get off the path. In one instance in the dataset, P5 over-turns following an instruction to turn diagonally to the left, and veers off the path in an open space. Hence, we can see how considering personal characteristics in reactions to turn instructions can be useful for developing interfaces which better prevent user-related navigation errors. 

\textbf{Significant Variability in Timing.} We further analyze reaction and task completion statistics following turning instructions, also finding significant variability among users. Following a `turn' notification, task timing can be clearly identified between the onset of the turning action and its completion. As a result, a variety of timing statistics can be computed. Here, we analyze the reaction time, overall task performance, and reaction to the vibration and sound feedback at the end of the turn indicating a correct heading has been achieved (the feedback is purposed to be used by the user for proper re-orientation during the turn, yet we find reaction to it takes another 0.91s on average, SD=0.71).

 \begin{figure}[!t]
  \centering
  \begin{tabular}{c}
  \includegraphics[width=1.5in]{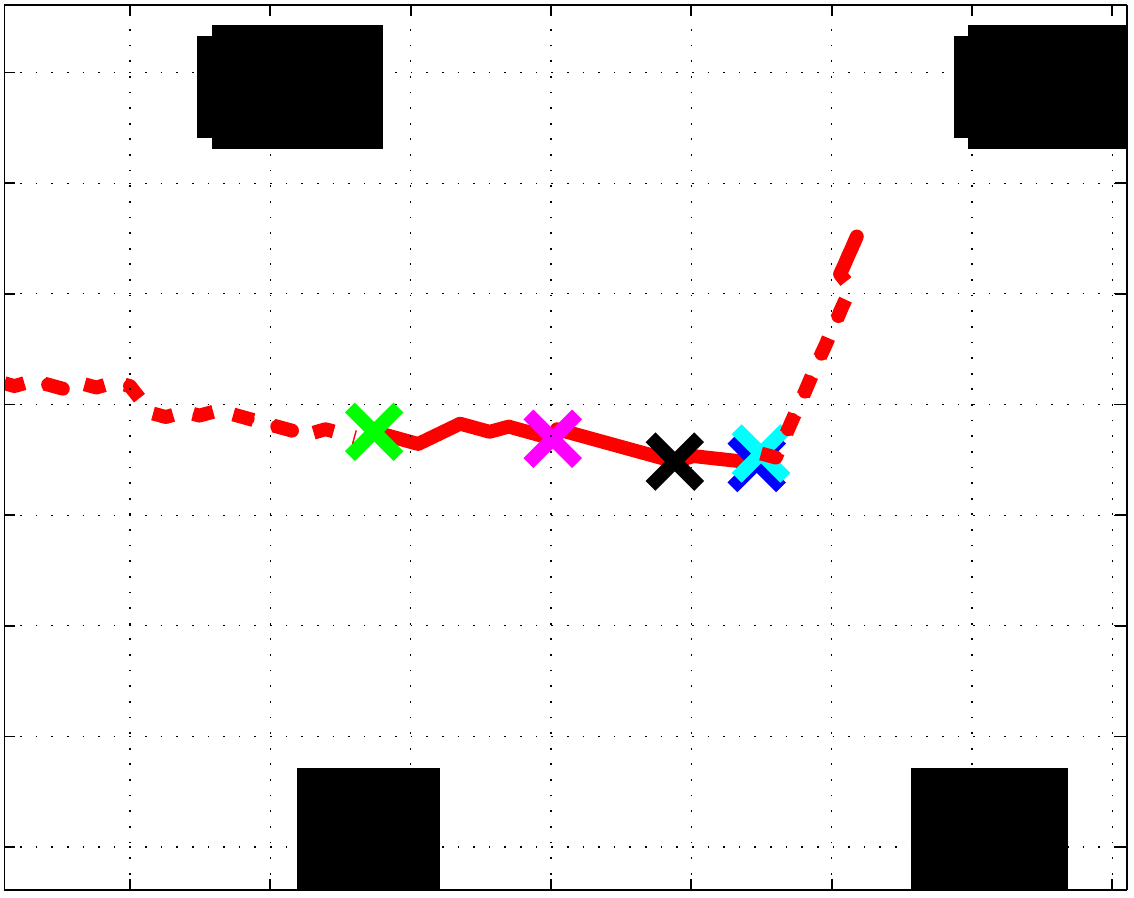} 
    \end{tabular}
  \caption{Example event definition for an instructed turn in the data. Grid lines are at $1$ meter intervals, and in black are obstacles/walls. Due to high variability in turns, we decompose turns by defining 5 temporal subevents, $I_S$ the instruction beginning time (in green), $I_E$ instructions end (pink), $T_S$, onset of turning by user (black), $I_H$ heading feedback from the interface (dark blue), and $T_E$ the end of the re-orientation by the user. Trajectory before and after the entire sequence is dashed. We then compute timing statistics between these events.  }~\label{fig:exampleturn}
  \end{figure}

  \begin{figure}[!t]
  \centering
  \begin{tabular}{ccc}
    \multicolumn{3}{c}{\includegraphics[width=4in]{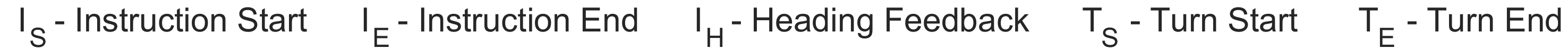}} \\
  \includegraphics[width=1.5in]{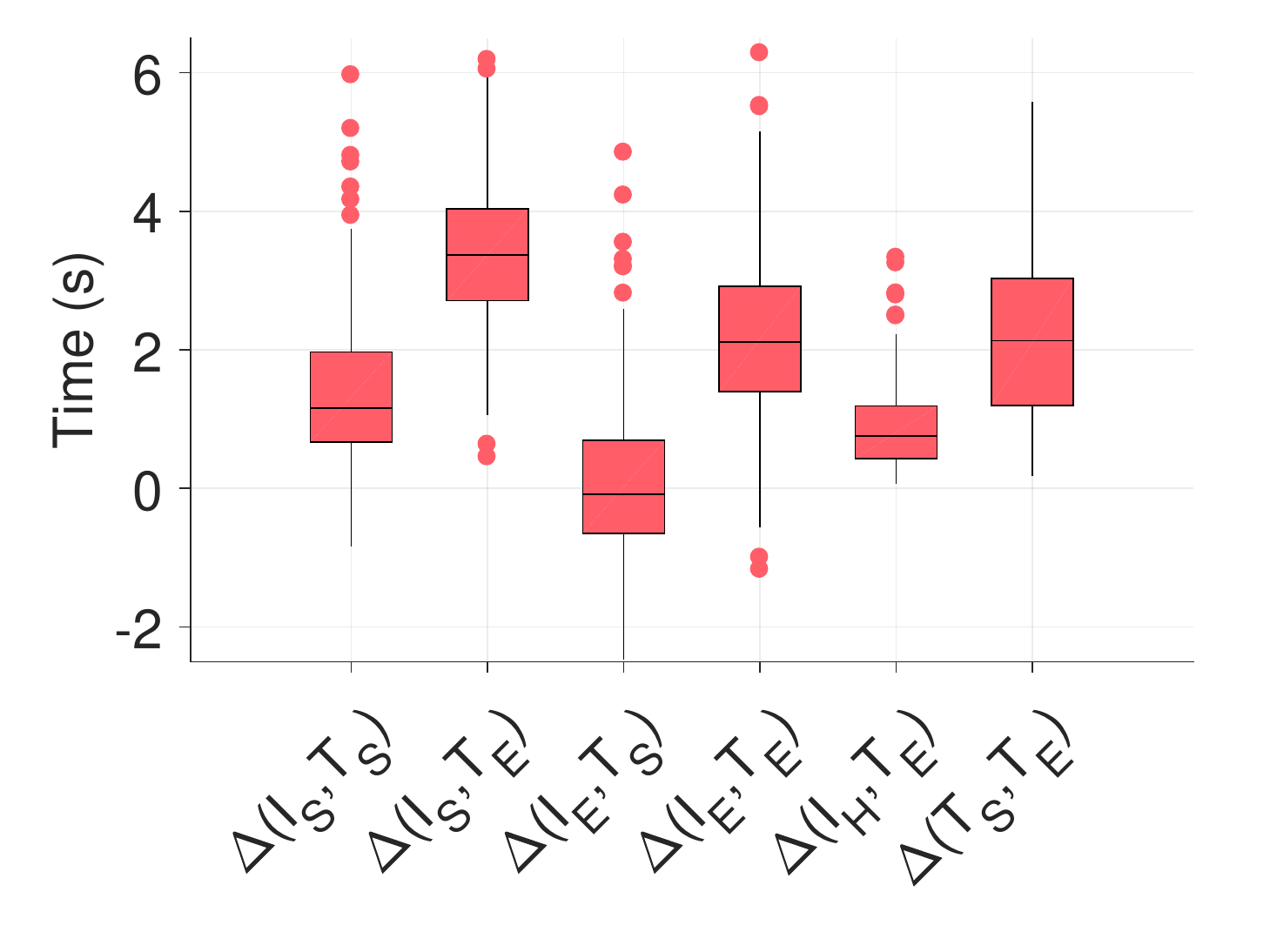}&
  \includegraphics[width=1.5in]{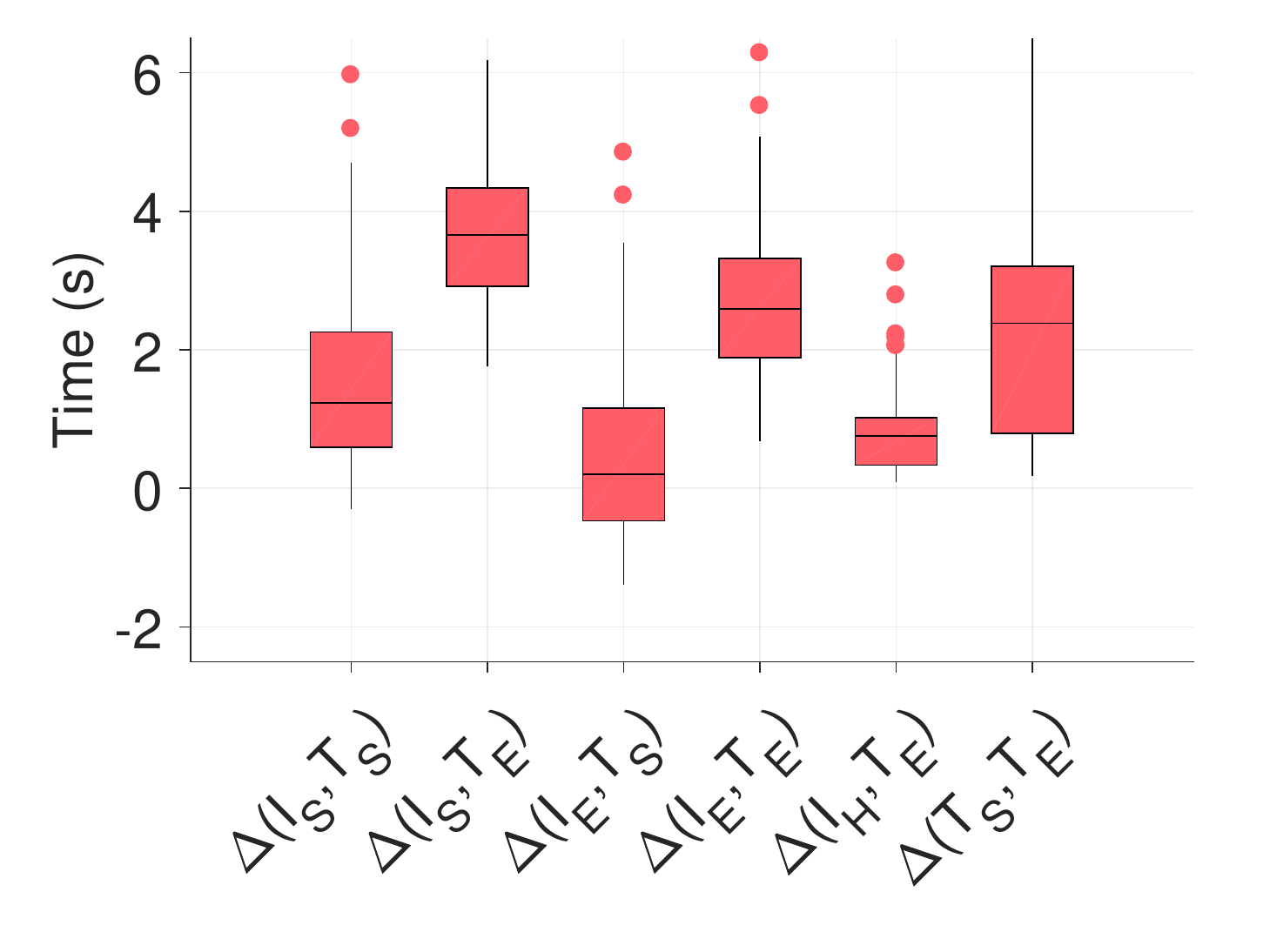}&
  \includegraphics[width=1.5in]{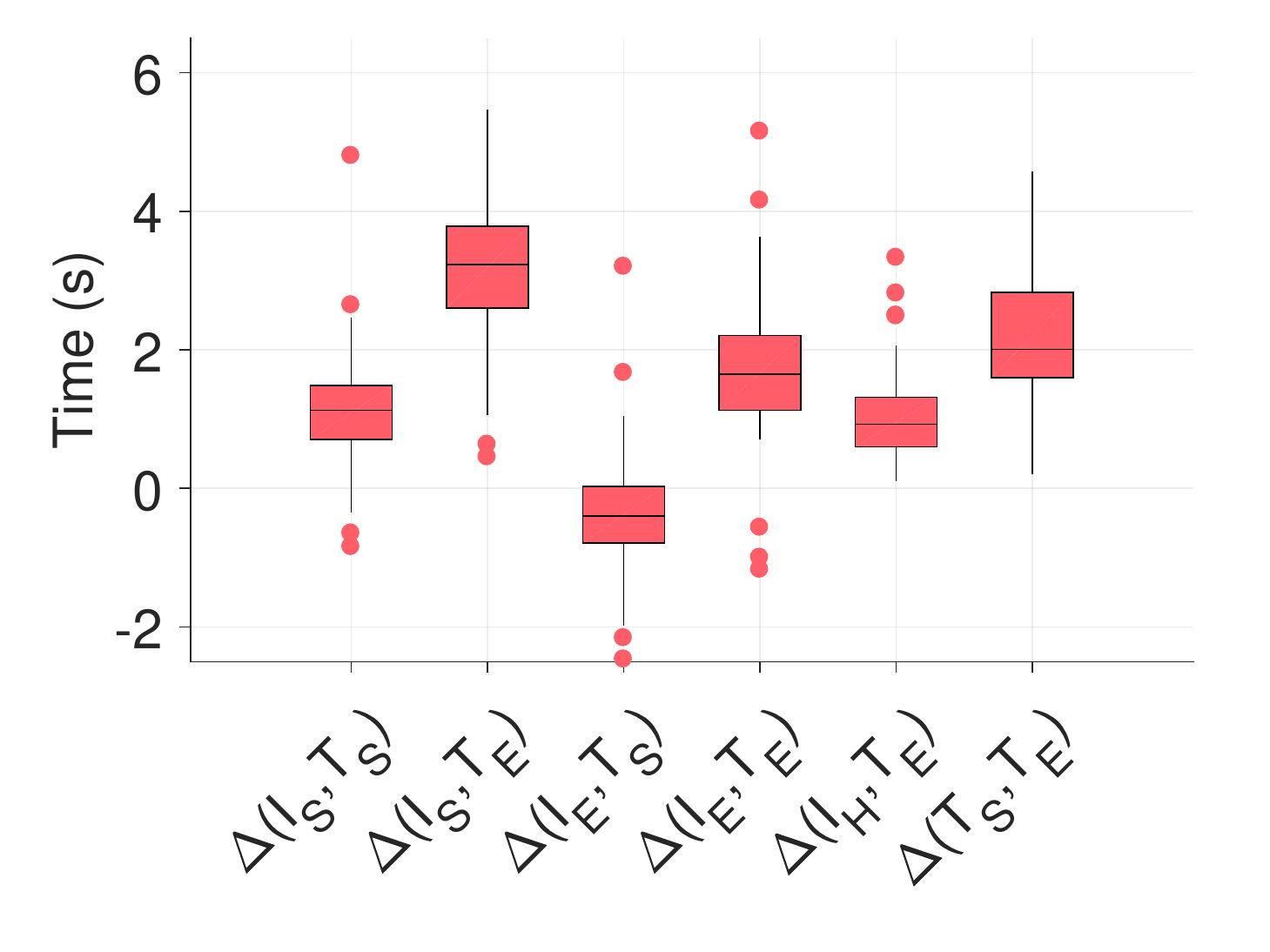} \\
  (a) Overall & (b) Large & (c) Small (Diagonal/Slight)
    \end{tabular}
  \caption{Boxplots showing overall (across all users) statistics for timing of events during different turn types. $\Delta$ stands for time difference in seconds, \emph{i.e.,} $\Delta ( I_S , T_S ) = T_S - I_S$. }~\label{fig:TLTRallstats}
\end{figure}

 \begin{figure}[!t]
  \centering
  \begin{tabular}{cc}
  \multicolumn{2}{c}{\includegraphics[width=5in]{fig/timingTurnsLegend.pdf}}\\
  \includegraphics[width=5in]{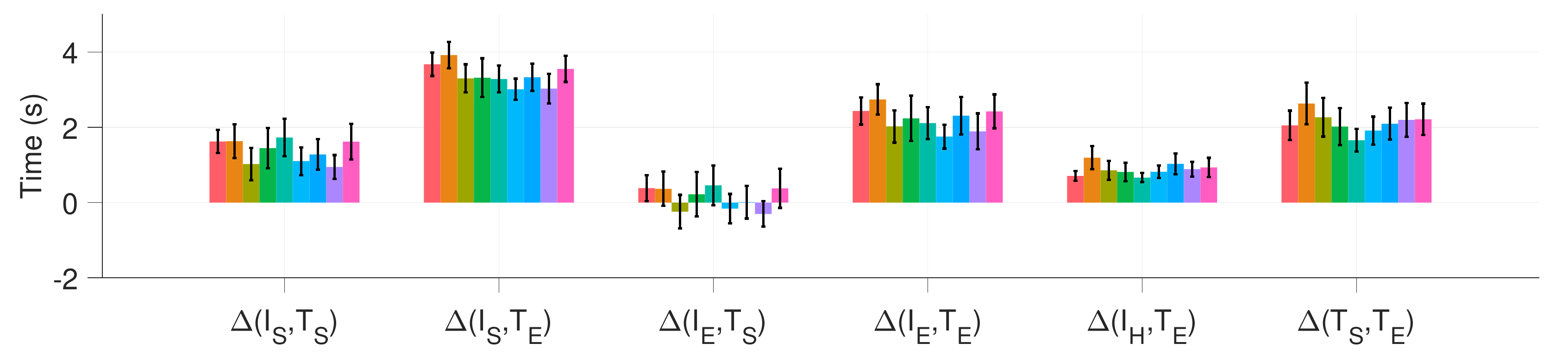} & \\
 \multicolumn{2}{c}{ (a) All turns}
 \\
 \includegraphics[width=5in]{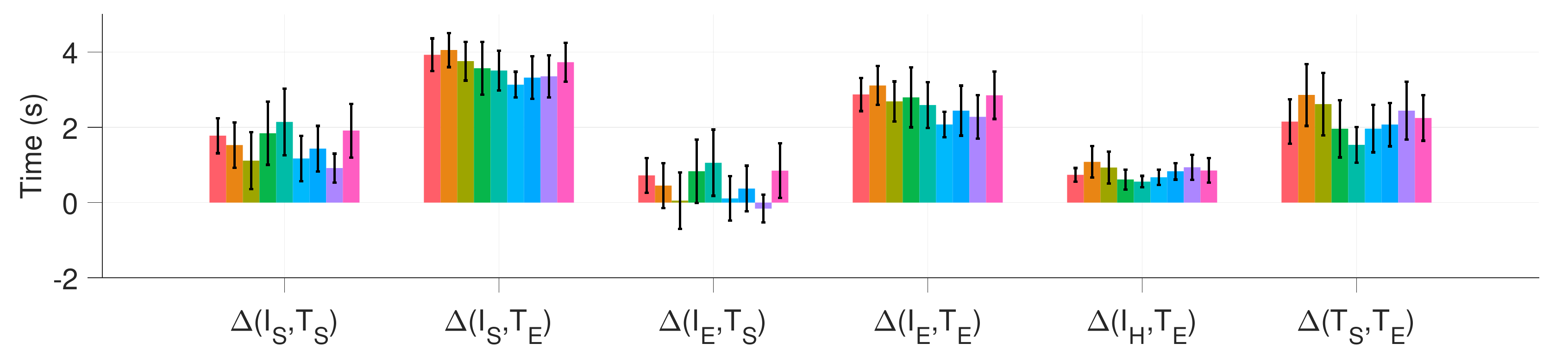} &  \includegraphics[width=0.3in,trim={0cm 2cm 13cm 0},clip]{fig/legp.pdf}  \\
  \multicolumn{2}{c}{(b) Large turns}
 \\
   \includegraphics[width=5in]{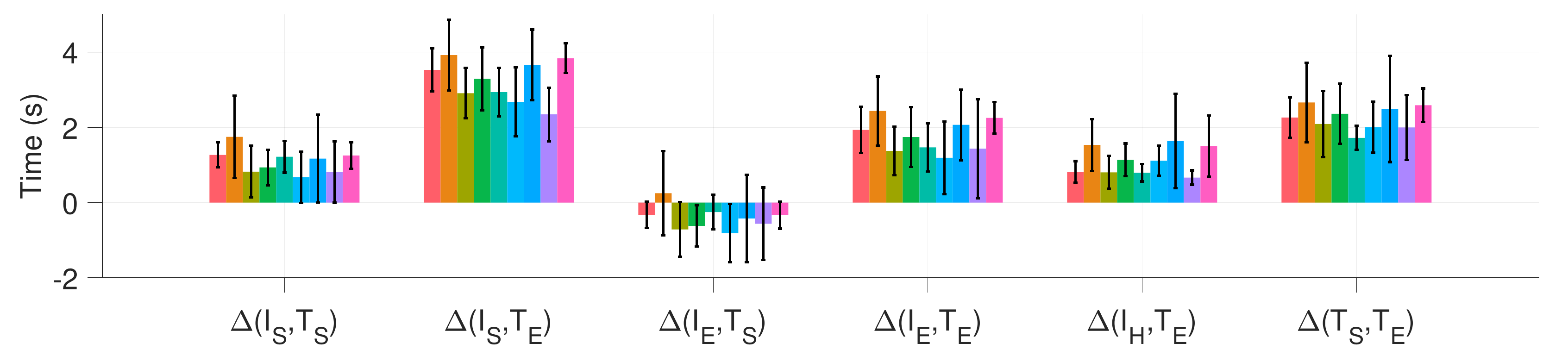} & \\
 \multicolumn{2}{c}{ (c) Small turns }
 
    \end{tabular}
  \caption{Per-user breakdown of turn timing statistics. }~\label{fig:ext1}
  \end{figure}

The reaction time between onset of instruction and beginning of the turning action is shown to be highly variable among users ($F(8,236)=2.15, p=.032$). While the average reaction time (average 1.37s, SD=1.08) is close to the one reported in related literature~\cite{adebiyi2017assessment}, other studies are often performed in controlled settings with short routes, thereby missing the crucial finding of significant variability among users. In contrast, we show it to be highly variable. We also find that the total time to complete this task may last over 4 seconds for some users (average 3.54s, SD=1.17), and is statistically significant among users ($p=.006$). 

\textbf{Is It Due to Age Differences?} One possible explanation for the variability could be users' age, but this was not confirmed by our data. While our user population is generally older (4 participants are under 46 years old), we attempted to group participants into two groups using several age group definitions, but found no statistically significant differences with respect to reaction time or overall task time. While we found older users (over vs. under 46 years old) to move slower towards the end of the turn ($p=.007$), which is somewhat to be expected, it does not fully explain the variability in turn timing. Additional data is required for this type of analysis.  

\textbf{Large Variability in Mobility Skills \& Strategies During Turns.} Instead, we are able to explain this phenomenon by inspecting each user's mobility during the navigation. Specifically, some users are able to better anticipate upcoming turns by leveraging their mobility skills. The insight came by inspecting the video data, in particular between the fastest `turner' P3 compared to a slower one, such as P1. P3 is able to use several navigation strategies to achieve correct and quick turning. Since users are informed of the upcoming navigation turning point and turn direction after completion of the previous turn, P3 constantly checks for an available turn with the cane in the notified direction (only with the cane, while still walking forward). Hence, the moment a turn is available, P3 immediately begins turning, even before the turn notification completes. On the other hand, P5 appears to heavily depend on the instructional guidance while walking in the middle of the hallway. P1 does not employ a similar strategy, and simply awaits the next instruction while using the cane to avoid obstacles (but not for anticipating or seeking turns). Also, P1 pauses before turning, and turns very cautiously as if hesitating before acting due to a notification. Inspecting videos with other users, we found each to be on a spectrum, between the behavior of P3 and P1 in terms of mobility strategies. Hence, we conclude that personal differences in mobility skill and style help explain variability in reaction to turn notifications.   

\textbf{Timing Variability Among Turn Types.} Timing results can be further analyzed by the turn type (small vs. large). When comparing among turn types in terms of timing, we observe more inter-user variability occurring during the large turns as opposed to small turns. Particularly relevant to our analysis is the finding that some users treat small turns with less care on a consistent basis. In the dataset, several users which over/under-turn got off the planned path in an open space area. 
  \subsubsection{All Instructions - User Variability Analysis}
\begin{figure}[!t]
   \centering
   \begin{tabular}{cccc}

\multicolumn{4}{c}{\textbf{User Motion Statistics - Overall}} \\
\includegraphics[width=1.55in]{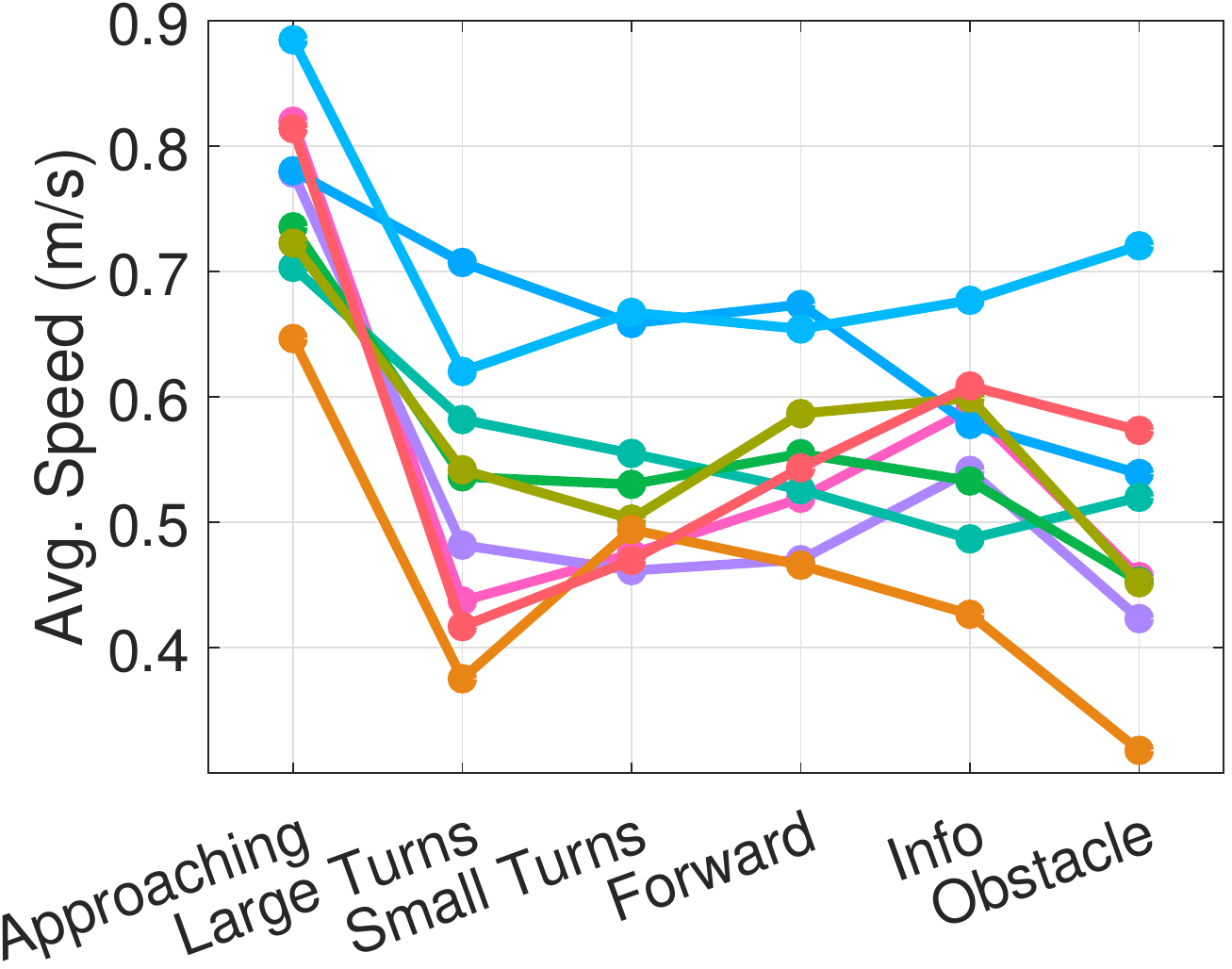}
&
\includegraphics[width=1.55in]{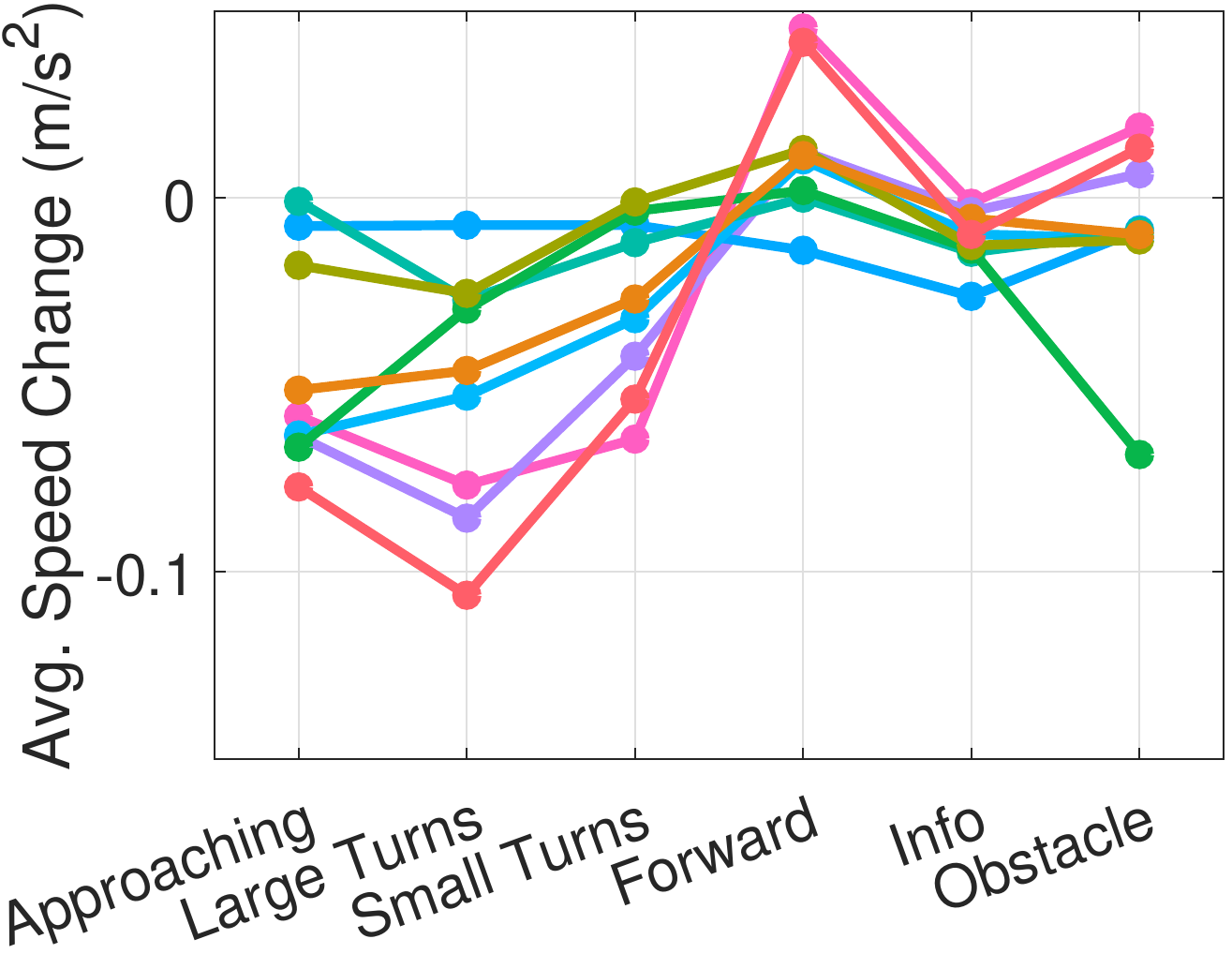}
&
\includegraphics[width=1.55in]{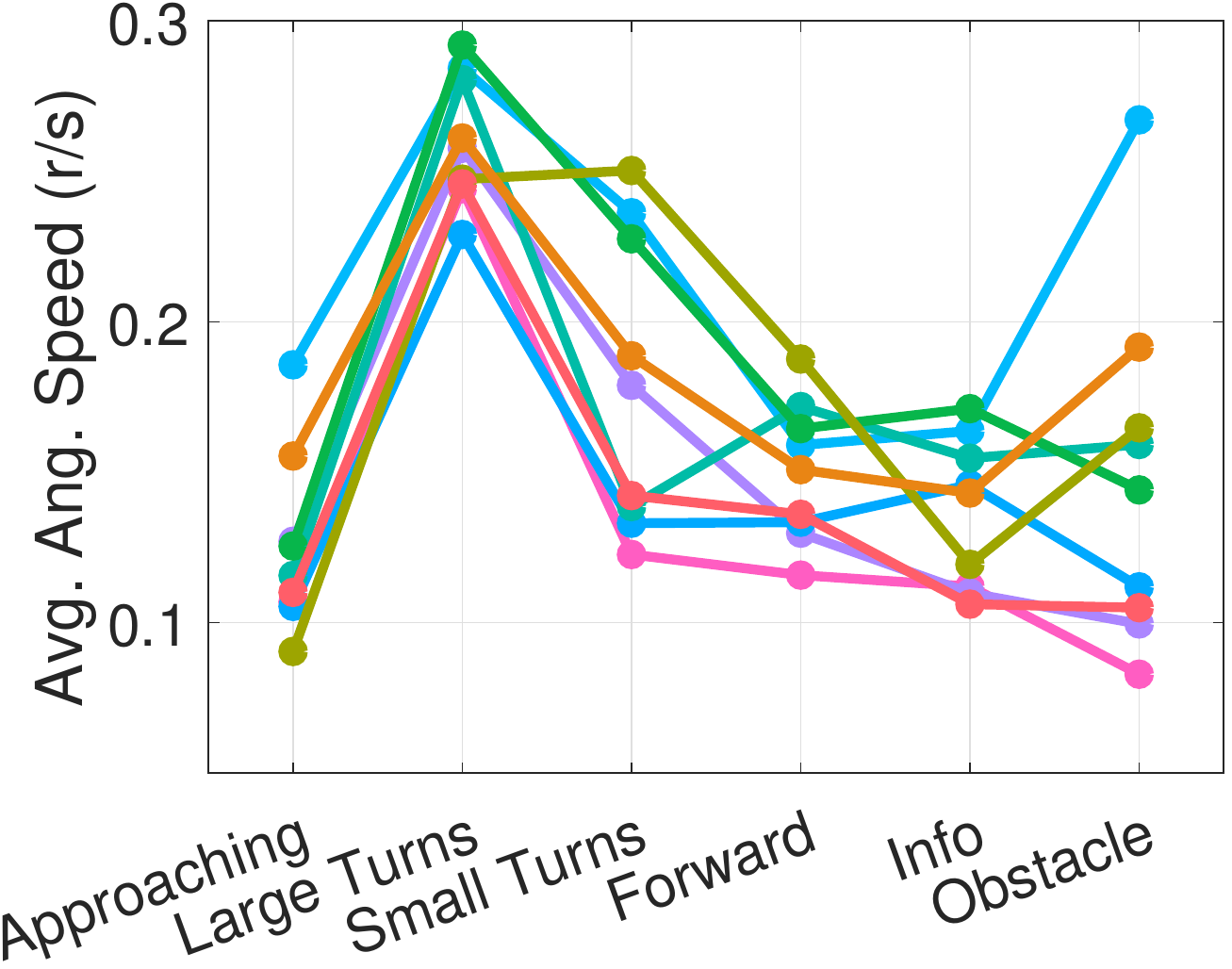}
& 
\includegraphics[width=0.24in,trim={0cm 2cm 13cm 0},clip]{fig/legp.pdf}
\\
(a) & (b) & (c) &

     \end{tabular}
   \caption{\textbf{Individual differences among users when responding to assistive guidance.} (a) Average speed (over a 6 second window) following the onset of instructional guidance, (b) speed change following the onset of instructional guidance (difference between end speed and starting speed divided by the time change of 6 second window), and (c) angular speed following different instructions. }~\label{fig:fig1newsupp}
   \end{figure}

As most navigation errors in the dataset occur around turns, we focus our analysis on variability followwing turn instructions. However, interaction under other instruction types besides turns also show significant inter-user variability, including `obstacle' notifications ($F(8,54)=.043, p=.042$), `forward' ($F(8,314)=2.02,p=.044$), `approaching' ($F(8,236)=2.8, p=.006$), and `info' ($F(8,333)=2.62, p=.009$), but generally to a lesser extent (and significance depends on the motion measure used). While turns contain the most variability (see Fig.~\ref{fig:fig1newsupp}) and non-linearity (posing a challenge to the dynamics model), we include analysis for all instructions in Fig.~\ref{fig:fig1newsupp} for completeness.

   \begin{figure}[!t]
  \centering
  \resizebox{7.5cm}{!}{
  \begin{tabular}{ccc}
      \includegraphics[width=1.5in,trim={0cm 0 2.8cm 0},clip]{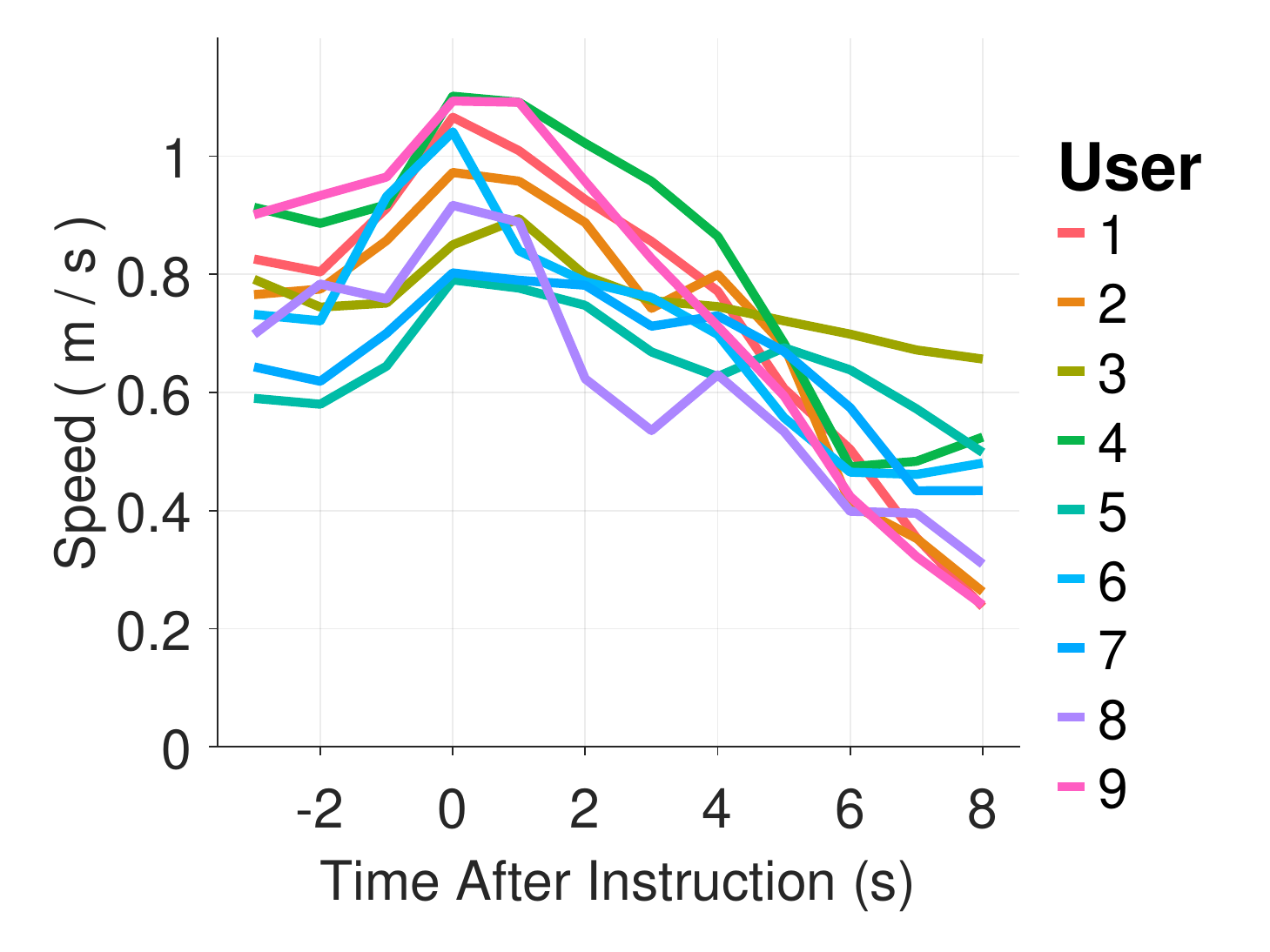} &
  \includegraphics[width=1.5in,trim={0cm 0 2.8cm 0},clip]{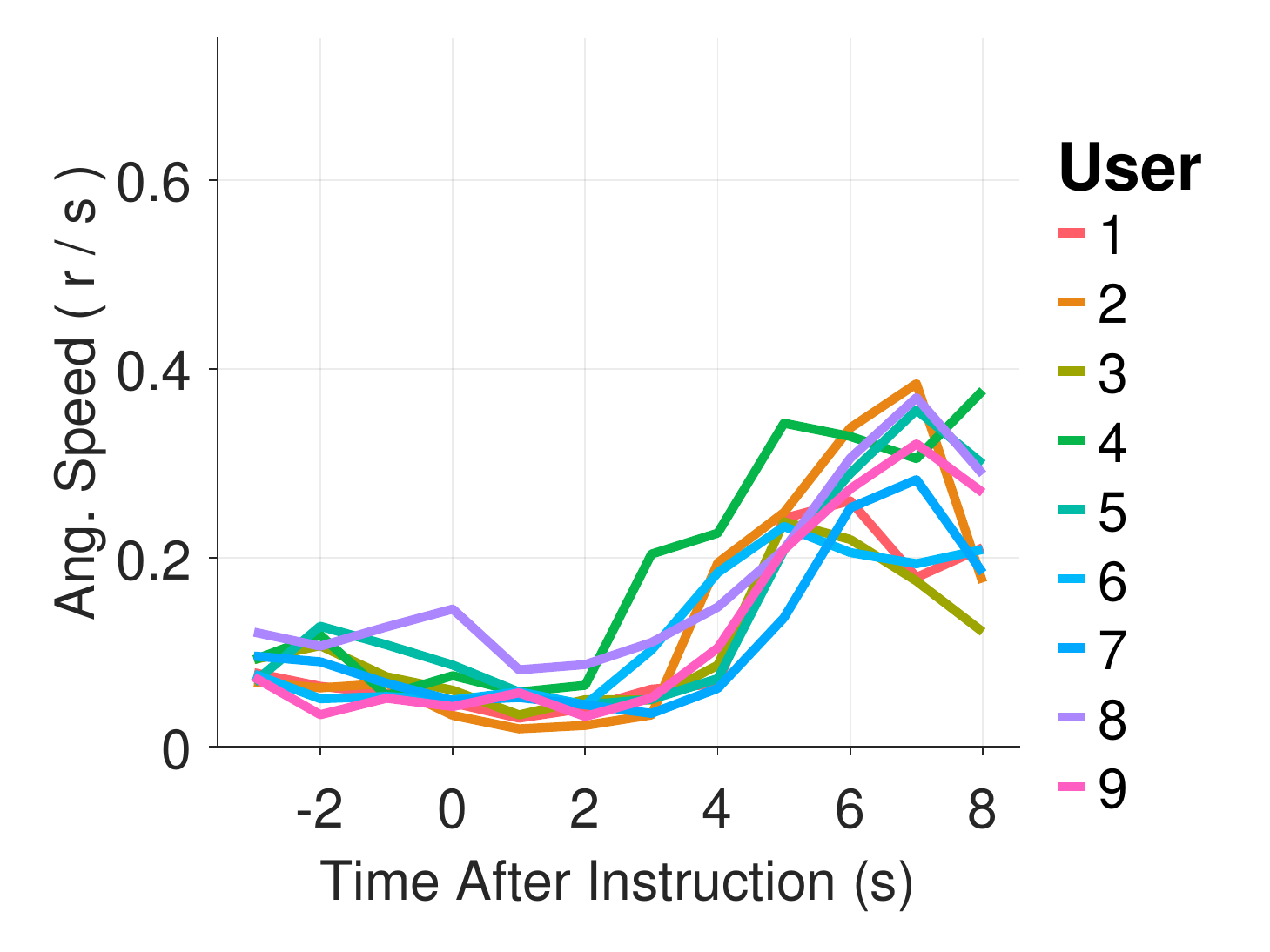} &  \includegraphics[width=0.2in,trim={0cm 2cm 13cm 0},clip]{fig/legp.pdf} 
  \\
    \multicolumn{3}{c}{(a) `Approaching' instructions}\\
    \includegraphics[width=1.5in,trim={0cm 0 2.8cm 0},clip]{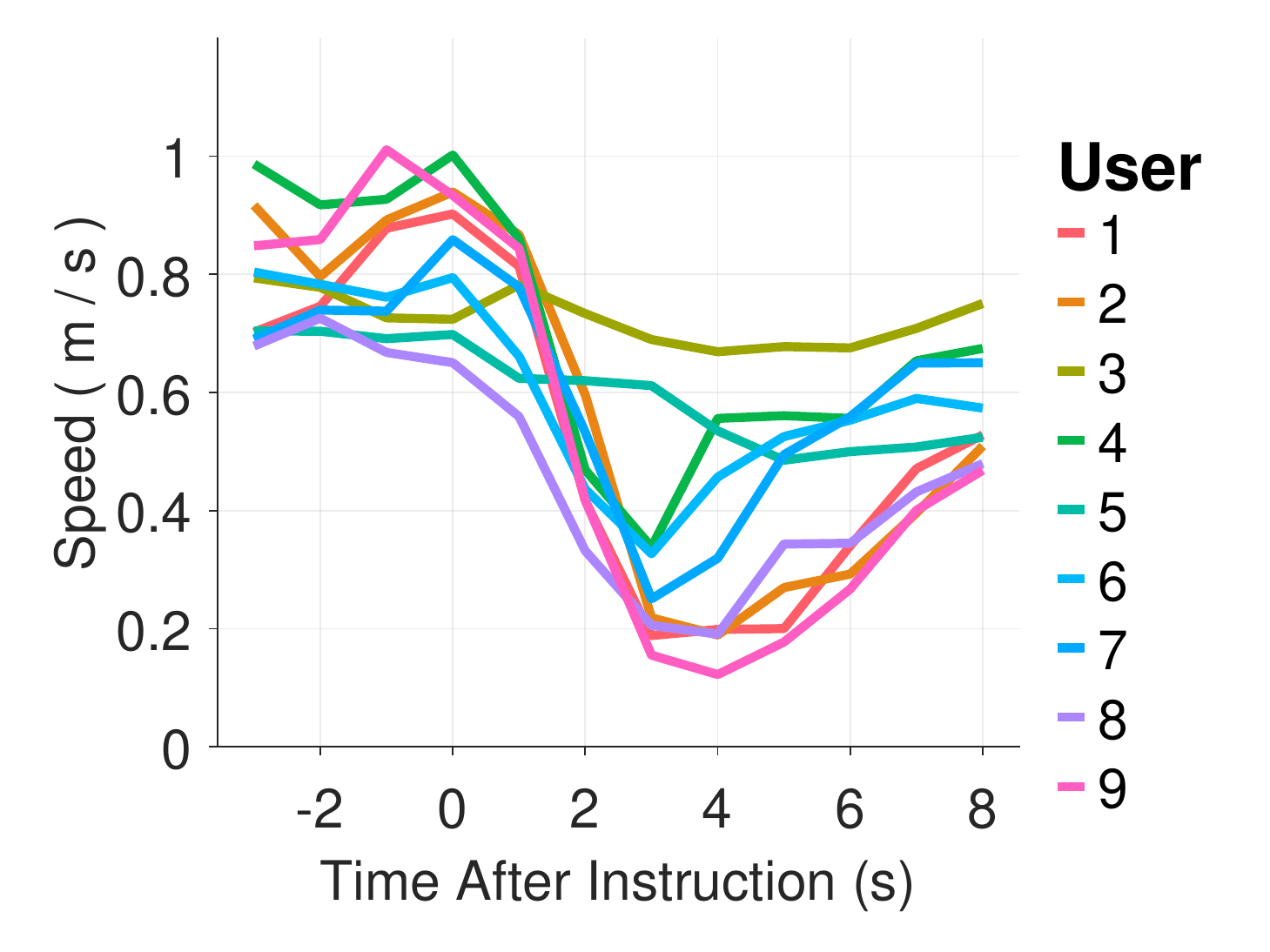} &
  \includegraphics[width=1.5in,trim={0cm 0 2.8cm 0},clip]{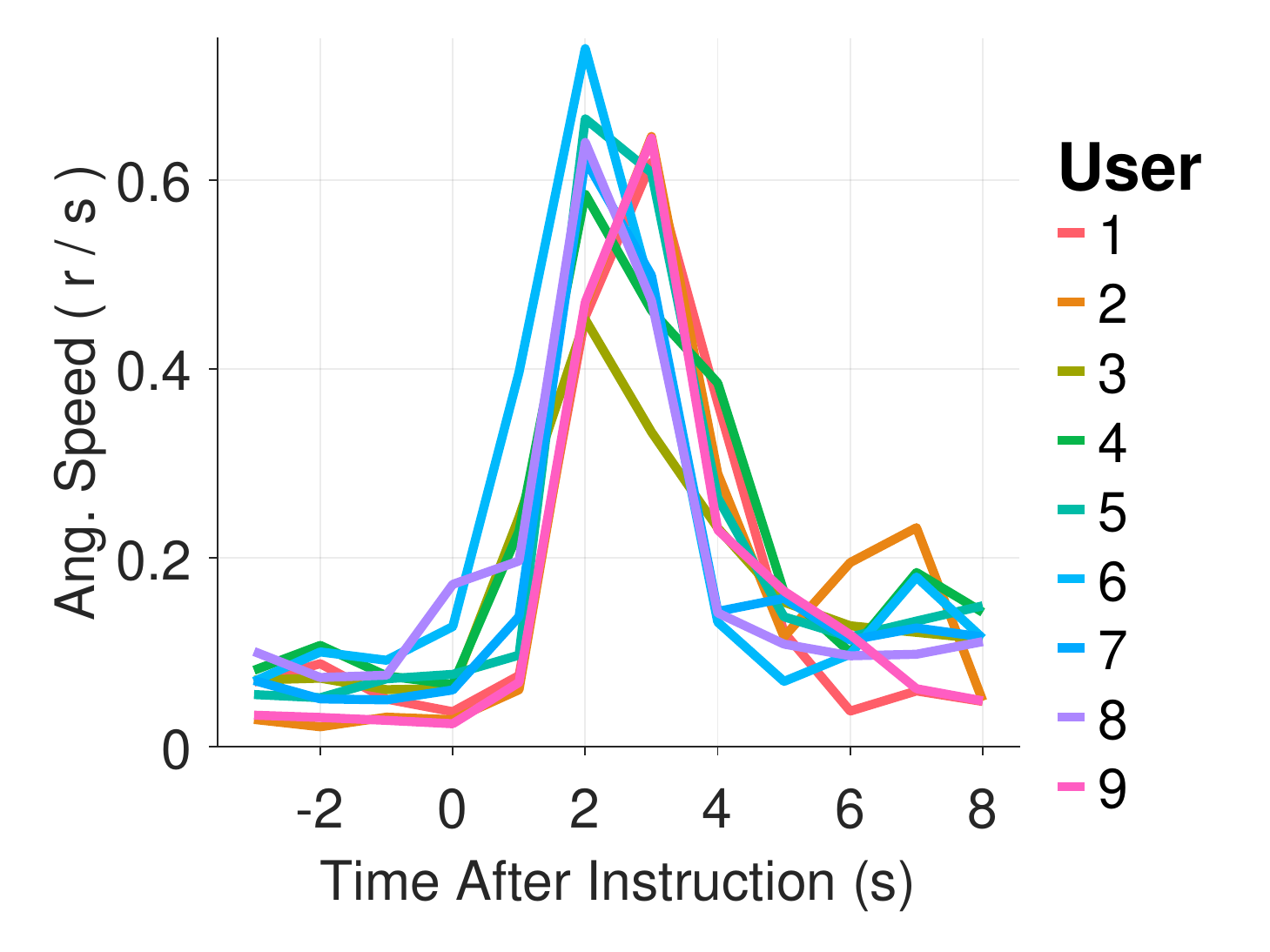} & 
    \includegraphics[width=0.2in,trim={0cm 2cm 13cm 0},clip]{fig/legp.pdf} 
  \\  \multicolumn{3}{c}{(b) `Large turn' instructions} \\
   \includegraphics[width=1.5in,trim={0cm 0 2.8cm 0},clip]{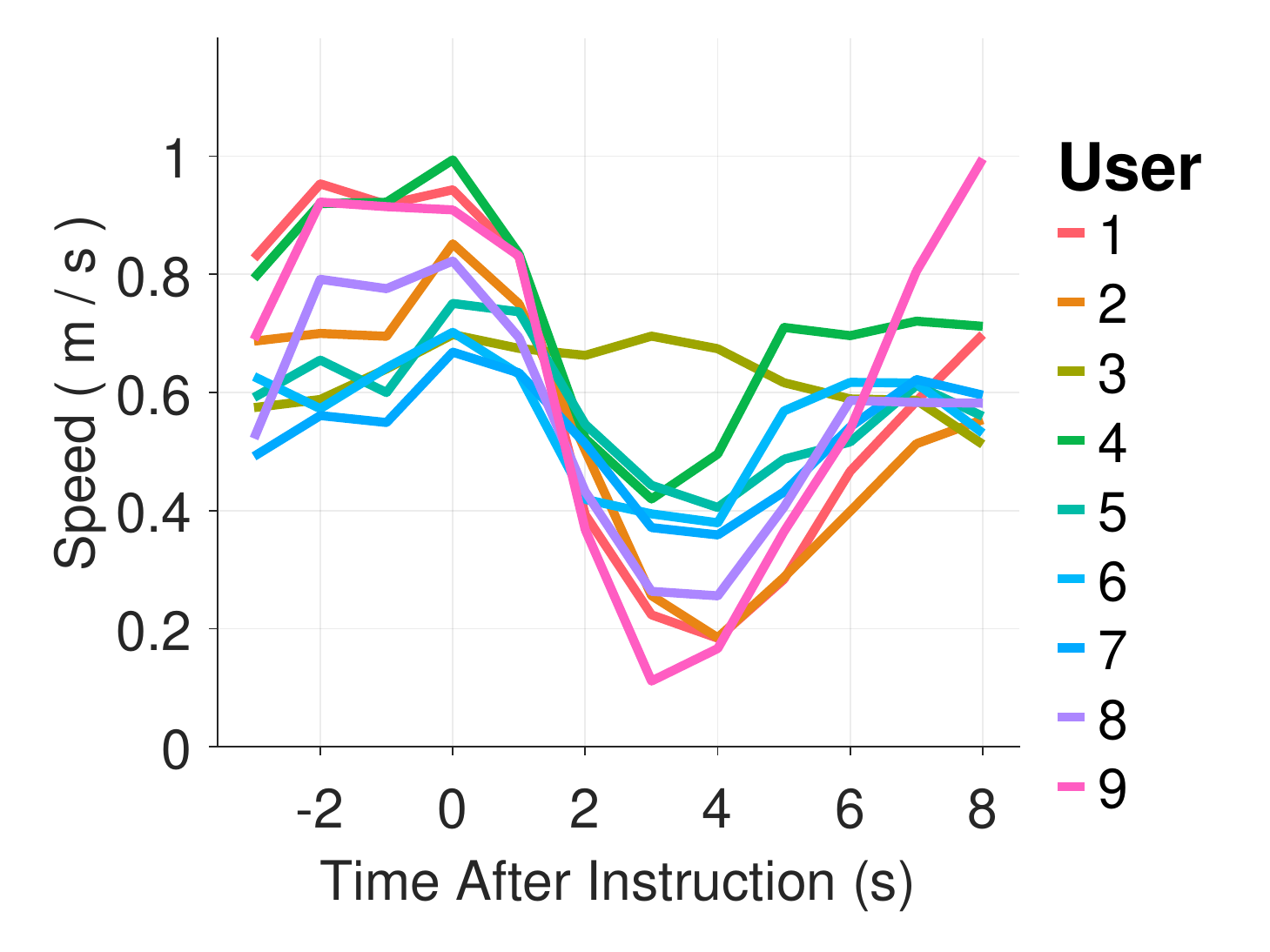} &
  \includegraphics[width=1.5in,trim={0cm 0 2.8cm 0},clip]{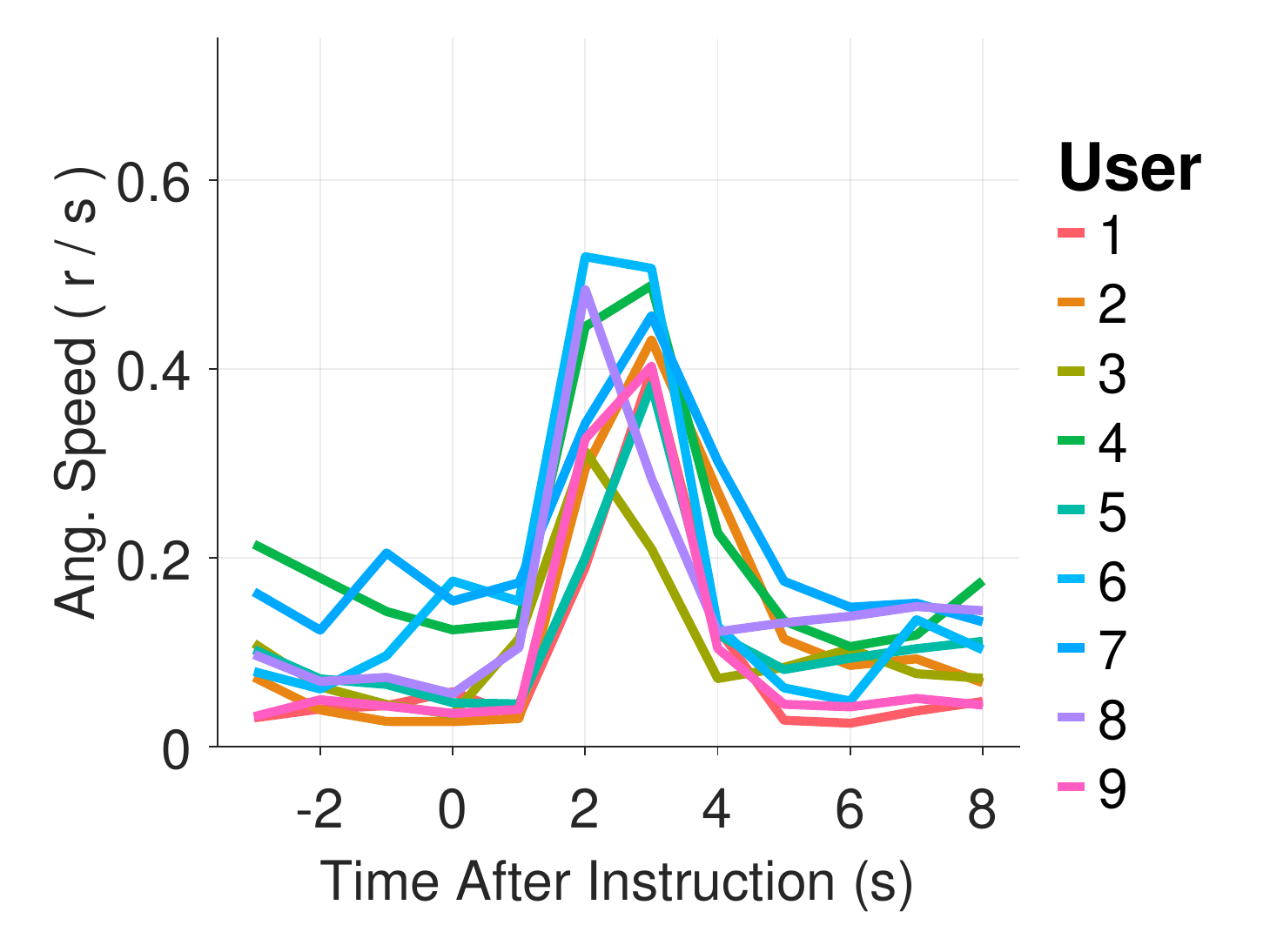} & 
    \includegraphics[width=0.2in,trim={0cm 2cm 13cm 0},clip]{fig/legp.pdf}  \\  \multicolumn{3}{c}{(c) `Small turn' instruction}
    \\
    \includegraphics[width=1.5in,trim={0cm 0 2.8cm 0},clip]{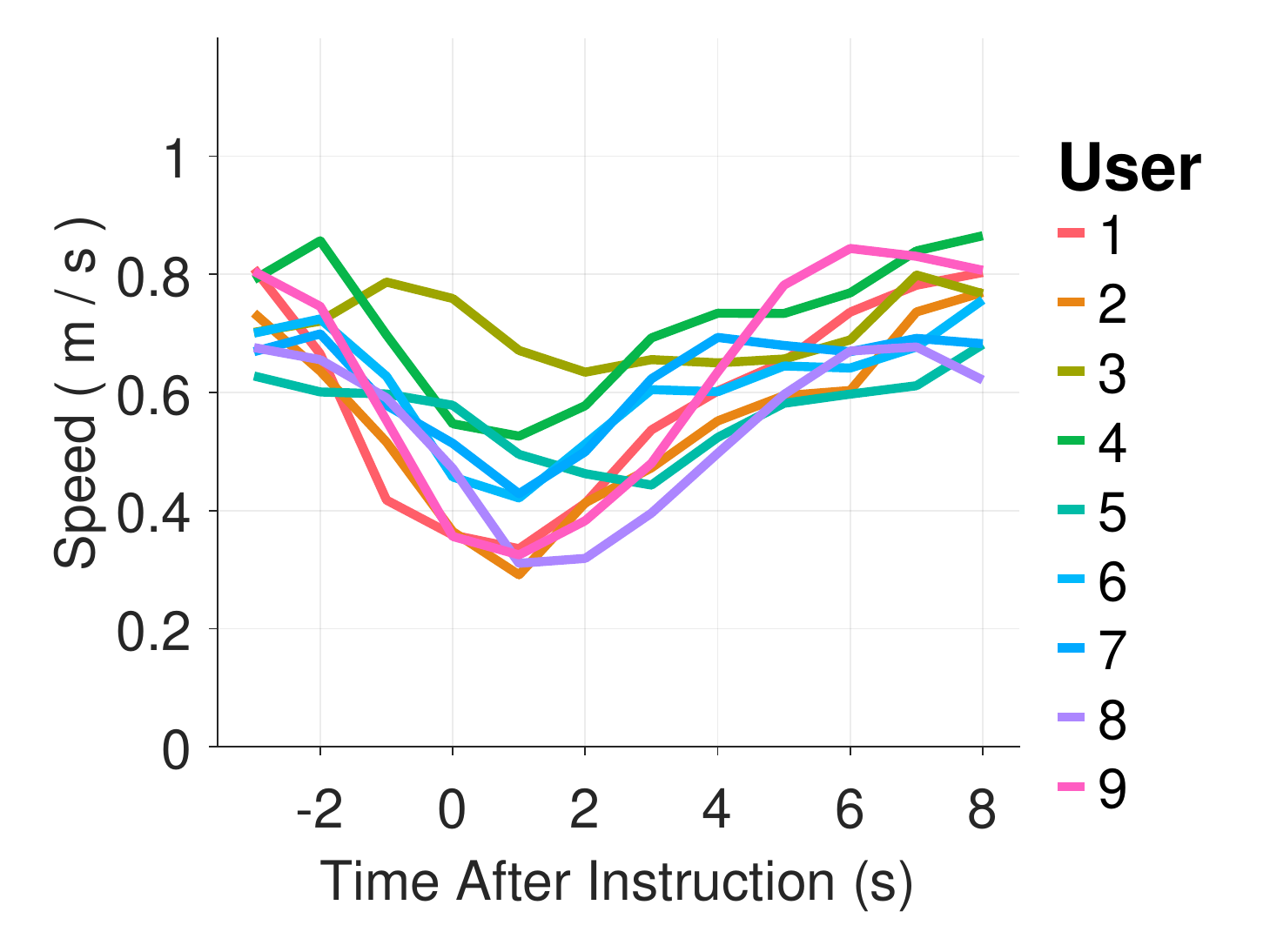}
&
 \includegraphics[width=1.5in,trim={0cm 0 2.8cm 0},clip]{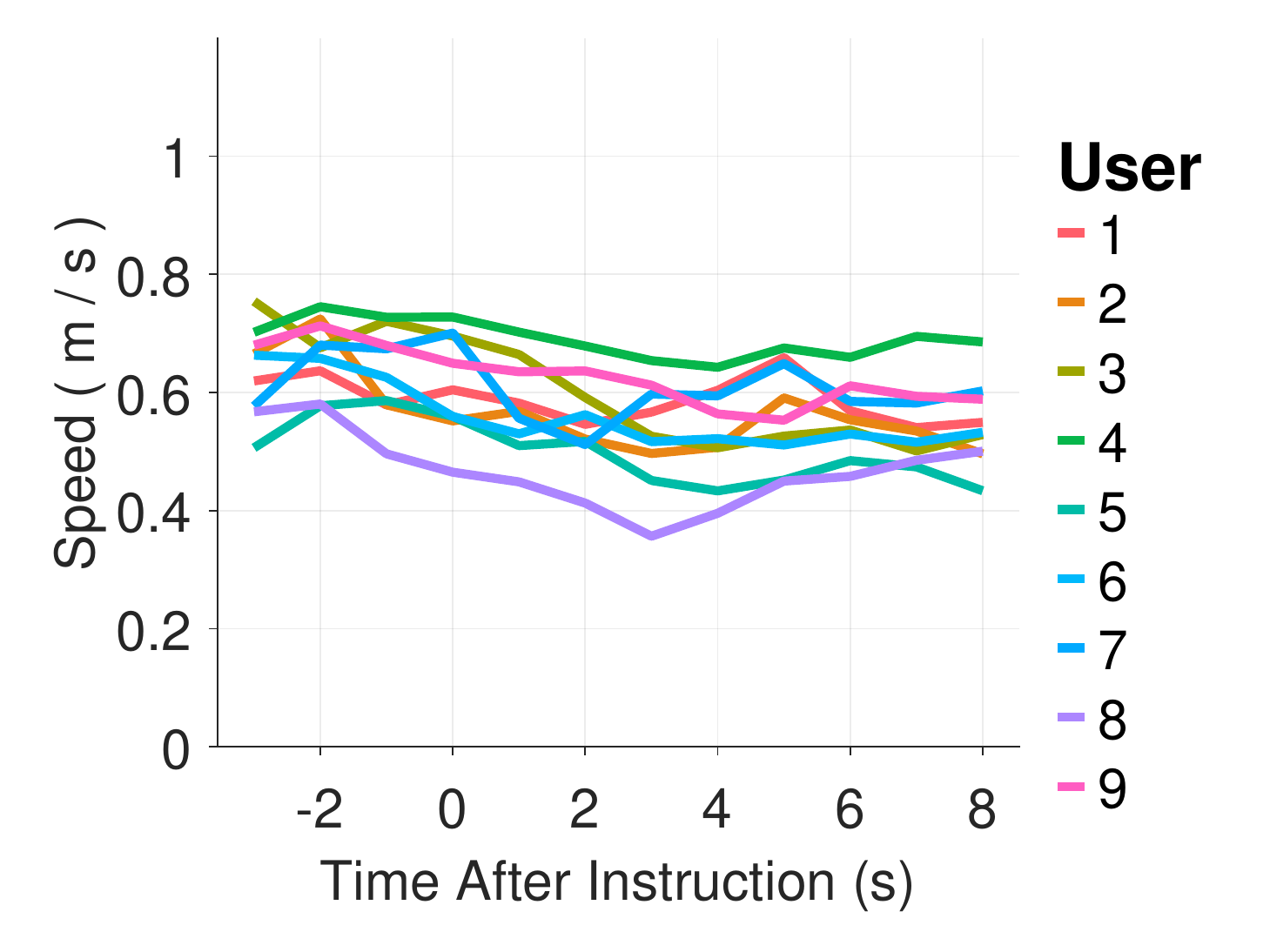}
 &
   \includegraphics[width=0.2in,trim={0cm 2cm 13cm 0},clip]{fig/legp.pdf} 
  \\(d) `Forward' instruction & (e) `Information' instruction &  \\
   \includegraphics[width=1.5in,trim={0cm 0 2.8cm 0},clip]{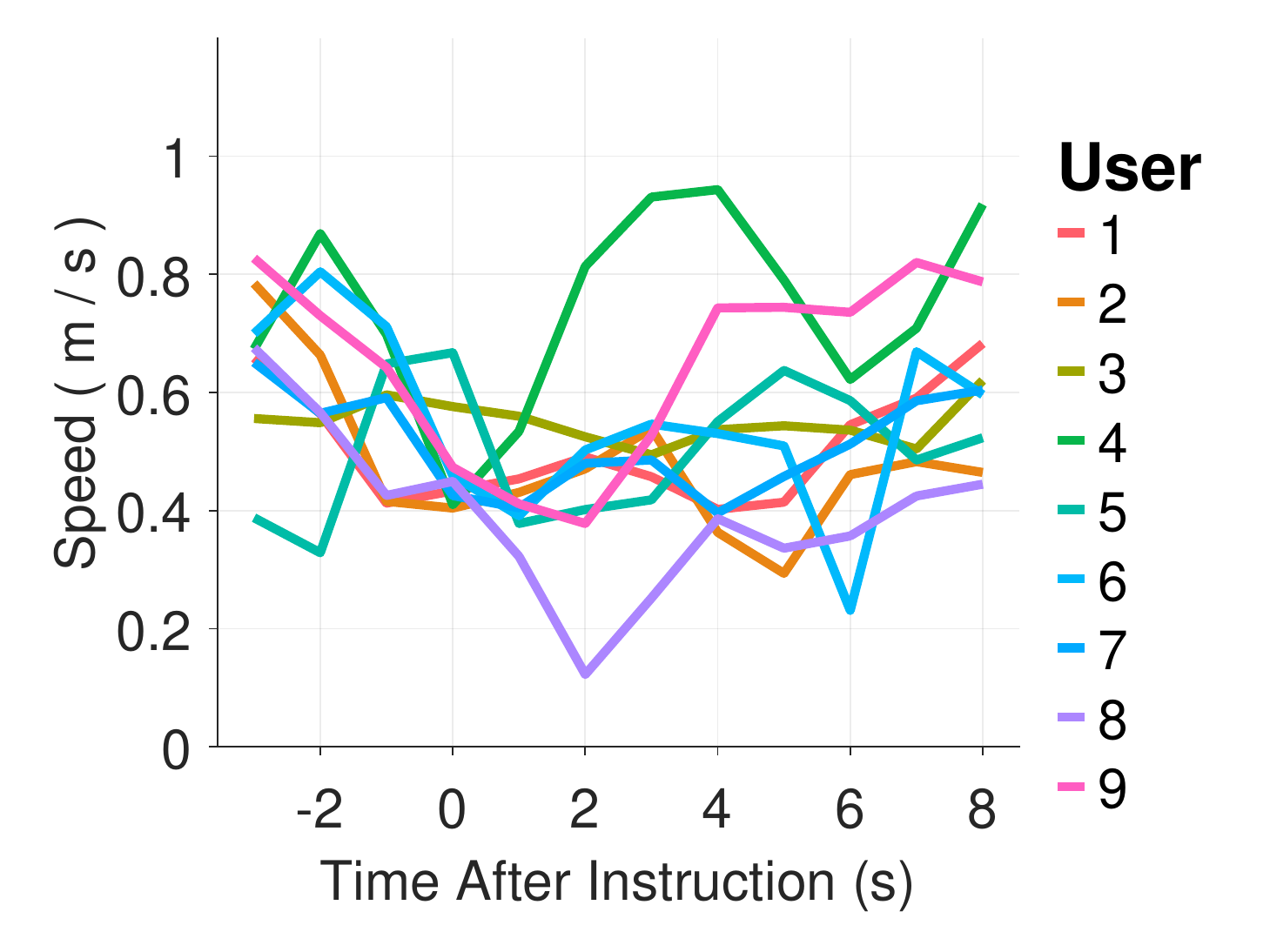}&
  \includegraphics[width=1.5in,trim={0cm 0 2.8cm 0},clip]{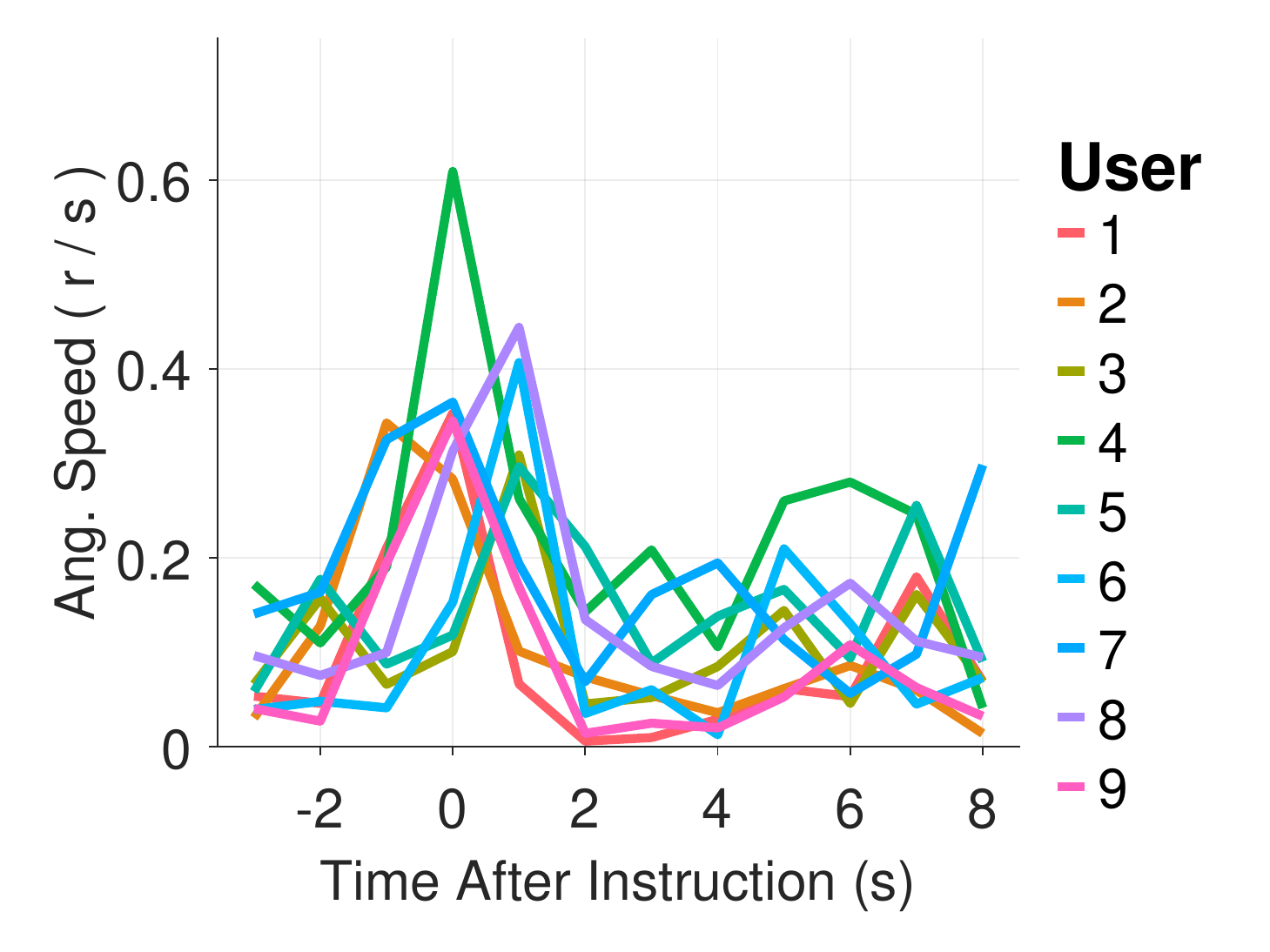}& 
    \includegraphics[width=0.2in,trim={0cm 2cm 13cm 0},clip]{fig/legp.pdf} 
  \\  \multicolumn{3}{c}{(f) `Obstacle' instruction}
    \end{tabular}
    }
  \caption{Per-user averaged (across all instances in the dataset) speed statistics for different notifications (onset at t=0).}~\label{fig:figlabel1}
\end{figure}

 \subsection{Walker Visualization}
\begin{figure}[!t]
  \centering
  \begin{tabular}{ccc}
  \includegraphics[width=1.5in]{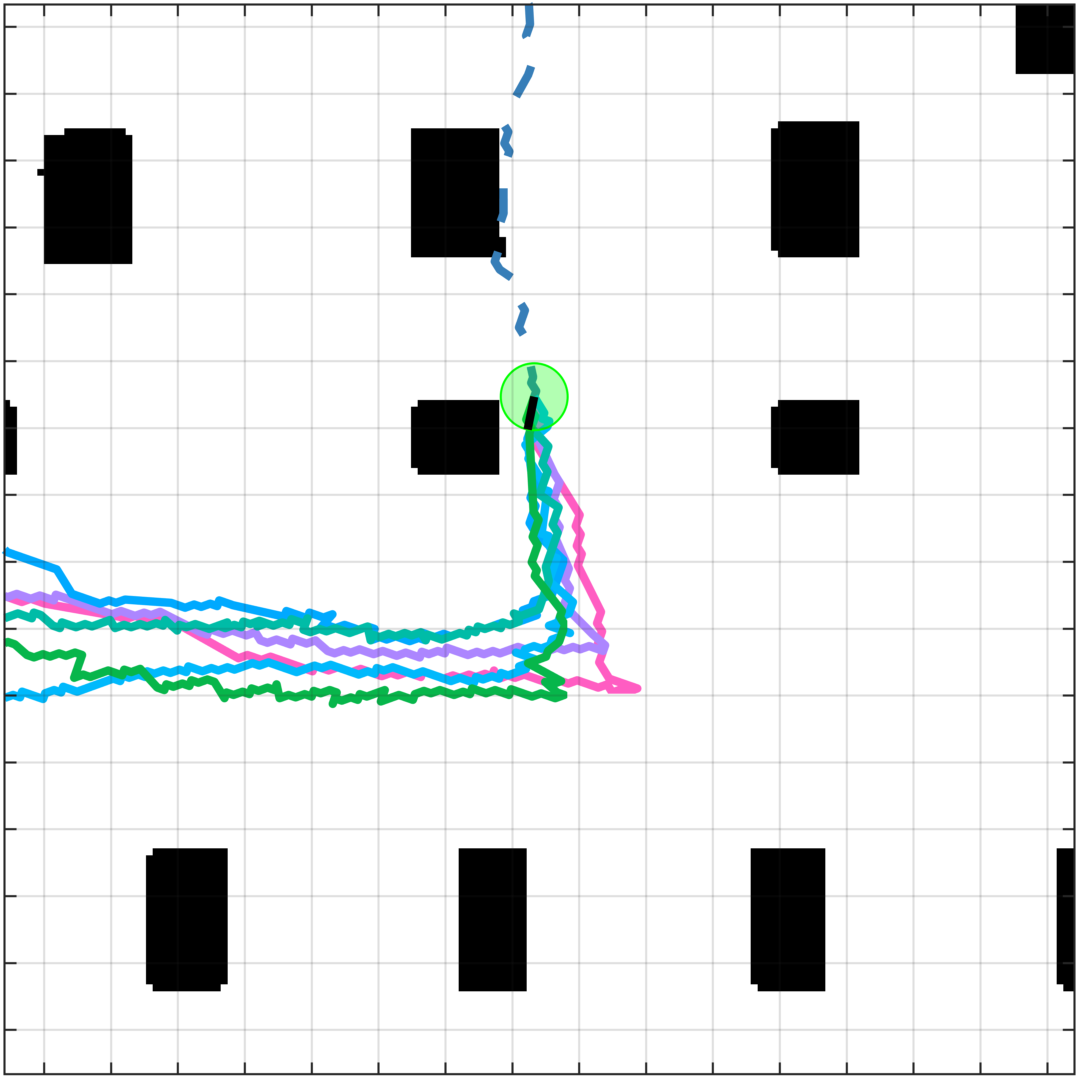}&
  \includegraphics[width=1.5in]{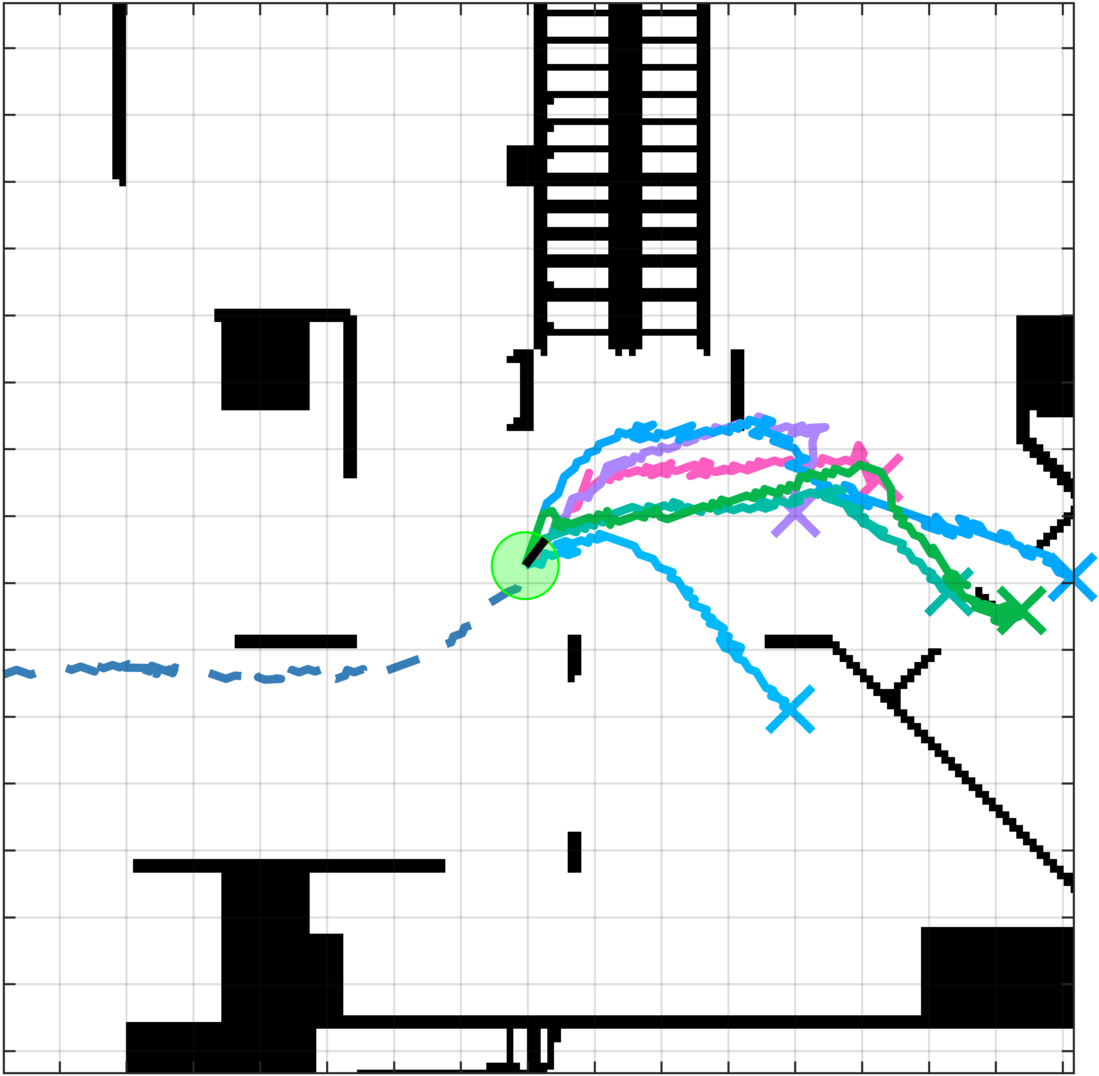}&
  \includegraphics[width=1.5in]{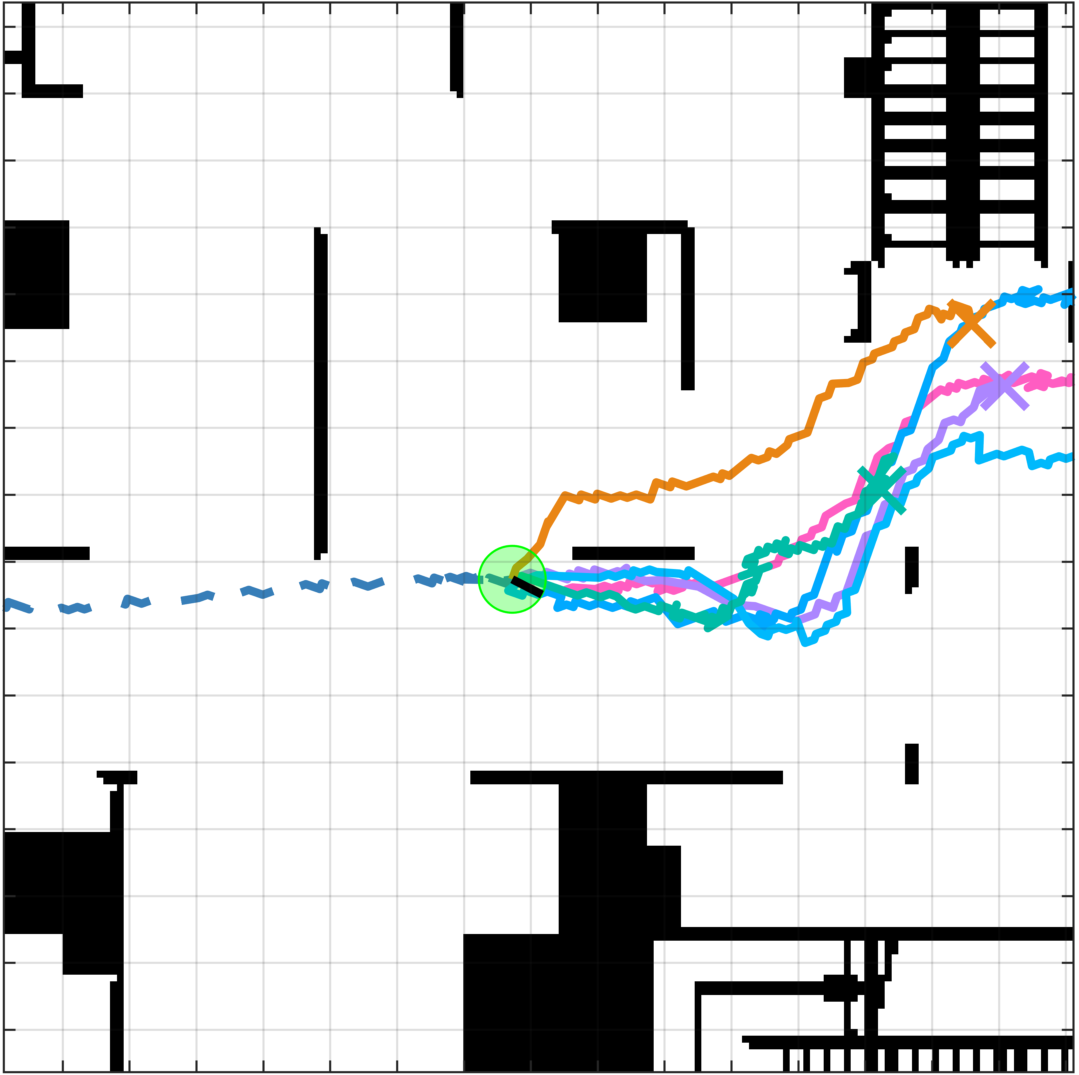}
    \end{tabular}
  \caption{Example diverse walkers simulated in the real-world environmental layout for evaluating PING.}~\label{fig:exwalkers}
\end{figure}

To evaluate PING we use the real-world motion statistics and environment to generate a large number of walkers (Fig.~\ref{fig:exwalkers}) with greater motion diversity and measure cumulative reward. 

 